\documentclass[acmsmall,screen]{acmart}

\AtBeginDocument{%
  \providecommand\BibTeX{{%
    \normalfont B\kern-0.5em{\scshape i\kern-0.25em b}\kern-0.8em\TeX}}}

\setcopyright{rightsretained}
\acmYear{2024} \acmVolume{1} \acmNumber{1} \acmArticle{1} \acmMonth{1}\acmDOI{10.1145/3657632}

\acmJournal{CSUR}

\begin{document}

\title{Visual Tuning}

\author{Bruce X.B. Yu}
\authornote{Both authors contributed equally to this research.}
\affiliation{%
  \institution{Zhejiang University-University of Illinois Urbana-Champaign Institute, Zhejiang University}
  \country{China}
}
  
\author{Jianlong Chang}
\authornotemark[1]
\affiliation{%
  \institution{Huawei}
  \country{China}}

\author{Haixin Wang}
\affiliation{%
  \institution{National Engineering Research Center for Software Engineering, Peking University}
  \country{China}
}

\author{Lingbo Liu}
\affiliation{%
  \institution{Peng Cheng Laboratory}
  \country{China}}

\author{Shijie Wang}
\affiliation{%
  \institution{Huawei}
  \country{China}}

\author{Zhiyu Wang}
\affiliation{%
 \institution{Huawei}
 \country{China}}

\author{Junfan Lin}
\affiliation{%
  \institution{Peng Cheng Laboratory}
  \country{China}}

\author{Lingxi Xie}
\affiliation{%
  \institution{Huawei}
  \country{China}}

\author{Haojie Li}
\affiliation{%
  \institution{College of Computer Science and Engineering, Shandong University of Science and Technology}
  \country{China}}
  
\author{Zhouchen Lin}
\affiliation{%
  \institution{National Key Lab of General AI, School of Intelligence Science and Technology, Peking University}
  \country{China}}
\affiliation{%
  \institution{Peng Cheng Laboratory}
  \country{China}}
  
\author{Qi Tian}
\authornote{Corresponding author(s).}
\affiliation{%
  \institution{Huawei}
  \country{China}}
\email{tian.qi1@huawei.com}

\author{Chang Wen Chen}
\authornotemark[2]
\affiliation{%
  \institution{Department of Computing, The Hong Kong Polytechnic University}
  \country{Hong Kong}}
\email{changwen.chen@polyu.edu.hk}

\renewcommand{\shortauthors}{Bruce and Jianlong, et al.}

\begin{abstract}
Fine-tuning visual models has been widely shown promising performance on many downstream visual tasks. With the surprising development of pre-trained visual foundation models, visual tuning jumped out of the standard modus operandi that fine-tunes the whole pre-trained model or just the fully connected layer. Instead, recent advances can achieve superior performance than full-tuning the whole pre-trained parameters by updating far fewer parameters, enabling edge devices and downstream applications to reuse the increasingly large foundation models deployed on the cloud. With the aim of helping researchers get the full picture and future directions of visual tuning, this survey characterizes a large and thoughtful selection of recent works, providing a systematic and comprehensive overview of existing work and models. Specifically, it provides a detailed background of visual tuning and categorizes recent visual tuning techniques into five groups: fine-tuning, prompt tuning, adapter tuning, parameter tuning, and remapping tuning. Meanwhile, it offers some exciting  research directions for prospective pre-training and various interactions in visual tuning.
\end{abstract}



\received{6 June 2023}
\received[revised]{3 April 2024}

\maketitle

\section{Introduction}\label{sec:introduction}
Since the wide usage of Transformer models \cite{han2022survey, khan2022transformers} and the emergence of large foundation models \cite{bommasani2021opportunities, zhou2023comprehensive}, the paradigm of deep learning in vision intelligence has been experiencing the hype of adapting downstream tasks to the foundation models. 
The astonishing performance of the recent Visual ChatGPT \cite{wu2023visual} is enabled with myriad computation resources in the pre-training process \cite{brown2020language}, and human feedback during the tuning process. The pre-trained foundation model (i.e., GPT-3) shows strong capability but entails large storage space, around 800GB, to store 175B parameters \cite{brown2020language}, which makes it expensive to retrain independent model copies for different downstream tasks. Foundation models are expected to continue to scale up, and how to reuse the foundation model via parameter-efficient transfer learning (PETL) methods (e.g., prompt, prefix, adapter, etc.) quickly becomes a research hype. In the past two years, taking the inspiration of PETL methods in natural language processing (NLP) \cite{tay2022efficient, liu2021pre, zhou2023comprehensive}, numerous visual tuning techniques have been proposed for adapting downstream tasks to pre-trained vision or visual-language models. 

\begin{figure*}
	\begin{center}
		\includegraphics[width=\linewidth]{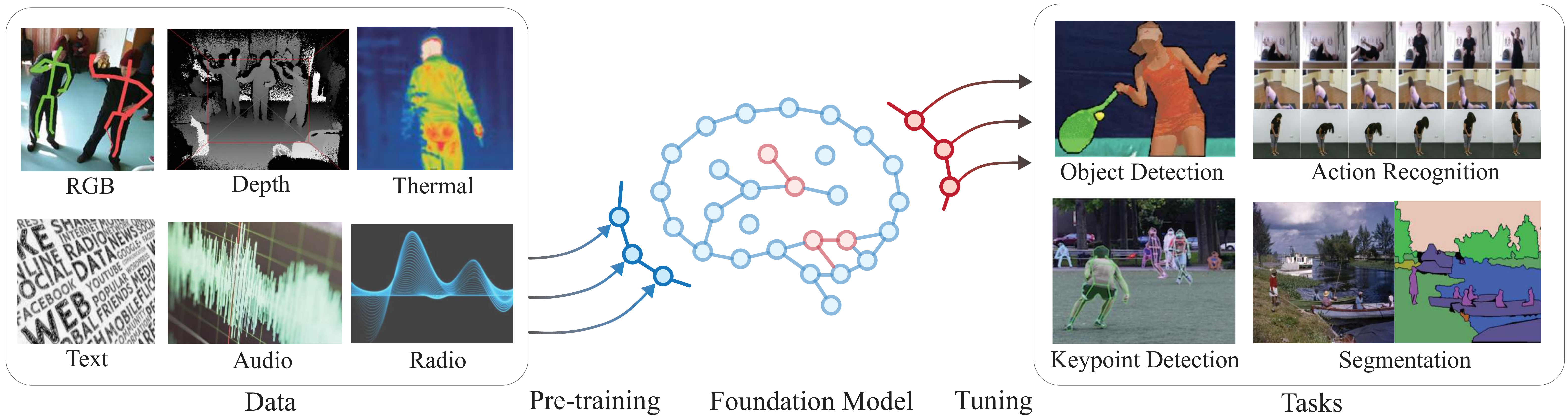}
	\end{center}
	\caption{Illustration of visual tuning. A pre-trained foundation model can accumulate knowledge via various pre-training techniques by scaling up in terms of model size, data modalities, tasks, etc. Given the pre-trained model, the focus of this survey is visual tuning, showing how to effectively reuse the knowledge of the pre-trained models by concerning important aspects such as tuned parameters, generalization ability, data efficacy, training memory, and inference memory, etc. }
	\label{fig:motivation}
\end{figure*}

In the era of increasingly large models, vision models have been scaled up from EfficientNet-based \cite{pham2021meta} ($480$M parameters) to Transformer-based \cite{yu2022coca} ($2,100$M parameters) and even larger scales such as $22$B parameters \cite{dehghani2023scaling} and $562$B \cite{driess2023palm}. For such large models, PETL methods aim to make good reuse of the shared parameter weights (usually interpreted as the knowledge of large models) deployed on the cloud to save storage overhead and to empower edge devices such as autonomous vehicles, drones, and robots that are intensive in computing and battery resources \cite{yuan2022roadmap}. This practice is different from the modus operandi of transfer learning that either fully fine-tunes the whole model or just fine-tunes the task head (e.g., the last fully connected layer)  \cite{zhuang2020comprehensive}. 

Given the emergence of increasingly large models (i.e., foundation models), we are in a new paradigm of visual tuning that is beyond tuning the entire model or the task head. How to effectively reuse the knowledge thereof with PETL methods, leading to less memory usage and higher inference speed is a hot topic in various vision tasks \cite{sung2022vl,lin2020exploring,sung2022lst}. 
Starting with a detailed background in Section \ref{section:background}, this paper provides an in-depth review of recent tuning advances in the vision domain, categorizing them into five common types and elaborating their current technical state with discussions in Section \ref{sec:petl_techniques}. Last but not the least, we provide insights into future research directions that hold significant promise in Section \ref{sec:visual_future}, followed by a conclusion. To the best of our knowledge, this is the first comprehensive survey on visual tuning, which bears great importance for researchers to understand the mechanisms and trends of this practice. 

\section{Background}\label{section:background}
In the early days, machine learning methods relied on feature engineering such as SIFT \cite{ke2004pca}, BRIEF \cite{calonder2010brief}, and ORB \cite{rublee2011orb} to handle specific tasks, which is later on dominated by the deep learning paradigm \cite{lecun2015deep} since the introduction of ImageNet \cite{deng2009imagenet}. Deep learning models \cite{szegedy2015going, he2016deep, krizhevsky2017imagenet} pre-trained on ImageNet are able to benefit various downstream vision tasks such as image recognition, object detection, and image segmentation via fine-tuning. Fine-tuning is the second step of typical transfer learning, which makes use of the knowledge acquired from the source domain to facilitate the learning process of the target domain \cite{tan2018survey,zhuang2020comprehensive}. 
Given the promise of large-scale pre-trained models, visual tuning techniques beyond fine-tuning have attracted increasing research interest, leading to the visual tuning paradigm as illustrated in Fig. \ref{fig:motivation}. This section will elaborate on the background of visual tuning from five perspectives: 
theory, definition, model architecture, model pre-training, and model tuning.

\subsection{Theories}
In the 1990s, the machine learning community largely ignored neural networks and backpropagation due to concerns about overfitting and the potential for poor local minima. However, in the recent era of deep learning, these concerns have been greatly alleviated via advancements in theories and empirical experiences \cite{lecun2015deep}. In this section, we present the fundamental theories that underpin the current state of visual tuning, exploring these theories from three distinct perspectives as follows.

\subsubsection{Biological Perspective}\label{sec:background_theory_biological}
Like the origin convolutional neural network (CNN) architecture was inspired by the receptive fields in the visual cortex \cite{hubel1959receptive}, learning models can be inspired or motivated by biological and neuroscience discoveries. 
Particularly, researchers are working on enabling computer vision to have capabilities that are similar to human vision. 
First, human vision can efficiently process huge amounts of continuous visual streams. Regarding the intrinsic mechanism of this ability, classical biological findings suggest that humans perceive real-world scenes by contextualizing information from local parts (such as small edges) as a whole (i.e., subjective contours), which are respectively handled by cortical areas V1 and V2 \cite{marr2010vision}. It is also suggested that human vision is embodied and developed in interactive ecological environments \cite{gibson2014ecological}. This motivates researchers to work on effective solutions concerning the aspects such as accuracy and efficiency. Second, humans are good at generalizing visual understanding to unseen brand-new scenes by reasoning their physical and geometric properties \cite{lake2015human}.   This motivates emerged foundation models to be tested via increasingly challenging setups such as zero-shot learning, continual learning, multi-task learning, etc. 
 
\subsubsection{Model Perspective}\label{sec:background_theory_model}
Taking inspiration from human vision, the recently emerged paradigm of tuning large vision models aims to effectively reuse the knowledge in the large pre-trained model in an efficient way regarding computation and data. 
Generative pre-trained large language models such as GPT-3 show significant continual performance improvements when the model size is scaling up from 0.1B to 175B parameters \cite{brown2020language}. This observation is known as the \textit{scaling up law}: larger pre-trained model will benefit downstream tasks, which shows insights that adapting from a larger knowledge base can lead to better performance for downstream tasks. This \textit{scaling up law} has also been proved in recent literature \cite{liu2023ten, dehghani2023scaling}.
Sung \textit{et al.} \cite{sung2022lst} elaborated on the reason for the reduced training memory of PETL techniques from the perspective of backpropagation and further reduced their training memory by skipping the gradient traversal through the frozen backbone, which steps further on the analysis of existing PETL technique regarding training and inference memories. 

\subsubsection{Statistical Perspective}\label{sec:background_static}
Machine learning models are restricted by some statistical assumptions such as independent and identically
distributed, the law of large numbers, central limit theorem, etc. \cite{dekking2005modern}, making practitioners conduct regularization techniques, collect large-scale datasets, and normalize the input data,  respectively. 
In the era of large models, breaking these statistical boundaries becomes imaginable with encouraging recent progress (surveyed in Section \ref{sec:petl_techniques}), which intrinsically improves models' generalization ability to out-of-distribution or long-tail data with less training data (from few-shot to zero-shot learning) and tunable parameters. To guide tuning with statistical rules, there are some works proposed based on measurable domain bound. For instance,  
Ye \textit{et al.} \cite{ye2021towards} proposed the concept of expansion function, quantifying regularization or bound restrictions as the “variation” between the source and target domains, and the “informativeness” of a feature.  
Liu and Zhang \cite{liu2022closer} also attempted to measure the domain gap by using the test error. 
Zhang \textit{et al.} \cite{zhang2019bridging} proposed to use the margin loss to replace the 0-1 loss for domain adaptation. The margin loss is expected to relax the restriction and provide a more informative generalization bound. 
Nilesh \textit{et al.} \cite{tripuraneni2020theory} defined task diversity from a statistical perspective, providing generalization upper bounds of sample complexity for multi-task transfer learning.

\begin{table}[t]
\setlength{\tabcolsep}{1.8em}
\caption{Notations used in the paper.}
\label{table_running_time}
\renewcommand{\arraystretch}{1}
\centering
\begin{tabular}{cl}
\hline
 \bfseries Symbol  & \bfseries Definition   \\
\hline
    $m$ & Number of domains \\
    $P$ & Distribution   \\
    $\mathcal{D}$ & Domain   \\
    $S$ & Source domain   \\
    $T$ & Target domain   \\
    $\mathcal{X}$ & Feature space \\
    $\mathcal{Y}$  & Label space \\
    $X$ & Instance set \\
    $Y$  & Label set \\

    $N$  & Number of samples in $X$  \\

    $a$ & Learnable vectors \\
    $M$ & Number of prompts \\
    $\mathcal{T}$ & Tokens of visual inputs \\


    $Z$  & Feature propagate through network \\
    $l$  & Neural network layer \\
    $k$  & Input size of a convolutional layer \\
    $d$  & Output size of a convolutional layer \\
    $K$  & Kernel size \\
    $G$  & Group size \\
    $r$  & Dimension of low rank \\
    
\hline 
\end{tabular}
\end{table}

\subsection{Notation and Definition}\label{sec:background_notation}
In order to understand efficient fine-tuning, let's start by defining domains, tasks, transfer learning, and other notations. 
A joint distribution  $\mathcal{X} \times \mathcal{Y}$ can be expressed as $P(X,Y)$ (i.e., $P_{XY}$), where $\mathcal{X}$ and $\mathcal{Y}$ represent its corresponding feature space and label space, respectively. ( $X$ and $Y$ represent the observed instance set and its corresponding label set. )
 Given $P_{XY}$, we refer $P(X)$ (i.e., $P_{X}$) as the marginal distribution on $X$, $P_{Y|X}$ the posterior distribution of $Y$, and $P_{X|Y}$ the class-conditional distribution of $X$ given $Y$.
 
\textbf{Definition 1} (Domain): 
A domain $\mathcal{D}=\{\mathcal{X},P(X)\}$ is defined by its feature space $\mathcal{X}$  and a marginal distribution $P(X)$, where $X$ denotes an instance set defined as $X = \{x| x_i \in \mathcal{X} , i=1,...,n\}$. A domain can be with or without labeling information.

\textbf{Definition 2} (Task): 
A task can be denoted as $\mathcal{T}=\{\mathcal{Y},f\}$, where  $\mathcal{Y}$  and $f$ represent a label space and a decision function, respectively. 
For the classification task of a source domain $\mathcal{T^S}$, the goal is usually to predict the conditional distribution of instances, which can be denoted as $f(x_j)=\{P(y_k|x_j)|y_k \in \mathcal{Y}, k=1,..., |\mathcal{Y}|\}$.  In this case, the task $\mathcal{T^S}$ can be regarded as forming a typical source domain $\mathcal{D^S}$ with labeling information, being denoted as $(\mathcal{D^S},\mathcal{T^S}) = \{ (x,y) | x_i \in  \mathcal{X}^S, y_i \in  \mathcal{Y}^S, i=1,..., n^S \}$.

\textbf{Definition 3} (Transfer learning): 
Given $m^S \in \mathbb{N}^+$ source domain(s) and $m^T \in \mathbb{N}^+$ target domain(s), their corresponding task(s) can be denoted as $\{ (\mathcal{D}^{S_i}, \mathcal{T}^{S_i}) | i=1,...,m^S \}$ and $\{ (\mathcal{D}^{T_j}, \mathcal{T}^{T_j}) | j=1,...,m^T \}$, respectively. Transfer learning aims to improve the performance of decision functions $f^{T_j}$ on the target domain(s) by making good use of the knowledge learned from the source domain(s). 

\textbf{Definition 4} (Parameter efficient fine-tuning): 
Given $m^S \in \mathbb{N}^+$ source domain(s) $\{ (\mathcal{D}^{S_i}, \mathcal{T}^{S_i}) | i=1,...,m^S \}$ and $m^T \in \mathbb{N}^+$ target domain(s) $\{ (\mathcal{D}^{T_j}, \mathcal{T}^{T_j}) | j=1,...,m^T \}$ defined in transfer learning, the goal of efficient fine-tuning $\{ (\mathcal{D}^{T_j}, \mathcal{Y}^{T_j}, f^{T_j}, f^{S_i})| i=1,...,m^S, j=1,...,m^T \}$ is to improve the performances of $f^{T_j}$ by reusing  ${f^{S_i}|i=1,...,m^S}$ learned from their corresponding tasks denoted as $P_{XY}$. In particular, the parameters of $f^{S_i}$ need to be frozen or tuned with a small portion. While $f^{T_j}$ denotes an extra small amount of model parameters that can be easily deployed on edge devices. 
In practice, for supervised or self-supervised pre-training, $f^{S_i}$ can be learned from $P_{Y|X}$ and $P_X$, respectively. 
This definition is an extension of typical transfer learning \cite{zhuang2020comprehensive}, which covers multisource efficient fine-tuning. 

\subsection{Model Architecture}
Pre-trained foundation models for vision have been surveyed in \cite{zhou2023comprehensive}, which develops from CNN- and GAN-based models to recent Transformer-based models. We recommend readers refer to \cite{zhou2023comprehensive} for the detailed pre-training strategies. This section briefly introduces these representative models' basic structures: CNN-based, Transformer-based, and CNN+Transformer.
 
CNN is one of the most popular deep learning models such as AlexNet \cite{krizhevsky2017imagenet}, VGGNet \cite{simonyan2014very}, Inception \cite{szegedy2016rethinking}, ResNet \cite{he2016deep}, EfficientNet \cite{tan2019efficientnet}, etc., which has been surveyed time to time \cite{albawi2017understanding,li2021survey}. EfficientNet is lightweight yet can achieve comparable performance to Transformer-based models via pre-trained initialization on various visual tasks such as image classification \cite{yuan2021incorporating} and video understanding \cite{bruce2022mmnet}. Except for 2D CNN, a couple of 3D CNN models such as C3D \cite{tran2015learning}, I3D \cite{carreira2017quo}, S3D \cite{xie2018rethinking}, and X3D \cite{feichtenhofer2020x3d} have been introduced for video understanding  tasks. In addition, graph convolutional network \cite{yan2018spatial} has also been proposed for tasks such as exercise evaluation \cite{bruce2024egcn} and pose estimation \cite{yu2023gla}. 

The typical architecture of a Transformer model is structured with several basic Transformer layers. Each layer can be made of a varied number of Transformer blocks composed of a multi-head self-attention (Attention) module and a fully connected feed-forward network (FFN) implemented with a 2-layer multilayer perceptron (MLP). Layer normalization (LN) and residual connection are respectively performed before and after both FFN and Attention modules. 
Building upon the basic Transformer model, Transformer has been dominating increasing tasks \cite{han2022survey, khan2022transformers}. Early Transformer models for vision are  Vision Transformer (ViT) \cite{dosovitskiy2020image}, Data-efficient image transformer
(DeiT) \cite{touvron2021training}, while their representative variations are TNT \cite{han2021transformer}, T2T \cite{yuan2021tokens}, PVT \cite{wang2021pyramid}, Swin-Vit \cite{liu2021swin}, Video Swin Transformer \cite{liu2022video}, CPVT \cite{chu2021conditional}. 

Transformer models are well known for their ability to capture long-range dependencies of input data. Whereas CNNs might be better at representing local features. Models combining Transformer and CNN can achieve better performance. 
Twins-SVT \cite{chu2021twins} proposed to add a positional encoding module implemented via a 2D depth-wise convolution 2D in between the Transformer encoders, and designed global and local attention modules to improve the model's representation ability, leading to improved performance on image-based tasks with slightly more model parameters. Representative methods combining CNN and Transformer are Shuffle \cite{huang2021shuffle},
CMT \cite{guo2022cmt},
VOLO \cite{yuan2022volo}, etc. Although they can achieve superior performance, how they can be used for visual tuning seems under-explored.

\subsection{Model Pre-training}\label{sec:background_pretrain}
Pre-training methods can be roughly grouped into supervised and self-supervised ones. 
Early vision models were pre-trained via supervised learning on large-scale datasets such as ImageNet \cite{deng2009imagenet}, JFT-300M \cite{sun2017revisiting}, Kinetics \cite{kay2017kinetics}, etc. Since fine-tuning models pre-trained with supervised learning, larger-scale pre-training has been conducted recently. For example, 
Gato \cite{reed2022generalist} uses multi-task learning with the supervision of varied tasks to enable the large model to acquire more knowledge for the adaptation of downstream tasks. 
Multi-label learning is used to pre-train a pure vision model that reaches 22B parameters  \cite{dehghani2023scaling},  showing fantastic performance on downstream visual tasks.

In the regime of supervised pre-training, the non-trivial annotation cost imposes a practical obstacle to scaling up the benefit of transfer learning. Alternatively, self-supervised learning on unannotated data can also make the models richer and potentially more useful  \cite{bommasani2021opportunities}.  The paradigm of fine-tuning models pre-trained via self-supervision brings the possibility of learning knowledge from unannotated data at a larger scale, which is enabled by advanced computing power, the Transformer model, and more data. Models pre-trained with self-supervised learning are termed “foundation models” by Bommasani \textit{et al.} \cite{bommasani2021opportunities}. 
Recent notable examples include MAE \citep{he2022masked,tong2022videomae} in vision; CLIP \cite{radford2021learning},  ALIGN \cite{jia2021scaling}, Florence \cite{yuan2021florence}, BEiT \cite{wang2022image}, Gato \cite{reed2022generalist}, CoCa \cite{yu2022coca}, SWAG \cite{singh2022revisiting}, etc. for visual-language models. More recently, generative models such as NeRF \cite{mildenhall2021nerf} and Diffusion \cite{peebles2023scalable} have also been fine-tuned for better image or video generation such as Latent-NeRF \cite{metzer2023latent}, DreamBooth \cite{ruiz2023dreambooth}, and Tune-a-video \cite{wu2023tune}. Taking the initial success in NLP, this paradigm has started showing success in vision and various other realms such as climate science \cite{nguyen2023climax}, protein design \cite{verkuil2022language}, intelligent transportation \cite{shi2023open}, etc. 
Bommasani \textit{et al.} \cite{bommasani2021opportunities} identified the key significance of foundation models as \textit{emergence} regarding capability and \textit{homogenization} regarding model, modality, tasks, and domains. 

\begin{table*}[htbp]
\footnotesize
\caption{A comprehensive review and classification of visual tuning methods. Red and blue parts are tunable and frozen parameters, respectively.}
\label{tab:efficient_finetune_methods}

\renewcommand{\arraystretch}{1.}
\centering
\resizebox{\linewidth}{!}{%
\begin{tabular}{ccc}

\toprule%
 \bfseries Category  & \bfseries Description & \bfseries Method  \\
 \midrule

   \begin{minipage}{.24\textwidth}
      \includegraphics[width=0.88\linewidth]{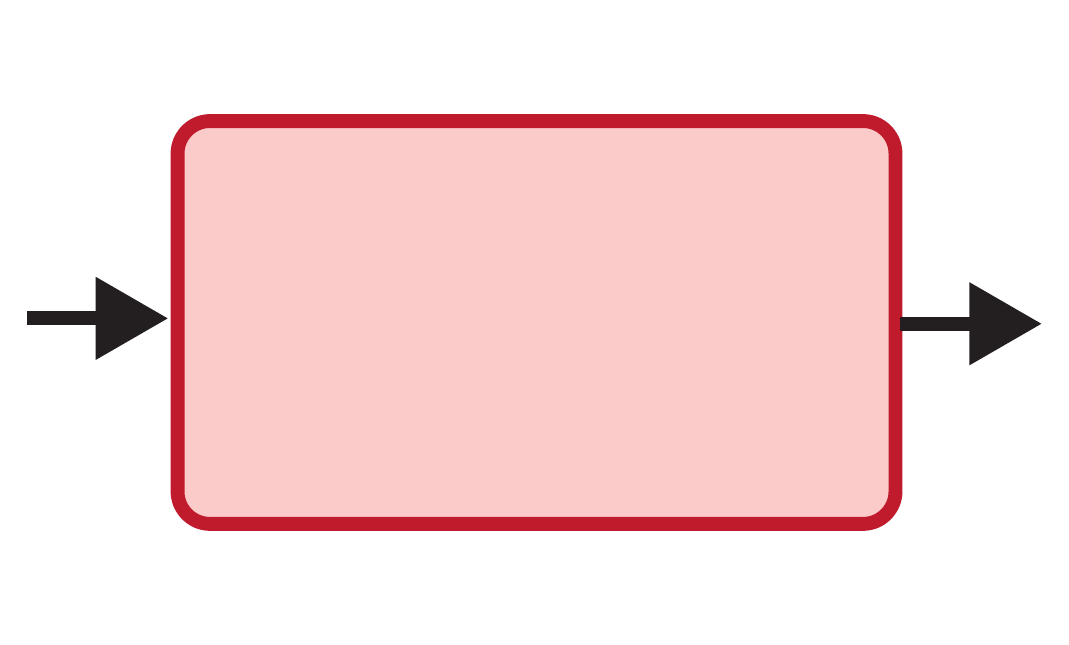}
      \centering
      Fine-tuning
    \end{minipage}

& 
    \begin{minipage}{.32\textwidth}
    All parameters in the pre-trained model are updated in the tuning process. This method is by far regarded as an effective practice to achieve state-of-the-art performance on many vision benchmark datasets. However, when vision models continue to scale up, this fine-tuning method becomes less practicable due to the storage and training overhead.
    \vspace{3pt}
    \end{minipage} 
& 
 \begin{minipage}{.52\textwidth}

    \textbf{CNN}: 
        VGGNet \cite{simonyan2014very}, 
        Inception \cite{szegedy2016rethinking}, 
        ResNet \cite{he2016deep}, 
        EfficientNet \cite{tan2019efficientnet}, 
        C3D \cite{tran2015learning}, 
        I3D \cite{carreira2017quo}, 
        S3D \cite{xie2018rethinking}, 
        X3D \cite{feichtenhofer2020x3d}
    
    \textbf{Transformer}:
        ViT \cite{dosovitskiy2020image}, 
        DeiT \cite{touvron2021training}, 
        TNT \cite{han2021transformer}, 
        T2T \cite{yuan2021tokens}, 
        PVT \cite{wang2021pyramid}, 
        Swin-Vit \cite{liu2021swin}, 
        Video Swin Transformer \cite{liu2022video}, 
        CPVT \cite{chu2021conditional}

    \textbf{CNN and Transformer}: 
        Shuffle \cite{huang2021shuffle},
        CMT \cite{guo2022cmt},
        VOLO \cite{yuan2022volo}
   \end{minipage}  
\\
\midrule
   \begin{minipage}{.24\textwidth}
      \includegraphics[width=\linewidth]{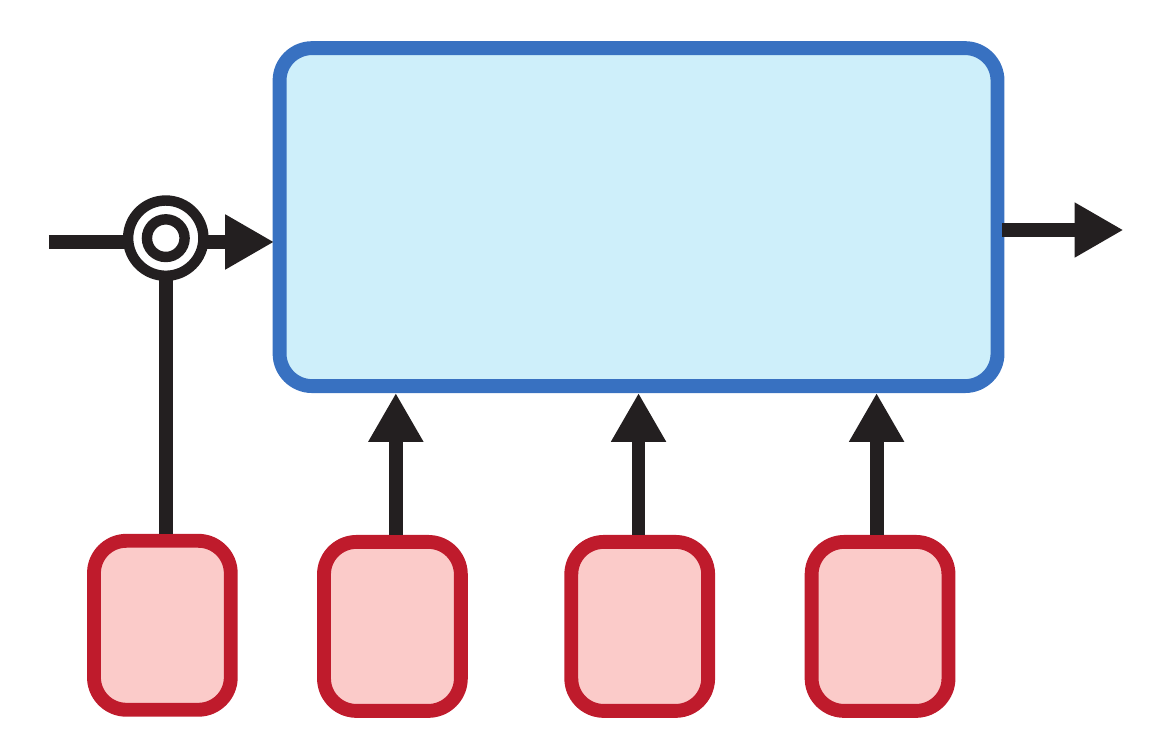}
      \centering 
      Prompt Tuning
    \end{minipage} 
& 
    \begin{minipage}{.32\textwidth}
    Prompt tuning unifies all downstream tasks into pre-trained tasks via designing a specific template to fully exploit the capabilities of foundation models. Prompt tuning usually learns few parameters and keeps pre-trained models frozen. In addition, the core mechanism of the vision prompts aims to exploit the potential of the upstream pre-trained model, so that the upstream pre-trained model can perform the downstream task as well as possible with some or fewer labeled data.

    \end{minipage} 
& 
    \begin{minipage}{.52\textwidth} 
    
    \textbf{Vision-driven Prompt}:
        VPT \cite{jia2022visual}, S-Prompting\cite{wang2022s},
        DePT \cite{gao2022visual},
        ZegCLIP \cite{zhou2022zegclip},
        ACT \cite{dong2022autoencoders},
        PViT \cite{herzig2022promptonomyvit},
        TeCoA \cite{mao2022understanding},
        EVP \cite{wu2022unleashing},
        ProSFDA \cite{hu2022prosfda},
        APT \cite{bowman2023carte},
        PAT \cite{yu2023prompting},
        LPT \cite{dong2022lpt},
        PointCLIP \cite{zhang2022pointclip},
        P2P \cite{wang2022p2p},
        PromptGen \cite{wu2022generative},
        NOAH \cite{zhang2022neural},
        PGN \cite{loedeman2022prompt},
        FPTrans \cite{zhang2022feature},
        FRPT \cite{wang2022fine},
        RePro \cite{gao2023compositional},
        ViLD \cite{gu2021open},
        LION \cite{wang2023lion} 
        
    \textbf{Language-driven Prompt}:  
        CoOp \cite{zhou2022learning}, 
        SubPT \cite{ma2023understanding},
        MPA \cite{chen2022multi},
        ZegOT \cite{kim2023zegot},
        X-CLIP \cite{ni2022expanding},
        ProGrad \cite{zhu2022prompt},
        Berg et. al \cite{berg2022prompt},
        PTP \cite{zhang2022prompting},
        LANIT \cite{park2022lanit},
        SgVA-CLIP \cite{peng2022sgva},
        LASP \cite{bulat2022language},
        DualCoOp \cite{sun2022dualcoop},
        PLOT \cite{chen2022prompt},
        CPL \cite{he2022cpl},
        DeFo \cite{wang2022learning},
        GALIP \cite{tao2023galip},
        CoCoOp \cite{zhou2022conditional},
        PointCLIP V2 \cite{zhu2022pointclip}
    
    \textbf{Vision-language Prompt}:
        UPT \cite{zang2022unified},
        DPT \cite{xing2022class},
        MaPLe \cite{khattak2022maple},
        MVLPT \cite{shen2022multitask},
        MetaPrompt \cite{zhao2022learning},
        TPT \cite{shu2022test}

    \vspace{3pt}  
    \end{minipage}         
        \\
\midrule
    \begin{minipage}{.24\textwidth}
          \includegraphics[width=0.93\linewidth]{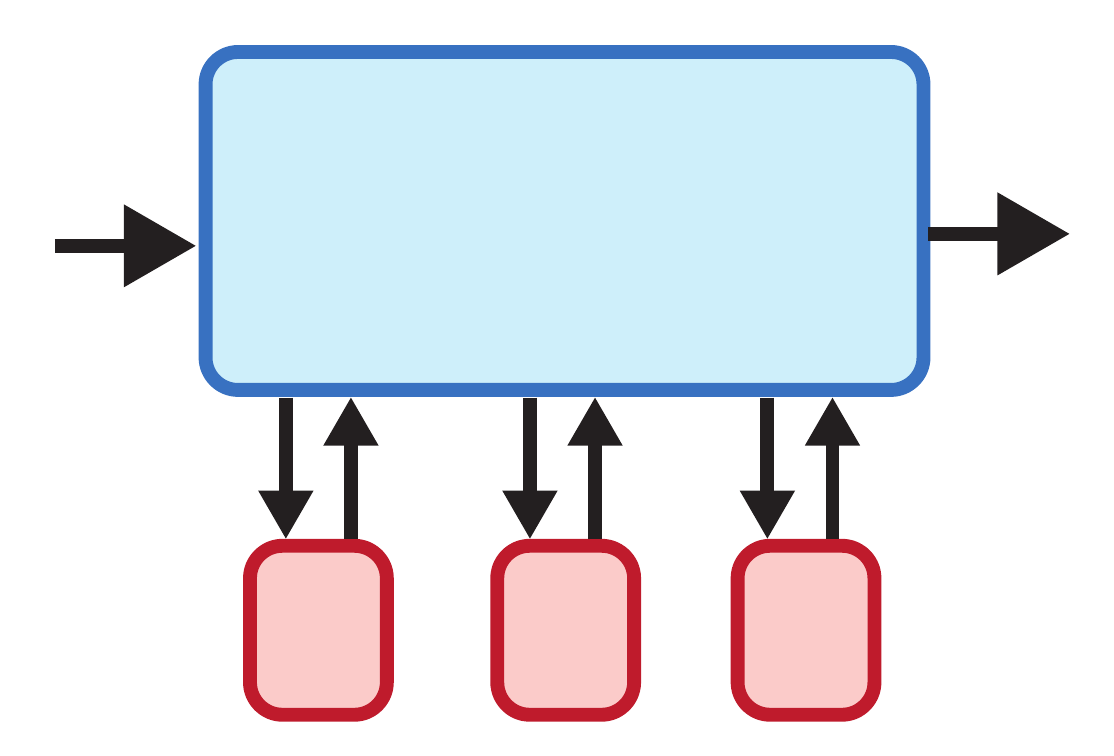}
          \centering 
          Adapter Tuning
    \end{minipage} 
&
       \begin{minipage}{.32\textwidth}
    Adapter tuning is a class of techniques that inserts additional trainable parameters into a pre-trained model frozen to facilitate learning for downstream tasks. The advantage of this method is its lightweight nature and ease of plug-and-play insertion into the middle of a pre-trained network, making it widely applicable in many visual tasks.
    \end{minipage}  
& 
    \begin{minipage}{.52\textwidth}

    \textbf{Sequential Adapter}:
    \vspace{2pt}
        Res-adapt \cite{rebuffi2017learning}, EPM \cite{rebuffi2018efficient}, DAN \cite{rosenfeld2018incremental}, LST \cite{sung2022lst}, Conv-Adapter \cite{chen2022conv}, Polyhistor \cite{liu2022polyhistor}, Pro-tuning \cite{nie2022pro},
        AMixer \cite{rao2022amixer}, Fit \cite{shysheya2023fit}, TINA \cite{marouf2023tiny}, RepAdapter \cite{luo2023towards}, BDTL \cite{li2021benchmarking}, ViTDet \cite{li2022exploring}, Florence \cite{yuan2021florence}, SND \cite{wang2020stacking}, MK-Adapter \cite{zhang2022collaboration}, ADA \cite{Ermis_2022_CVPR}, AIM \cite{yang2023aim}, ST-Adapter \cite{panst}, PEA \cite{sharma2023lossless}, CAOA \cite{tsubota2023universal}, HA \cite{kim2021adapt}, CLIP-Adapter \cite{gao2021clip}, Tip-Adapter \cite{zhang2021tip}, BALLAD \cite{ma2021simple}, MAGMA \cite{eichenberg2021magma}, VL-Adapter \cite{sung2022vl}, Hierarchical3D \cite{papalampidi2022hierarchical3d}, HyperPELT \cite{zhang2022hyperpelt}, SVL-Adapter \cite{pantazis2022svl}, LAVISH \cite{lin2022vision}, CrossModal-Adapter \cite{jiang2022cross}, MV-Adapter \cite{zhang2023multimodal}
    
    \textbf{Parallel Adapter}: 
        ViT-Adapter \cite{chen2022vision}, PESF-KD \cite{rao2022parameter}, AdaptMLP \cite{chen2022adaptformer}, Convpass \cite{jie2022convolutional}, AMA \cite{yu2023rethinking}, UniAdapter \cite{lu2023uniadapter}
   
    \textbf{Mix Adapter}:  
        Consolidator \cite{hao2023consolidator}, ETT \cite{xu2023exploring}, PATT \cite{he2022parameter}, PALT \cite{valipour2022dylora}, TVG \cite{shimomoto2022towards}, VQT \cite{tu2022visual}

    \end{minipage}  
    \\
\midrule
 
 \begin{minipage}{.24\textwidth}
      \includegraphics[width=\linewidth]{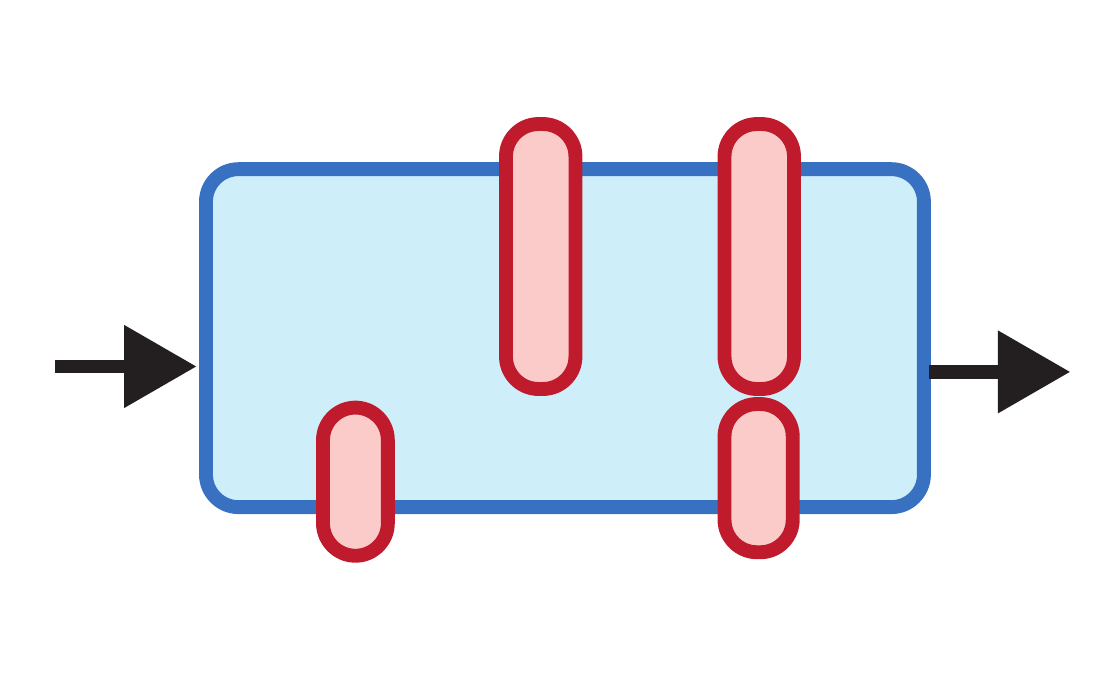}
      \centering 
      Parameter Tuning
    \end{minipage} 
& 
    \begin{minipage}{.32\textwidth}
    Parameter tuning aims to directly modify the model parameters (i.e., weight and bias). They can be grouped into three categories: bias part, weight part, and both. Common modification schemes can be addition, decomposition, or without extra parameters (i.e., directly tune part of parameters). Representative methods are bias tuning, LoRA, and Compacter. 
    \end{minipage} 
& 
    \begin{minipage}{.52\textwidth}

        \textbf{Bias Part}:
         \vspace{2pt}      
        Bitfit \cite{zaken2022bitfit},
        Side Adapter \cite{xu2023side},
        AdapterBias \cite{fu2022adapterbias},
        DP-BiTFiT \cite{bu2022differentially}
        
       \textbf{Weight Part}: 
       \vspace{2pt} 
        LoRA \cite{hu2022lora},
        MoSA \cite{kothari2022motion},
        DyLoRA \cite{valipour2022dylora}
        DnA \cite{jiang2022dna},
        Compacter \cite{karimi2021compacter},
        KAdaptation \cite{he2022parameter},
        PHM \cite{zhangbeyond},
        PHNNs \cite{grassucci2022phnns},
        TARP \cite{hou2022meta},
        FacT \cite{jie2022fact},
        KronA \cite{edalati2022krona},
        DLDR \cite{li2022low},
        Aurora \cite{wang2023mode}
        
    \textbf{Weight and Bias}: 
        SSF \cite{lian2022scaling}
       \vspace{2pt}
    \end{minipage} 
    \\
\midrule
   \begin{minipage}{.24\textwidth}
      \includegraphics[width=\linewidth]{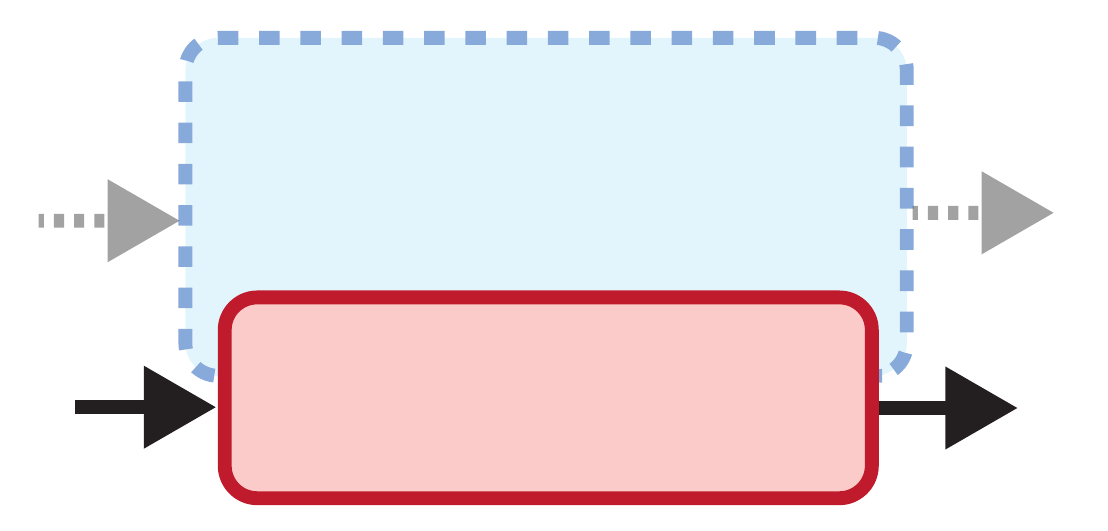}
      \centering 
      Remapping Tuning 
    \end{minipage}  
& 
    \begin{minipage}{.32\textwidth}
    Remapping-based tuning is a novel approach that involves transferring the learned knowledge of a pre-existing model to a new downstream model. This technique has shown promising results in improving the performance of downstream models and can be categorized into three different types according to the use of the pre-trained model.
    \end{minipage} 
&  
    \begin{minipage}{.52\textwidth}
    \textbf{Knowledge Distillation}: 
    \vspace{2pt}
        KD~\cite{hinton2015distilling},  Fitnet~\cite{romero2015fitnets}, Student~\cite{chen2020learning}, DFA~\cite{guan2020differentiable}, AdaIN~\cite{yang2020knowledge}, Normalized KD~\cite{xu2020feature},  Heterogeneous KD~\cite{passalis2020heterogeneous}, DeiT~\cite{touvron2021training}, Manifold KD~\cite{jia2022efficient}, Paraphrasing KD~\cite{kim2018paraphrasing}, RKD~\cite{park2019relational}, AKDNet~\cite{liu2020search}, SemCKD~\cite{chen2021cross},  HKD~\cite{zhou2021distilling},  Review~\cite{chen2021distilling}, DKD~\cite{zhao2022decoupled} 
    \hspace*{\fill}
    
    \textbf{Weight Remapping}: 
    \vspace{2pt}
        Net2Net~\cite{chen2016net2net}, EAS~\cite{cai2018efficient}, N2N Learning~\cite{ashok2018n2n}, NASH~\cite{elsken2018simple}, Path-level EAS~\cite{cai2018path}, FNA~\cite{fang2020fast}, FNA++~\cite{fang2020fna++} 

    \textbf{Architecture Remapping}: 
    \vspace{2pt}        
        DARTS~\cite{liu2019darts},
        DATA~\cite{DBLP:conf/nips/ChangZGMXP19},
        DATA-GS~\cite{DBLP:journals/pami/ZhangCGMXLP21},
        P-DARTS~\cite{chen2019progressive},
        DARTS+~\cite{liang2019darts+}, SGAS~\cite{li2020sgas}, SNAS~\cite{xie2019snas}, MiLeNAS~\cite{he2020milenas}, DARTS-~\cite{chu2021darts}
    \vspace{1pt}
\end{minipage} 

\\
\bottomrule

\end{tabular}%
}
\end{table*}

\subsection{Model Tuning}\label{subsection:background_finetuning} 

Given the knowledge learned via pre-trained models, downstream tasks can greatly benefit from them. 
Early modus operandis of fine-tuning includes updating the whole parameters of the pre-trained model and tuning the task head only (e.g., fully connected layer). With the popularity of large language models such as GPT-3 \cite{brown2020language} pre-trained via meta-learning in an unsupervised manner, enabling them to handle a broad set of skills (the inner loop termed “in-context learning”). %
Given the ability of multiple skills, the current leading paradigm in NLP is to adapt downstream tasks to the large language models, entering the learning paradigm of “pre-train, prompt, predict” from “pre-train, fine-tune” \cite{bommasani2021opportunities}.

On the one hand,  a couple of recent works \cite{zang2022unified, pan2022st} achieve promising performances on vision downstream tasks by fine-tuning visual-language models.
However, according to the results in \cite{pan2022st} and \cite{he2022parameter}, fine-tuning visual-language models do not lead to results as good as fine-tuning supervised pre-trained vision models. In addition, pure vision models are also increasingly large (reach 22B parameters) \cite{dehghani2023scaling} and gain great advances recently \cite{chen2020generative,li2021efficient,reed2022generalist} with varied pre-training strategies \cite{khan2022transformers, zhou2023comprehensive}. As such, it needs further investigation on proper pre-training techniques and fine-tuning techniques for vision downstream tasks.

\section{Visual Tuning}\label{sec:petl_techniques}
To the best of our knowledge, there is no survey that systematically summarizes the recent state of visual tuning from the technical perspective. He \textit{et al.} \cite{he2022towards} analyzed different PETL tuning methods such as prompt-tuning, prefix-tuning, and adapters in the NLP domain, showing they are intrinsically similar (i.e., they bring a certain amount of tunable parameters for adaptation). Taking parameter efficient transfer learning methods in NLP into consideration, we group visual tuning methods into five categories: fine-tuning, prompt tuning, adapter tuning, parameter tuning, and remapping tuning (see Table \ref{tab:efficient_finetune_methods}) according to their structures 
and motivations. In the remainder of this section, we introduce the five groups of tuning techniques with discussions of their advantages and disadvantages.

\subsection{Fine-tuning}\label{subsection:vt_finetuning}
We use fine-tuning to denote the standard practice of transfer learning, which either tunes the whole parameters of pre-trained models or just tunes the task head. Many state-of-the-art methods adopted this practice to achieve impressive performance on vision benchmarks such as ImageNet 
\cite{deng2009imagenet}, Kinetics \cite{kay2017kinetics}, COCO \cite{lin2014microsoft},  NTU RGB+D 120 \cite{liu2019ntu}, Human3.6M \cite{ionescu2013human3}, etc. Tuning the whole pre-trained model intrinsically initiates the learning process of the downstream tasks via the learned model weights. While tuning the task head treats the pre-trained model as a feature extractor.

The full fine-tuning strategy comes with obstacles for adapting large models to downstream tasks. First, it requires one to update and store separate model parameters for different downstream tasks, which can be expensive and infeasible when the foundation models become increasingly large.
Second, it relies on high-quality downstream data and can hardly adapt to unseen scenarios that have large distribution shift \cite{kumar2021fine}, which is unlike the learning process of humans who can learn from few samples and generalize well to new circumstances. This issue has been researched in directions such as zero-shot learning, few-shot learning, and continual learning \cite{li2021else}. Alternatively, fine-tuning the downstream task head can avoid updating the entire backbone model, but it usually leads to unsatisfactory experimental performance. 

\subsection{Prompt Tuning}\label{sec:prompt_tuning}

Prompt-based learning is first introduced in NLP to efficiently adapt downstream language tasks to foundation models. 
Unlike the traditional “pre-training, fine-tuning” paradigm which initializes the weight parameters of pre-trained model and optimizes these parameters under the guidance of downstream task-specific loss functions, prompt-based learning leverages textual prompts to reformulate various downstream tasks as the original pre-trained task. Inspired by prompt techniques in NLP, prompt tuning is also introduced into the computer vision field. Specifically, vision prompt tuning could be divided into three groups, \textit{i.e.}, vision-driven prompt, language-driven prompt, and vision-language prompt.

\subsubsection{Vision-driven Prompt}
\begin{figure}[t]
    \centering
    \includegraphics[width=0.6\linewidth]{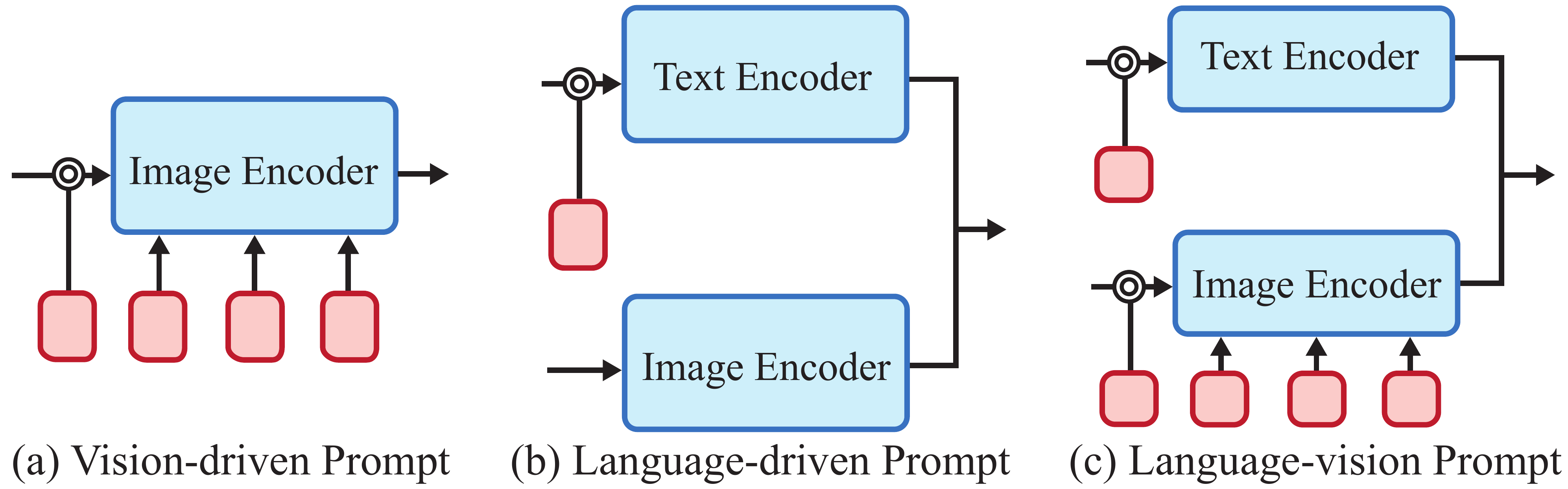}
    \caption{Three different types of prompt methods. The red and blue parts are tunable and frozen parameters, respectively.}
    \label{fig:prompt_detail}
\end{figure}

Vision-driven prompt tuning \cite{zhou2022zegclip,wu2022unleashing,hu2022prosfda,bowman2023carte,yu2023prompting,dong2022lpt,zhang2022neural,loedeman2022prompt} has become a popular parameter-efficient way to transfer the remarkable generalization ability of pre-trained vision models to various downstream tasks.
The research efforts of vision-driven prompt strategies can be roughly categorized into two groups, {\it i.e.}, modifying inputs directly, and designing vision prompt sub-networks to produce vision prompts.
Studies of the first family  \cite{jia2022visual, wang2022s, sohn2022visual, gao2022visual, dong2022autoencoders, herzig2022promptonomyvit, mao2022understanding} usually tend to directly modify inputs, {\it e.g.}, \textit{adding a set of learnable parameters into input images}, which aims to modify the input distribution and further makes downstream tasks close to the solved task during the original pre-training, as shown in Fig. \ref{fig:prompt_detail}(a).
Formally, the mathematical formulation can be described as:
\begin{equation}
    \mathcal{P}_V = [\mathcal{T},\textbf{\textit{a}}_1, \cdots, \textbf{\textit{a}}_i, \cdots, \textbf{\textit{a}}_n],
\end{equation}
where $\mathcal{P}_V$ indicates the vision-driven prompts, $\mathcal{T}$ denotes the embeddings of local images or tokens outputted by Transformer, and $\textbf{\textit{a}}_i$ is the $i$-th learnable vector.

Existing extensive works utilize the above principle to design vision prompts to instruct frozen vision pre-trained models to various downstream tasks. Concretely, VPT \cite{jia2022visual} plugs solely a few learnable parameters and regards these parameters as a part of input tokens of Transformer, which steers pre-trained vision models to perform various downstream tasks.
Similar with VPT, DePT \cite{gao2022visual} also introduces learnable visual prompts into the vision Transformer and only optimizes these source-initialized prompts while keeping the vision Transformer frozen during adaptation.
In addition, PViT \cite{herzig2022promptonomyvit} designs task-specific prompts by introducing a small set of specialized parameters to adopt a shared video transformer backbone to perform synthetic scene tasks and a real video downstream task. 
LPT \cite{dong2022lpt} optimizes shared prompts to explore the general features across the entire long-tailed dataset, and group-specific prompts to endow the fine-grained discrimination ability into frozen pre-trained vision models. 

As proven by the above works, prompt learning enables pre-trained visual models to adapt to a variety of visual tasks in natural scenarios.
However, prompt learning still has great potential in transferring visual knowledge of pre-trained vision models trained in natural scenarios to downstream tasks that have large domain gaps.
A recent study has extended vision prompt from natural scene understanding to diverse vision tasks with huge domain discrepancies, such as point cloud analysis \cite{zhang2022pointclip, wang2022p2p}, image generation \cite{wu2022generative} and even speech understanding \cite{kim2023prompt}.
Concretely, PointCLIP \cite{zhang2022pointclip} converts the raw points into scatter depth maps by projecting them onto predefined image planes, termed as vision prompt, which effectively transfers the remarkable ability of CLIP model. In addition, PointCLIP also narrows the modality discrepancies between unordered point clouds and the visual images, thus producing a unique insight for processing vision tasks with significant domain gaps using prompt technology.
P2P \cite{wang2022p2p} proposes the geometry-preserved projection and geometry-aware coloring operations to translate point cloud data into colorful images, which are regarded as vision prompt and further adapt the pre-trained vision model for various point cloud analysis tasks.
These works show that vision-driven prompts can transfer pre-trained vision models from natural scenarios to various downstream tasks even with domain discrepancies.

Excitingly, the above work takes only a simple manner (e.g., adding extra parameters into inputs) to construct visual prompts but makes great progress on transferring the remarkable discrimination and generalization of pre-trained vision models.
To further investigate and dig into the effectiveness of vision prompt, the other family of approaches \cite{zhang2022neural, loedeman2022prompt, zhang2022feature, wang2022fine, gao2023compositional, gan2022decorate, gu2021open} also tend to \textit{design a sub-network to construct vision prompts}, as shown in Fig. \ref{fig:prompt_detail}(c). 
Specifically, the vision-driven prompts $\mathcal{P}_V$ can be denoted as:
\begin{equation}
    \mathcal{P}_V = \Phi(X, \boldsymbol{\theta}),
\end{equation}
where $\Phi(,)$ denotes the designed sub-network to produce vision prompts $\mathcal{P}_V$, $\boldsymbol{\theta}$ is the learnable parameters in $\Phi(,)$, and $X$ is the input image.
For instance,
NOAH \cite{zhang2022neural} combines adapter, prompt, and LoRA. It utilizes the neural architecture search algorithm to learn the down-sampled dimension of adapters, the down-projection dimension of LoRA, and the learnable token length of prompts, leading to better parameter efficiency and performance trade-off. 
PGN \cite{loedeman2022prompt} learns to produce input-dependent prompts via selectively sampling input images from a commonly learned library of tokens.
FRPT \cite{wang2022fine} explicitly zooms the discriminative regions of input images via designing a lightweight sampling network to obtain the vision prompts. 
RePro \cite{gao2023compositional} localizes objects from videos as vision prompts utilizing a tracklet detector and further learns the correlation between subjects and objects according to the learned vision prompts.
ViLD \cite{gu2021open} generates multiple regions of interest based on the region proposal network, regarded as vision prompts, to align their visual embeddings and textual embeddings for open-vocabulary object detection.
These works can produce appropriate prompts according to downstream tasks, thus effectively exploring the remarkable generalization and discrimination ability of pre-trained vision models. 
More importantly, compared to introducing learnable parameters directly, they can improve the interpretability of these vision prompts such as directly modifying the pixels.

\subsubsection{Language-driven Prompt}
Recently, large-scale vision-language models are pre-trained by extensive image-text pairs and focus on open-world visual concepts. 
Following this ideology of prompt learning in NLP, most existing works tend to transfer large-scale vision-language models into various downstream vision tasks via designing appropriate language-driven prompts  \cite{ma2023understanding,berg2022prompt,zhang2022prompting,park2022lanit,peng2022sgva,tao2023galip,zhou2022conditional,zhu2022pointclip,lin2023comes}. 
As shown in Fig. \ref{fig:prompt_detail}(b), 
most works, such as CoOp \cite{zhou2022learning}, firstly extract unified context or class-specific context of visual images as language-driven prompts to adapt frozen pre-trained vision-language models to diverse vision tasks.
Formally, the language-driven prompts can be formulated as below:
\begin{equation}
    \mathcal{P}_T = [\textbf{\textit{a}}_1, \cdots, \textbf{\textit{a}}_i, \cdots, \textbf{\textit{a}}_n, <class>],
\end{equation}
where $\mathcal{P}_T$ denotes the language-driven prompts, $a_i$ represents the $i$-th learnable vectors, the number of learnable vectors is $n$, and $<class>$ is the class embeddings.
Extensive existing works have focused widely on this line of language-driven prompt and utilize this prompt analogous to the designed in CoOp to adapt various downstream tasks, {\it e.g.}, domain adaption \cite{chen2022multi}, semantic segmentation \cite{kim2023zegot}, video understanding \cite{ju2022prompting, ni2022expanding}, and few-shot learning \cite{zhu2022prompt}. 
In addition, recent methods \cite{saito2022prefix, bulat2022language, sun2022dualcoop, chen2022prompt, he2022cpl, wang2022learning} extend the original language-driven prompt and thus design multiple complementary language-driven prompts to  better mine the task-specific knowledge from pre-trained vision-language models.
The multiple complementary language-driven prompts $\mathcal{P}_{MT}$ can be represented as 
\begin{equation}
   \mathcal{P}_{MT} = [ \mathcal{P}_T^1, \cdots,  \mathcal{P}_T^i, \cdots,  \mathcal{P}_T^M].
\end{equation}
For instance, PLOT \cite{chen2022prompt} learns multiple comprehensive prompts to capture different attributes of classes and aligns visual embeddings and multiple textual embeddings via optimizing the optimal transport distances between multiple prompts.

\subsubsection{Vision-language Prompt}
Vision-driven and language-driven prompts have been explored to simultaneously modify the vision and text inputs for pre-trained vision-language models, thus transferring the discrimination and generalization ability of pre-trained vision-language models thanks to effectively  aligning visual and textual embeddings \cite{zang2022unified,xing2022class,khattak2022maple,shen2022multitask,zhao2022learning,shu2022test,DBLP:conf/cvpr/WangCLWO023}. 
For an instance, UPT \cite{zang2022unified} designs a shared prompt network to produce the vision prompt and text prompt, thus narrowing the gap between visual representations and textual embeddings.
DPT \cite{xing2022class} simultaneously optimizes the visual and textual prompts from the vision and text input perspectives, which aims to modify the textual classifier and visual representations of pre-trained vision-language models. 
TPT \cite{shu2022test} introduces learnable text prompts with random vectors and category names and designs vision prompts generated by cropping input images randomly.
These methods can transfer the pre-trained vision-language models to various downstream tasks from the perspective of text and vision inputs.

\subsubsection{Discussion}
It is well known that the quantity of labeled data largely determines the upper limit of the vision algorithm. Vision prompt learning usually focuses on solving the problem of few-shot or zero-shot learning, which allows the model to perform relatively well even without labeled data.
Moreover, visual prompt learning integrates all subsequent tasks into pre-training tasks by creating a distinct template. Through this approach, data from downstream tasks are converted into new inputs to leverage the inherent capacities of pre-trained models. 
In other words, the core mechanism of the vision prompts aims to harness the capabilities of the upstream pre-trained model, allowing it to excel in downstream tasks even with minimal reliance on annotated data.

However, prompt-based tuning also suffers from some limitations, vitally sacrificing the applicability in the real world. Firstly, a significant challenge facing prompt tuning is how to construct or highlight effective visual cues of inputs and seamlessly integrate them with downstream tasks. 
This necessitates a profound understanding and solid technical expertise in both the fields of original pre-training tasks and downstream tasks. 
Additionally, the prompt-based tuning approach still demands substantial computational resources for model training and optimization, inevitably leading to increased training time and costs. 
Lastly, despite prompt tuning showcasing notable performance improvements in many tasks, its generalizability requires further exploration and validation when facing huge domain differences between original pretraining tasks and downstream ones.

Despite these limitations, prompt tuning will assume an increasingly pivotal role in the realm of artificial intelligence.
We posit that exploring three specific avenues could mitigate some of the current limitations.
Firstly, prompt tuning tends to transparent and interpretable adjustments to the input prompts. This transparency enables researchers and practitioners to understand and validate the model's decision-making process, performing better against different data distributions or interferences.
Furthermore, researchers and users can tailor prompts to steer the model's attention towards specific features or classes of interest, making the model more usable for various applications and tasks.
Prompt tuning contributes to the usability of deep learning models by facilitating model interpretability and controllability. 
Lastly,  prompt tuning tends to promote consistency in model performance by enabling standardized methodologies for adjusting prompts across different datasets and tasks. This consistency ensures that models behave predictably and reliably in various scenarios, enhancing their overall usability and applicability.

\subsection{Adapter Tuning}
Adapter-based methods are a class of techniques that involves additional trainable parameters into a pre-trained model that has been frozen to facilitate learning for downstream tasks. In the NLP domain, adapters were first introduced by Houlsby \textit{et al.} \cite{houlsby2019parameter} as a means of achieving PETL. However, efficient adaptation, particularly in the field of computer vision, has received comparatively little attention.
Initial efforts to develop adaptive methods for computer vision have included incremental learning methods \cite{rosenfeld2018incremental} and domain adaptation methods \cite{rebuffi2017learning, rebuffi2018efficient}. Subsequently, adapters have garnered interest across domains and have been successfully applied in the computer vision field. Adapters provide a lightweight alternative to extensive model fine-tuning.

In this section, we have sorted out the existing vision-related adapter-based tuning methods, which can be roughly divided into three ideas, {\it i.e.}, sequential adapter, parallel adapter, and mix adapter one by one as follows.

\begin{figure}[t]
    \centering
    \includegraphics[width=0.6\linewidth]{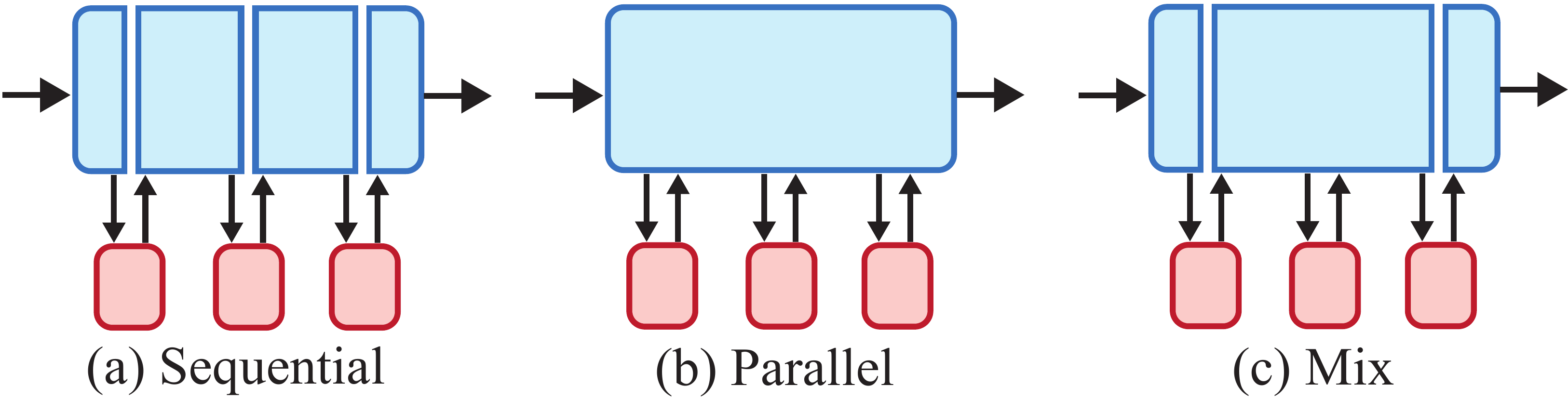}
    \caption{Three different types of adapter methods. Red and blue parts are tunable and frozen parameters, respectively.}
    \label{fig:adapter_detail}
\end{figure}

\subsubsection{Sequential Adapter}
Sequential adapter refers to the technique of inserting parameters into a sequential forward network shown in Fig. \ref{fig:adapter_detail}(a), which typically includes a linear down projection, a non-linear activation function, an up projection, and a residual connection. This approach is commonly applied after the multi-head attention layer and/or the feed-forward layer to enhance model performance. In particular, given a $d$-dimensional input feature map $Z ^{(l)}$, the number of parameters of adapter can be adjusted by a hyperparameter $d_{\text{bottle}}$ $(d_{\text{bottle}}\ll d)$. The sequential adapter module first uses a down-projection ({\it i.e.}, downsampling) with $\textbf{\textit{W}}_{\text{down}} \in \mathbb{R}^{d{\times}d_{\text{bottle}}}$ to project the feature to the lower-dimensional representation, followed by a ReLU activation function and an up-projection ({\it i.e.}, upsampling) with $\textbf{\textit{W}}_{\text{up}} \in \mathbb{R}^{d_{\text{bottle}}{\times}d}$. 
The above formulation can be written as:
\begin{equation}
  \hat{Z}^{(l)}=\text{ReLU}(\text{LN}(Z^{(l)})\textbf{\textit{W}}_{\text{down}})\textbf{\textit{W}}_{\text{up}},
  \label{eq:adapter_sequential}
\end{equation}
where $\hat{Z}^{(l)}$ denotes the optimized features outputted by the sequential adapter.

In sequential adapter strategies, research can be categorized into two groups: inserting residual blocks directly, and using parameter optimization techniques to minimize adapter size.
Studies of the first group \cite{rebuffi2017learning,rosenfeld2018incremental,sung2022lst,chen2022conv} emerged early without large-scale models. Res-adapt \cite{rebuffi2017learning} involves a customized deep network with adapter residual modules to adapt to different visual domains in real-time.
DAN \cite{rosenfeld2018incremental} converges to comparable or higher performance with a fraction (typically 13\%) of the parameters of standard fine-tuning after Res-adapt.
Recent work~\cite{sung2022lst} introduces LST, which trains a separate ladder network using intermediate activations and shortcut connections to improve accuracy and reduce computational complexity.
Additionally, Conv-Adapter\cite{chen2022conv} investigates feasible solutions to learn task-specific knowledge by adapting intermediate features of each residual block using four variants.

EPM \cite{rebuffi2018efficient}  suggests using universal parametric neural network families with limited parameters, while Polyhistor \cite{liu2022polyhistor} decomposes a hyper-network into separate hyper-networks and factorizes adapter weight matrices. Additionally, Pro-tuning enriches the feature space with multiple prompt blocks \cite{nie2022pro}, while AMixer captures long and short-term dependencies without self-attention \cite{rao2022amixer}. Shysheya et al. propose Fit \cite{shysheya2023fit}, which scales and shifts activations and uses a Naive Bayes final layer classifier for image classification. Marouf et al. introduce TINA \cite{marouf2023tiny}, which iteratively reduces adapter size using a scoring function compared to neuron importance, improving overall model efficiency. Finally, Luo et al. propose RepAdapter \cite{luo2023towards}, which uses re-parameterization of sparse structure to approach nearby projection weights, reducing model parameters while maintaining effectiveness and lightweight nature.
 
Adapters have become a popular technique for foundation tasks where the pre-training task is often image classification. However, other tasks such as high-level vision tasks \cite{yuan2021florence,li2021benchmarking,liu2021cross,li2022exploring,zhang2022collaboration,Ermis_2022_CVPR,luo2023survey}, low-level vision tasks \cite{wang2020stacking,tsubota2023universal}, video understanding \cite{yuan2021florence,yang2023aim,panst}, and robotic control \cite{sharma2023lossless} all require designs that are tailored to their specific architecture, in order to efficiently transfer learned parameters and achieve good performance through PETL.
In addition to these task differences, recent research has proposed innovative ways to utilize adapters in different applications. 
Recent research proposes innovative ways to use adapters, such as BDTL and ViTDet \cite{li2021benchmarking, li2022exploring} adjusting a plain backbone with minimal adaptation for object detection, and Florence \cite{yuan2021florence} incorporating universal visual-language representations for a wide range of tasks such as retrieval, classification, object detection, visual question answering, and action recognition. SND \cite{wang2020stacking} uses a dynamic stacked network for image restoration, MK-Adapter \cite{zhang2022collaboration} blends predictions for few-shot classification, and ADA \cite{Ermis_2022_CVPR} performs continual learning. AIM \cite{yang2023aim} and ST-Adapter \cite{panst} equip models with spatio-temporal reasoning for video understanding. PEA \cite{sharma2023lossless} addresses robotic manipulation limitations, and CAOA \cite{tsubota2023universal} optimizes image compression with adapters.

In the field of multi-modal learning, with the development of large-scale cross-modal pre-trained models, {\it i.e.}, CLIP~\cite{radford2021learning} and ALIGN~\cite{jia2021scaling}, adapter technique has been widely adopted, using a design analogous to the one mentioned above, to adapt various downstream tasks for efficient fine-tuning \cite{gao2021clip, zhang2021tip, ma2021simple,eichenberg2021magma,sung2022vl,zhang2022hyperpelt,jiang2022cross, lin2022vision,pantazis2022svl, zhang2023multimodal} with excellent results. 
HA \cite{kim2021adapt} recommends general recipes for efficient multi-modal transfer learning. CLIP-Adapter \cite{gao2021clip} uses residual-style feature blending with an additional bottleneck adapter, while Tip-Adapter \cite{zhang2021tip} enhances few-shot capability without backpropagation during training. MAGMA \cite{eichenberg2021magma} combines visual and textual inputs for generative language models, and BALLAD \cite{ma2021simple} augments representations for long-tailed vision language learning. Hierarchical3D \cite{papalampidi2022hierarchical3d} integrates multi-modal content into a textual summarizer, while VL-Adapter \cite{sung2022vl} adjusts pre-trained models with sequential adapter layers for cross-modal domains. HyperPELT \cite{zhang2022hyperpelt} fine-tunes small modules using a shared hyper-network, while CrossModal-Adapter and MV-Adapter \cite{jiang2022cross, zhang2023multimodal} allow early cross-modal interactions. SVL-Adapter \cite{pantazis2022svl} combines vision-language pre-training and self-supervised representation learning, and LAVISH \cite{lin2022vision} migrates adapters for pre-trained ViTs to audio-visual tasks. These approaches demonstrate the versatility of adapters and their potential for various applications beyond traditional classification tasks in multi-modal learning.

\subsubsection{Parallel Adapter}
Parallel adapter \cite{chen2022vision,rao2022parameter,chen2022adaptformer,jie2022convolutional,lu2023uniadapter,liu2022prompt,yu2023rethinking} has been proposed as a variant of the classic sequential adapter architecture shown in Fig. \ref{fig:adapter_detail}(b). Here, activations are passed via the module layer in parallel to the adapted sub-layer (i.e. feed-forward or attention layer), as opposed to the established, sequential, order of computations. The parallel adapter module also uses a down-projection ({\it i.e.}, downsampling) with $\textbf{\textit{W}}_{\text{down}} \in \mathbb{R}^{d{\times}d_{\text{bottle}}}$ to project the feature to the lower-dimensional representation, followed by a ReLU activation function, and an up-projection (a.k.a. ({\it i.e.}, upsampling)) with $\textbf{\textit{W}}_{\text{up}} \in \mathbb{R}^{d_{\text{bottle}}{\times}d}$ in parallel. 
Formally, the process of parallel adapter can be described:
\begin{equation}
  \hat{Z}^{(l)}=\text{ReLU}(\text{LN}(Z^{(l)})\textbf{\textit{W}}_{\text{down}})\textbf{\textit{W}}_{\text{up}}+\text{LN}(Z^{(l)}),
  \label{eq:adapter_z_wave}
\end{equation}
where $\hat{Z}^{(l)}$ denotes the optimized features outputted by the parallel adapter.

The simplest application of adapters is to insert a module in parallel. ViT-Adapter \cite{chen2022vision} introduces image-related biases by a pre-training-free adapter, while PESF-KD \cite{rao2022parameter} updates only the adapter for soft labels. AdaptMLP \cite{chen2022adaptformer} adapts to large video action recognition using two parallel branches. Convpass \cite{jie2022convolutional} uses trainable convolutional blocks to improve inductive bias. AMA \cite{yu2023rethinking} restores 2D structure for each modality, and UniAdapter \cite{lu2023uniadapter} unifies uni-modal and multi-modal adapters with partial weight sharing. These approaches demonstrate the versatility of adapter modules in various applications.

\subsubsection{Mix Adapter}
Mix adapter \cite{yu2022towards,wu2022pruning,tu2022visual,shimomoto2022towards,hao2023consolidator,xu2023exploring} introduces new parameters in different positions with mixed architecture demonstrated in Fig. \ref{fig:adapter_detail}(c), {\it i.e.}, the multi-head attention blocks in each Transformer layer.

PATT \cite{yu2022towards} explores efficient parameter techniques for video-based downstream tasks with a prefix-tuning module. ETT \cite{xu2023exploring} uses attentive prefix tuning and domain residual adapters for few-shot learning. PALT \cite{wu2022pruning} prunes adapters based on the lottery ticket hypothesis. VQT \cite{tu2022visual} aggregates intermediate features for parameter and memory-efficient transfer learning. Consolidator \cite{hao2023consolidator} structures tunable parts for efficient transfer learning with group-wise convolution. TVG \cite{shimomoto2022towards} compares pre-trained models and adapters for video grounding tasks. These approaches demonstrate the versatility of efficient adapter techniques in various applications.

\subsubsection{Discussion}

Adapter-based methods represent a popular PETL approach within vision and multi-modal learning, emphasizing the modification of a small set of parameters within a frozen backbone to address downstream tasks. 
This not only economizes on computational expense but also introduces a high degree of modularity. Such modularity facilitates the swift adaptation of pre-trained models to new tasks without necessitating significant architectural overhauls. Meanwhile, by focusing adaptation efforts on a concise set of parameters, adapter-based techniques maintain the integrity of the original model's learned representations, thereby enhancing the model's generalization capabilities across various tasks. 
Moreover, Adapters introduce variability through methods like projecting down and up with intermediate non-linear layers, offering a range of model adjustments not typically available through direct parameter tuning.

However, adapter tuning has its limitations when compared with other methods. On one hand, adapter tuning lacks of interpretable semantic meaning compared with prompt tuning. On the other hand, it can be slightly less parameter efficient than parameter tuning such as LoRA. 
Regarding the comparison with remapping methods, adapter tuning is faced with the challenge of where to insert parameters (such as Transformer models' attention and feed-forward modules, between the Transformer layers or blocks, etc.). Existing adapter tuning methods seem to have no consistent rule but just insert parameters to specific layers. 

We posit that exploring two specific avenues could mitigate some of the current limitations. Firstly, introducing more efficient operations could broaden the applicability of adapter-based methods across various communities. Not all layers of a foundation model may require adapters; a unified rule, akin to the scaling principles used in foundation models, could dictate their strategic implementation, enhancing efficiency.
Secondly, more adapter architecture can be studied. For instance, in the realm of NLP, there exist adapter architectures that exhibit promising performance in adapting to new tasks \cite{pfeiffer2020adapterhub}, which can be leveraged and applied to visual tuning.
Furthermore, emerging integration techniques will likely enable adapters to achieve improved performance in practical applications.

\subsection{Parameter Tuning}
\begin{figure}[t]
    \centering
    \includegraphics[width=0.6\linewidth]{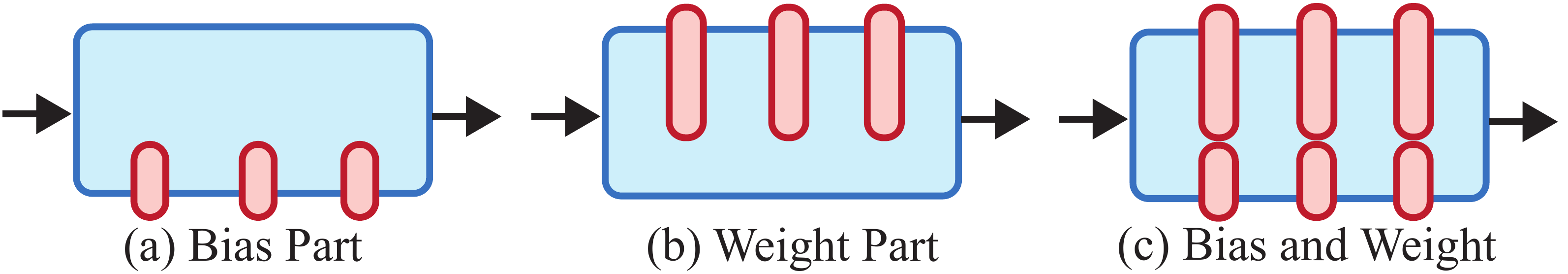}
    \caption{Three types of parameter tuning. Red and blue parts are tunable and frozen parameters, respectively.}
    \label{fig:variable-based}
\end{figure}

An effective parameter-based tuning involves directly modifying the parameters (either weights or biases) of the pre-trained model in a more aggressive manner. Given a specific layer, it can have its weight-term multiplied to the feature map and a bias-term added to the feature map. As shown in Fig. \ref{fig:variable-based}, this section introduces parameter-based methods based on which part of the parameters are tuned: weight part, bias part, and both. Techniques can be grouped into addition and decomposition. Existing works also termed this technique as reparameterization-based methods \cite{ding2023parameter, luo2023towards}.

Given a neural network layer with parameters {($k,d,K,G$), where $k=C_{\text{in}}$ is the input channel, $d=C_{\text{out}}$ is the output channel, $K$ is the kernel size, and $G$ is the group size. When $G=1$, we will have $\textbf{\textit{W}} \in \mathbb{R}^{d \times k \times K}$ and $\textbf{\textit{b}} \in \mathbb{R}^{d}$. Then a typical neural convolutional operation can be denoted as:
\begin{equation}
 Z^{(l)}=Z^{(l-1)}\textbf{\textit{W}} + \textbf{\textit{b}},
  \label{eq:lora_1}
\end{equation}
where $Z^{(l)}$ and $Z^{(l-1)}$ denote the output and input features at the $l$-th neural network layer.
The group parameter $G$ can be used to control the connections between inputs and outputs, leading to the weight becoming $\textbf{\textit{W}} \in \mathbb{R}^{d \times \frac{k}{G} \times K}$. For ease of explanation, we do not consider the kernel size and feature size but focus on the variable size.  
In the remainder of this section, parameter tuning methods are introduced based on three groups: bias part, weight part, and both.

\subsubsection{Bias Part} 

Bitfit \cite{zaken2022bitfit} is also known as side-tuning, which only tunes the bias part of the pre-trained model (see Fig. \ref{fig:variable-based}[a]) and can be represented as:
\begin{equation}
 Z^{(l)}=Z^{(l-1)}\textbf{\textit{W}} + \textbf{\textit{b}},
  \label{eq:lora_2}
\end{equation}
where the weight parameters $\textbf{\textit{W}}$ are frozen, and the bias $\textbf{\textit{b}}$ contains the parameters optimized in the tuning process.
Avoiding change in the bias of the pre-trained model, AdapterBias \cite{fu2022adapterbias} targets the bias term at the MLP layer by using a linear layer $L$ with weight ($\boldsymbol{\alpha} \in \mathbb{R}^{d}$) and a tunable vector $\textbf{\textit{v}} \in  \mathbb{R}^{r}$, which can be calculated as:
\begin{equation}
 Z^{(l)}=Z^{(l-1)}\textbf{\textit{W}} + \textbf{\textit{b}}+\textbf{\textit{v}} \otimes \boldsymbol{\alpha}.
  \label{eq:adapter_bias}
\end{equation}

Xu \textit{et al.} \cite{xu2023side} introduced side-tuning as two branches: one
for predicting mask proposals, and the other for predicting attention bias which is applied in the CLIP model for semantic segmentation. It adds a bias term to the results of the Softmax layer of the attention module. 
Differentially Private Bias-Term Fine-Tuning (DP-BiTFiT) \cite{bu2022differentially} proposed a differentially private version of bias-tuning. DP-BiTFiT  used the optimizer DP-SGD to make the bias term private: first aggregate bias gradient norms across all layers, then use it to compute clipping factor, add Gaussian noise to the sum of clipped gradients, and descend on bias term. DP-BiTFiT basically changed the way for optimizing the bias term, which 
achieves comparable performance with bias-tuning. DP-BiTFiT's implementation is worth noting as it does not calculate the gradients for the pre-trained weights, which helps to save over $60\%$ training time. 

Namazifar \textit{et al.} \cite{namazifar2023role} studied the role bias-term of Transformer for NLP tasks. From the mathematical perspective with empirical verification, it concludes that the bias term of the key linear transformation is redundant and can be omitted without
any impact on the attention module. Moreover, 
the bias term of the value linear transformation has a more
prominent role than that of the bias term of the query linear
transformation. 

\subsubsection{Weight Part}
Fig. \ref{fig:variable-based}(b) shows models that tune the weight part of some layers.
Given the parameter of a neural network layer with weight $\textbf{\textit{W}} \in \mathbb{R}^{d \times k}$, LoRA \cite{hu2022lora} learns parameters $\textbf{\textit{W}}_{\text{down}} \in \mathbb{R}^{d \times r}$ and $\textbf{\textit{W}}_{\text{up}} \in \mathbb{R}^{r \times k}$ on top of $\textbf{\textit{W}}$, which can be denoted as:
\begin{equation}
 Z^{(l)}=Z^{(l-1)}+Z^{(l-1)}(\textbf{\textit{W}}+\textbf{\textit{W}}_{\text{down}}\textbf{\textit{W}}_{\text{up}}).
  \label{eq:lora_3}
\end{equation}

The LoRA structure has been applied to an encoder-decoder model called motion style adapters (MoSA) \cite{kothari2022motion}. MoSA uses a lightweight LoRA structure for adapting the motion style (e.g., pedestrians) from a source domain with sufficient labeled data to a target domain (e.g., cyclists).
DyLoRA \cite{valipour2022dylora} proposes to truncate the parameters of rank to multiple parts ({\it i.e.}, ranks) and optimize them separately sequentially without relying on a search mechanism. 

Decomposition-and-Alignment (DnA) \cite{jiang2022dna} uses  GreBsmo (replaced with SVD in implementation) to decompose the weight matrix $\textbf{\textit{W}} \in \mathbb{R}^{d \times k}$ to a low-rank form: $\textbf{\textit{W}}=\textbf{\textit{UV}}+\textbf{\textit{S}},~\textbf{\textit{U}}\in \mathbb{R}^{d \times r}$ is the “alignable” part, $\textbf{\textit{V}} \in  \mathbb{R}^{r \times k}$ is the “fixed support” from the pre-trained model, $\textbf{\textit{S}} \in  \mathbb{R}^{d \times k}$ is the residual term. Two additional variables $\Delta \textbf{\textit{U}}$ and $\Delta \textbf{\textit{S}}$ are added to the “alignable” of the decomposed $\textbf{\textit{W}}$, which can be denoted as:
\begin{equation}
 Z^{(l)}=Z^{(l-1)}((\textbf{\textit{U}}+\Delta \textbf{\textit{U}})\textbf{\textit{V}}+\textbf{\textit{S}}+\Delta \textbf{\textit{S}}).
  \label{eq:lora_4}
\end{equation}
DnA remains needs to use SVD to implement the GreBsmo algorithm, bringing additional complexity to the iterative 
 optimization process.

Compacter \cite{karimi2021compacter}, KAdaptation \cite{he2022parameter}, and Aurora \cite{wang2023mode} use Kronecker products to decompose weight parameter to a  $\textbf{\textit{W}}=\sum_{i=1}^{n} \textbf{\textit{A}}_i \otimes \textbf{\textit{B}}_i, ~\textbf{\textit{A}}_i \in  \mathbb{R}^{n \times n},~ \textbf{\textit{B}}_i \in  \mathbb{R}^{\frac{k}{n} \times \frac{d}{n}}$ and tune one part of the decomposed term $\textbf{\textit{B}}_i$ with a low rank formed parameters $\textbf{\textit{B}}_i=\textbf{\textit{u}}_i\textbf{\textit{v}}_i, ~\textbf{\textit{u}}_i \in  \mathbb{R}^{\frac{k}{n} \times r},~ \textbf{\textit{v}}_i \in  \mathbb{R}^{r \times \frac{d}{n}}$, wihch can be represented as:
\begin{equation}
 Z^{(l)}=Z^{(l-1)}(\sum_{i=1}^{n} \textbf{\textit{A}}_i \otimes \textbf{\textit{u}}_i\textbf{\textit{v}}_i).
  \label{eq:lora_5}
\end{equation}
The decomposition method using the Kronecker product is also named as a parameterized hypercomplex multiplication/convolutional (PHM/PHC) layer \cite{zhangbeyond,grassucci2022phnns}, being applied for varied tasks such as vision and audio tasks. PHM \cite{zhangbeyond} inspires \cite{amplayo2022attribute} to form a tunable weight with three terms $\textbf{\textit{z}}_i,\textbf{\textit{s}}_i$, and $\textbf{\textit{A}}_i$, being added to the pre-trained weight for PETL of NLP tasks. 
FacT \cite{jie2022fact} considers two decomposition methods Fact-TT and Fact-TK, using Kronecker product and a multilinear generalization of the SVD (i.e., the Trucker model) \cite{de2000multilinear}, respectively. Fact-TK generally performs better than Fact-TT with slightly more parameters than Fact-TT across 19 image-based tasks, which is far fewer parameters than the basic LoRA method.
Dynamic Linear Dimensionality Reduction (DLDR) 
 \cite{li2022low} claims that only optimizing the low-dimensional subspace of a large model can achieve comparable performance. DLDR used SVD to decompose the weight to find the tuned subspace, which achieves comparable performance by training a small number of epochs. 
 
RepAdapter \cite{luo2023towards} is built based on LoRA structure and introduced a group-wise transformation \cite{luo2022towards} method to reparameterize the weight term. RepAdapter also interpreted its group-wise divided LoRA layers as a reparameterization process. RepAdapter \cite{luo2023towards} aims to reduce the inference time and seamlessly integrate the RepAdapter into most
giant vision models via structural re-parameterization. 

Similarly in NLP, task-adaptive reparameterization (TARP) \cite{hou2022meta} uses the Kronecker product as a dynamic low-rank decomposition for the MLP module for domain adaptation. Kronecker Adapter  (KronA) \cite{edalati2022krona} also introduces the Kronecker product to improve the limited representation power low-rank representation for NLP tasks. 

\subsubsection{Weight and Bias}

As illustrated in Fig. \ref{fig:variable-based}(c), some methods modify parameters of both weight and bias parts. Scale and Shift the deep Features (SSF) \cite{lian2022scaling} works towards the weights and bias terms by using two vectors $\boldsymbol{\gamma} \in \mathbb{R}^d$ and $\boldsymbol{\beta} \in \mathbb{R}^d$, which can be represented as: 
\begin{equation}
 Z^{(l)}=\boldsymbol{\gamma}(\textbf{\textit{W}}Z^{(l-1)}+\textbf{\textit{b}}) +\boldsymbol{\beta},
  \label{eq:ssf}
\end{equation}
where are respectively interpreted as scale and shift factors.
Note that both $\boldsymbol{\gamma}$ and $\boldsymbol{\beta}$ are learnable vectors, which can be much smaller than matrix variables of LoRA or decomposed forms of DnA, Compacter, and KAdaptation.

\subsubsection{Discussion}

Differences from the prompt-based and adapter-based methods, parameter-based tuning can use fewer parameters to achieve a similar effect of adaptation. On the tested image-based tasks \cite{lian2022scaling} can even outperform adapter and VPT. SSF \cite{lian2022scaling} introduces the bias-tuning technique to the weight variable via dot product. According to the analysis of SSF, it intrinsically modifies both the weight and bias variables, which is interpreted as follows: 
\begin{equation}
 Z^{(l)}=\boldsymbol{\gamma}(\textbf{\textit{W}}Z^{(l-1)}+\textbf{\textit{b}}) +\boldsymbol{\gamma} = (\boldsymbol{\gamma} \odot \textbf{\textit{W}})Z^{(l-1)} + \boldsymbol{\gamma} \odot \textbf{\textit{b}} + \boldsymbol{\gamma},
  \label{eq:lora_6}
\end{equation}
where $\odot$ is dot product. 
Given the varied techniques available, Mao \textit{et al.} \cite{mao2021unipelt} unified these methods with a gate mechanism. \cite{jiangback} uses pruning techniques to drop the activations during back-propagation, leading to sparse activations. This track of techniques will be further introduced in Section \ref{sec:remapping}.
In addition to Transformer-based structures, LoRA Winograd convolution \cite{qin2023low} aims to use the LoRA mechanism to prune the 3D CNN backbone model (e.g., C3D and R3D-18) for accelerating the  Winograd operation \cite{lavin2016fast} with less trainable parameters. 

Although parameter-based tuning can be less expensive from the perspective of tuned parameters, sometimes they will underperform the former two methods (i.e., prompt tuning and adapter tuning). This might be because fewer parameters can reduce the adaptation ability to a target domain with a large domain gap. Another limitation of existing parameter-based tuning methods is the lack of exploring based on the semantics of pre-trained models, leading to insufficient explainability. By far, most methods are tested on Transformer-based structures, but there remains exploration of their effect on CNN-based structures.

In the future, there will be continual exploration along this track of technique for more progressive parameter efficiency via further factorization on pre-trained models' weight or bias terms. Meanwhile, visual semantics are expected to be considered based on different types of pre-trained models (i.e., foundation models pre-trained with varied levels of vision tasks: low-level, middle-level, and high-level). It can also be combined with other tuning techniques for better interpretability. In addition, existing methods can also be expanded to CNN-based methods that have superior performance over corresponding Transformer-based methods on some specific tasks.

\begin{figure}[t]
    \centering
    \includegraphics[width=0.6\linewidth]{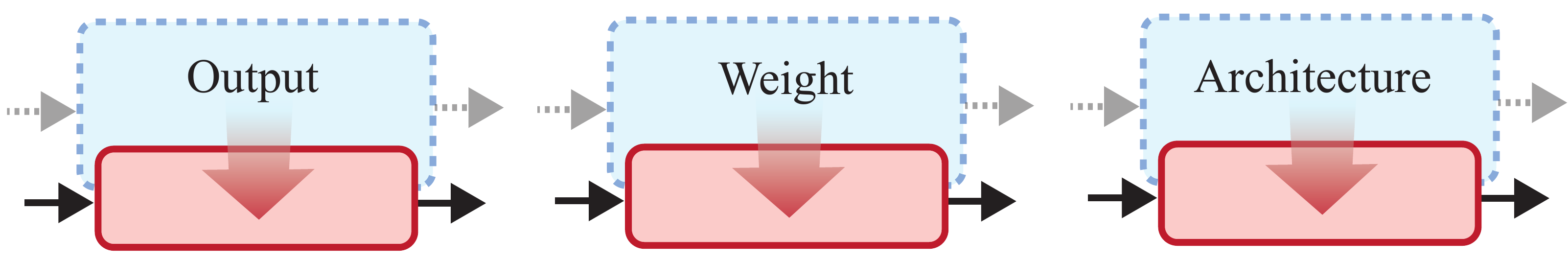}
    \caption{Three different types of remapping tuning methods. Red and blue parts are tunable and frozen parameters, respectively.}
    \label{fig:remap}
\end{figure}

\subsection{Remapping Tuning}
\label{sec:remapping}

Instead of directly fine-tuning or processing the pre-existing model, remapping-based tuning is a category of techniques that transfer the knowledge learned by a pre-trained model to a new downstream model. Based on how to utilize the pre-trained model, \textit{i.e.}, output, weight, and network architecture (see Fig. \ref{fig:remap}), we discuss three forms of knowledge transfer in the following categories: knowledge distillation-based remapping, weight-based remapping, and architecture-based remapping.

\subsubsection{Knowledge Distillation}

Knowledge distillation aims to regularize the downstream model by enforcing it to mimic the output of pre-trained models. Note that, the output typically refers to the final response or intermediate features. In the meantime, knowledge distillation is also an important model compression technique. In this section, we do not involve other model compression techniques such as network pruning~\cite{gong2014compressing, wen2016learning, ma2022dhwp}, since they are not typically motivated to transfer knowledge from teacher network to student network. 

The fundamental idea of knowledge distillation is to transfer the learned knowledge from a large pre-trained teacher model into a small student model by learning the network output or intermediate features of the teacher.
Typically, knowledge is distilled from the teacher model to the student model using a soft target distribution for each case. The probability $q_i$ of each case can be formulated as:
\begin{align}
    q_i = \frac{exp(z_i/T)}{\sum_j exp(z_j/T)},
\end{align}
where $\textbf{\textit{z}}$ is the output logit of the teacher networks and $T$ is the temperature of the distillation process.

To our best knowledge, the work of ~\cite{buciluǎ2006model} first introduces knowledge distillation to extract knowledge from a pre-existing model. They trained a compressed model with the pseudo data produced by an ensemble of shallow networks while no significant loss occurs in performance. This idea has been expanded to compress the deep and wide networks into shallower ones in~\cite{ba2014deep}. Hinton \textit{et al.}~\cite{hinton2015distilling} introduced the teacher-student knowledge distillation framework, where the student network is penalized based on the softened class distribution output of the teacher network.

One of the characteristics of deep neural networks is to obtain increasingly expressive power by learning hierarchical feature representations, as pointed out in~\cite{bengio2013representation}. Based on this theory, both the final response and the intermediate feature maps of the teacher network can be employed as the target for training the student model. To substantially exploit the information of intermediate layers, Fitnets~\cite{romero2015fitnets} introduces intermediate-level hints of the teacher to facilitate training the student. It enforces the intermediate feature alignment between the teacher and student networks via the teacher’s intermediate feature maps as hints. Subsequently, a rich line of work is devoted to aligning the features indirectly~\cite{kim2018paraphrasing, chen2020learning, guan2020differentiable, yang2020knowledge, xu2020feature, passalis2020heterogeneous, touvron2021training, jia2022efficient}. Concretely, Kim \textit{et al.}~\cite{kim2018paraphrasing} developed a factor transfer method that employs paraphrased intermediate features of the teacher as a factor, rendering the knowledge of the teacher network more understandable for the student network. Inspired by neural architecture search~(NAS)~\cite{liu2019darts}, Guan \textit{et al.}~\cite{guan2020differentiable} developed a two-stage distillation approach that adopts the differentiable search strategy to simultaneously improve the efficiency and the effectiveness of knowledge distillation. Xu \textit{et al.}~\cite{xu2020feature} developed a feature-normalized distillation method by introducing a sample-specific correction factor for the replacement of the temperature, with the goal of suppressing the impact of noise resulting from the one-hot label. Passalis \textit{et al.}~\cite{passalis2020heterogeneous} modeled the information flow of the teacher's multiple intermediate layers and then trained a student model to match this information flow. To realize knowledge transfer for vision transformers, Touvron \textit{et al.}~\cite{touvron2021training} introduced a token-based distillation strategy termed DeiT, which enforces the student transformer to directly reproduce the label estimated by the pre-trained teacher network using a distillation token. Hao \textit{et al.}~\cite{jia2022efficient} introduced a manifold distillation approach for vision transformers by substantially utilizing patch-level information. 

There are also some extensions to further explore knowledge transfer patterns. Park \textit{et al.}~\cite{park2019relational} proposed a relational knowledge distillation scheme for mutual relations transfer of outputs instead of individual outputs. As a generalization of vanilla knowledge distillation, they introduced distance-wise and angle-wise distillation losses to sufficiently extract the structural relations in data examples. Liu \textit{et al.}~\cite{liu2020search} proposed an architecture-aware knowledge distillation approach termed AKD, with the goal of finding the optimal student networks for distilling a given teacher network. Chen \textit{et al.}~\cite{chen2021cross} studied the semantics of intermediate layers and employed an attention mechanism to automatically assign the soft layer association between teacher and student networks, which can reduce the impact of over-regularization
during the training process. Zhou \textit{et al.}~\cite{zhou2021distilling} proposed a holistic knowledge distillation with graph neural networks, where the holistic knowledge contains individual knowledge and relational knowledge~\cite{liu2019knowledge, peng2019correlation}. To integrate two knowledge and refine their correlations, graph neural networks are adopted to learn holistic knowledge to provide supervision for the student network by aggregating feature representation from correlated data examples. Chen \textit{et al.}~\cite{chen2021distilling} proposed a residual distillation framework termed Review to effectively learn informative features from multi-level information in the teacher network. Review utilizes multiple layers in the teacher to guide the training for one layer in the student with great performance gains. Zhao \textit{et al.}~\cite{zhao2022decoupled} modeled the traditional knowledge distillation loss into target class knowledge distillation and non-target class knowledge distillation, then dived into their effects. Based on the observations, they found that the traditional knowledge distillation loss is a highly entangled formulation and introduced a decoupled method to facilitate the knowledge distillation.

For application, most of the above methods focus on image classification. Furthermore, knowledge distillation also demonstrates promising results in more vision tasks, such as object detection~\cite{li2017mimicking, guo2021distilling, zheng2022localization, zhang2021improve}, image segmentation~\cite{liu2019structured, liu2020structured, wang2020intra, yang2022cross}, person re-identification~\cite{porrello2020robust, remigereau2022knowledge}, super-resolution~\cite{zhang2021data, angarano2022generative}, depth estimation~\cite{hu2021boosting, wang2021knowledge}, and crowd counting~\cite{liu2020efficient}.

\subsubsection{Weight Remapping}

Rather than relying on the teacher's output as supervision to train the student, weight remapping directly transfers the model weights from the teacher network to the student ones. Specifically, assume that a teacher network is a function $f(x; \boldsymbol{\theta})$ parameterized by $\boldsymbol{\theta}$, where $x$ is the network's input. Weight remapping for student network $g$ is to reassemble a new set of parameters $\boldsymbol{\theta}'$ from the existing parameters of $\boldsymbol{\theta}$, such that:
\begin{align}
    \forall x, f(x; \boldsymbol{\theta}) = g(x; \boldsymbol{\theta}').
\end{align}
Net2Net~\cite{chen2016net2net} is a pioneering effort that rapidly transfers the knowledge stored in a pre-existing network into another network by remapping the weight of a pre-existing teacher network to the student. 
Subsequently, EAS~\cite{cai2018efficient} introduces the concept of weight remapping into neural architecture search by exploring the search space according to a pre-existing network and reusing its weights.
To tackle variable-length architecture and consider the entire input architecture, EAS employs a bidirectional recurrent network~\cite{schuster1997bidirectional} as the encoder network. In this way, the previously trained network can be further exploited to efficiently explore the architecture space and greatly accelerate the training process of the new network. To efficiently compress the teacher network for knowledge transfer, Ashoket \textit{et al.}~\cite{ashok2018n2n} proposed a reinforcement learning-based approach termed N2N learning, which models the conversion from the teacher network into a student network as a Markov Decision Process (MDP). N2N learning formulates the process of knowledge transfer as a two-stage action selection. In the first stage, a recurrent policy network selects a sequence of actions including keeping or removing layers of the large teacher network. In the second stage, another policy network performs further reduction in each remaining layer to meet the attenuate configuration.

Furthermore, some interesting weight remapping methods take the network path topology into consideration instead of merely adding or removing network layers. Elsken \textit{et al.}~\cite{elsken2018simple} introduced a hill climbing-based approach named NASH, which can automatically search the optimal student architecture. By using a series of alternative network morphisms, NASH can train the child networks with a short optimization process by cosine annealing. At each training step, NASH searches for the optimal architectures by a simple hill-climbing strategy~\cite{russell2010artificial}. Path-level EAS~\cite{cai2018path} enforces the meta-controller to change the topology of network connection paths while using function-preserving transformation operations to remap weights. 
To achieve this, Path-level EAS develops a bidirectional tree-structured meta-controller based on reinforcement learning, in order to enrich the architecture space to the generalization of multi-branch structures. 
And Yang \textit{et al.}~\cite{yang2022deep} proposed to customize networks efficiently via reassembling various pre-trained network blocks subject to downstream constraints. 

Previous object detection and semantic segmentation approaches use the network weights pre-trained on image classification for performance gains. However, one of the major challenges is that ImageNet pre-training typically requires highly large computation costs. To address this issue, Fang \textit{et al.}~\cite{fang2020fast} introduced a fast neural network adaptation approach dubbed FNA, making a pre-trained network adapt to a new task by modifying the network such as depth and kernels. In this way, FNA can expand NAS techniques to object detection and semantic segmentation with negligible computation costs. 
Technically, FNA first designs a seed network by selecting a manually designed network pre-trained on ImageNet such as~\cite{sandler2018mobilenetv2} and then enlarges it to a super network. By applying the weight remapping technique, the seed network is used to assign new model parameters.  
The follow-up work FNA++~\cite{fang2020fna++} extends the weight remapping of FNA to one more task (\textit{i.e.}, human pose estimation) and more network architectures including ResNet~\cite{he2016deep} and NAS networks with diverse widths, depths, and kernel sizes.

\subsubsection{Architecture Remapping}

Architecture remapping refers to the knowledge transfer about network architecture from a pre-existing model. To our best knowledge, this line of work is mainly used in weight-sharing neural network search~(NAS). Formally, $\mathcal{F}$ denotes the architecture and $\boldsymbol{\omega}$ denotes the weight of $\mathcal{F}$. The goal of NAS is to find the optimal architecture $\mathcal{F}^*$ that produces the best performance on the test set:
\begin{align}
    \mathcal{F}^* = \arg \max_{\mathcal{F}} \text{Eval}(\{\mathcal{F}, \boldsymbol{\omega}\};\mathcal{D}_\text{test}). 
\end{align}
Specifically, this type of NAS formulates the search space into an over-parameterized super-network, \textit{e.g.}, modeling the search space as multiple repeatable cells~\cite{pham2018efficient, liu2019darts, chu2021fairnas}. When transferring the searched architecture to downstream tasks, direct architecture transfer, which stacks several searched cells to form a downstream model and then retrains it on the downstream data, is the current mainstream scheme. Canonical examples include DARTS~\cite{liu2019darts} and its variants ~\cite{xu2020pc, chen2019progressive, li2020sgas, xie2019snas, he2020milenas, chu2021darts}. Direct architecture transfer has shown impressive results on downstream tasks.  

\subsubsection{Discussion}

Different from traditional transfer learning approaches, remapping tuning focuses on training a new downstream model isolated from the pre-existing model. Thus remapping tuning methods own their exclusive advantages. In this line of work, knowledge distillation involves training a smaller student model to mimic the output or intermediate features of a teacher model. This method is advantageous for efficient model compression as well as flexible student architecture designs. Weight remapping directly transfers model weights from a teacher network to a student network. This approach is beneficial for speed and efficiency, as transferring weights can be faster than retraining a new model from scratch. Architecture remapping focuses on transferring network architecture knowledge, often used in weight-sharing neural network search. This method enables the transferability of architectures discovered in one task to other tasks, accelerating the development of new models for various applications.

While remapping tuning offers many advantages, there are also some limitations for different approaches. For instance, knowledge distillation may lead to a loss of information from the teacher model. It can also be sensitive to hyperparameters, such as temperature and weighting of different losses. Weight remapping is a simple and effective solution without manually adding constraints. However, this type of work typically struggles to obtain a lightweight student network compared with knowledge distillation. Architecture remapping faces challenges in designing an effective search space for NAS, which can be complex and time-consuming. Additionally, NAS methods often require significant computational resources to explore and evaluate a large number of candidate architectures, increasing the overall computational cost.

Their advantages and challenges also provide valuable insights for future advancements. For knowledge distillation, beyond learning from the pre-trained model's output, its high flexibility suggests the potential to incorporate grounded information from the downstream tasks for distillation. For example, combining the pre-trained model's output with a physics simulation could help guide the knowledge distillation process, resulting in an accurate and efficient student model for the downstream tasks~\cite{lin2023comes}. Regarding weight remapping, a potential improvement is to combine with knowledge distillation to reduce the model size. As for architecture remapping, exploring stable and reusable modules from multiple pre-existing models could largely reduce the complexity of search space design, e.g., earlier layers of CNN are often reused for extracting lower-level visual features. This would help to flexibly and efficiently integrate various domain-specific models, based on the semantic associations among different domains.

\section{Visual Tuning Future}\label{sec:visual_future}
To date, the state of visual intelligence forms a transfer learning paradigm of pre-training and tuning, showing great promising performance on numerous benchmarks. Vision contributes a large portion of knowledge acquisition of human intelligence. However, due to the high dimensionality of vision data, the intelligence of machine vision suffers from a relatively small data scale in comparison to that of NLP and remains far behind the general human vision. The future promise of intelligent vision will be expanded beyond the competed benchmark datasets, realizing transformative impacts on more domains via a multidisciplinary coevolutionary process. On one hand, we expect that future pre-training techniques play the role of knowledge acquisition and storage in a “collection-labeling-training-feedback” cycle system. While future tuning is around how to make use of the learned knowledge through more diversified interactions beyond the prompts around conversational systems. Along the way to further understand the mechanisms of deep neural network models and even the human brain, we discuss the future works of vision from perspectives of pre-training and tuning techniques.

\subsection{Advanced Pre-training}\label{sec:future_pre_training}
Previous works use supervised or self-supervised methods to guide models to learn representations of our visual and visual-text world. The supervised pre-training method is a mainstream practice of the traditional transfer learning paradigm  \cite{tan2018survey,zhuang2020comprehensive, zhang2021survey}, While self-supervised pre-training scales pre-trained models up to foundation models (introduced in Section \ref{sec:background_pretrain}). Although encouraging progress has been made, as data continues to accumulate, we expect that future pre-training techniques will be able to constantly scale up the model size and improve the capabilities of foundation models. Here, we discuss the future directions of model pre-training from three perspectives: data, models, and optimization.

\subsubsection{Data}\label{sec:future_pre_training_data}
Quality data are the nourishment of foundation models. To realize the promise of future foundation models, it is expected to acquire more fundamental knowledge from the open-world multi-modal data with characteristics as follows: 
\begin{itemize}
\item Increasing scales: Concerning the data volume, large vision models that learn knowledge from large-scale datasets are empirically proven effective for adapting to downstream tasks via tuning techniques. However, compared with human vision, existing large-scale vision datasets remain far from the amount of data that humans learn from. On the contrary, the situation in NLP can be different as large language models can be regarded as having a wide knowledge of the Internet, making them more capable in some NLP tasks, \textit{i.e.}, chatGPT. 
To scale up the data volume, multi-modal data (e.g., image, video, audio, and text), multi-source data (e.g., Internet, generative models such as NeRF \cite{mildenhall2021nerf} and Diffusion \cite{rombach2022high, yang2023diffusion}), and multi-sensor data (e.g., different types of cameras, biomarkers, and ambient sensors) can be considered for training large models. 
\item 
High quality: Before arbitrarily collecting large-scale data, determining what data and how much data are essential concerns. The newly collected data can be redundant or noisy, respectively leading to limited or even negative effects on the model. Chen \textit{et al.} \cite{chen2022principle} introduced the diversity rule at the level of feature representation. However, there remains a lack of investigation into the quality at the data level for existing benchmark datasets, giving rise to research on topics such as out-of-distribution generalization and tolerance to noise (see Section \ref{sec:background_static}). Further investigating measurable factors of data quality (e.g., 16 dimensions summarized in \cite{evans2006scaling}) and their corresponding consequences on the large model can bring a large impact on the machine intelligence community. Findings will guide evidence-oriented data collection and effectively reduce the expensive labeling cost.
\item
Security and privacy are always the priority throughout the life cycle of the data, especially for domains such as healthcare and finance when interacting with large models  on the cloud \cite{chen2012data}. Issues around cloud computing can be grouped into four aspects: 1) users' control over the data, 2) authorized replication, 3) legal requirements, and 4) cloud subcontractors' processing  
 \cite{sun2014data}. Protective actions can be taken at the data level to prevent attacks such as re-identification, dataset reconstruction, and tracing \cite{kaissis2020secure}. 

\end{itemize}

\subsubsection{Models}\label{sec:pretraining_model}
Given the multimodal, multi-source, multi-sensor data, vision large pre-trained models are expected to continuously accumulate knowledge from the new data in an interpretable and secure mechanism. 

\begin{itemize}
\item Theoretical support: Training models with theoretical support from statistical and biological perspectives can make them more interpretable, explainable, and improvable.    
In the regime of large models, there are a certain number of recent works motivated by some theoretical definitions from the statistical perspective \cite{ye2021towards,liu2022closer,zhang2019bridging,tripuraneni2020theory}, by which a generalization bound are used to promise efficient knowledge transfer. Except for the statistical aspect, biological and neuroscience discoveries also benefit the development of deep neural networks, which can provide more insights and inspire new ideas for future large vision models. Recent works \cite{gibson2014ecological,lake2015human} discussed in Section \ref{sec:background_theory_biological} are mainly delayed from one to another as they are intended to explain each other's observed domain empirical realities instead of truly inspiring new ideas. Basic neural network connections are inspired by how brain neuron works, but it has not yet been known exactly how the human brain learns new knowledge. As such, it is also not clear if the knowledge acquired by existing large models via back-propagation can be effective. 
On one hand, we humans are sentient beings and acquire knowledge via multiple sensations: vision, sound, haptic, taste, etc. On the other hand, the human brain can be very efficient to activate just a small portion of neurons to complete a task, while existing foundation models do not. By far, the built intelligence is different from the most intelligent and efficient machine in the world (i.e., the human brain). Understanding the brain can be the next turning point (i.e., artificial general intelligence), which brings serious ethical issues. 

\item Continuously updating:
As introduced in Section \ref{sec:background_notation}, a model can be featured with its domains and tasks with their feature space and label space. We expect that foundation models will scale up not only at the parameter aspect but also at aspects of domains and tasks. 
For a single domain, it can have multiple tasks. Models such as Gato \cite{reed2022generalist} and  Flamingo \cite{alayrac2022flamingo} are pre-trained with multiple tasks, where the former covers vision tasks while the latter even covers both NLP and vision tasks. 
For a single task, a classification task needs to handle novel unseen classes, 
which is defined by an active learning paradigm (i.e., lifelong learning or continual learning). In contrast to batch learning where all training data is available at once, continual learning represents a family of methods that accumulate knowledge and learn continuously with data available in sequential order \cite{qu2021recent}. The future stronger vision and visual-language models will bring a more profound impact on other domains via the multidisciplinary coevolutionary process.

\item Security:
Aside from privacy issues at the data level, large foundation models (i.e., at the model level) can also be vulnerable to attacks. Foundation models allow users to easily plug and
unplug via APIs, which raises security and privacy concerns. such as adversarial attacks and model-inversion. 
Kaissis \textit{et al.} \cite{kaissis2020secure} introduced the advantages of federated learning and provided an outlook for future works. 
Although federated learning can mitigate data-level privacy issues, it can be vulnerable to adversarial attack \cite{usynin2021adversarial}. Around privacy-preserving AI, the adversarial attack will attract more research in the near future. 

\end{itemize}

\subsubsection{Optimization}\label{sec:pretraining_opt}
Current foundation models are generally optimized with back-propagation and reinforcement learning with human feedback. Optimization itself can be relying on hardware devices, hyperparameter configuration, and algorithms as follows: 
\begin{itemize}
\item Hardware:
Recent large models are trained with GPUs, which is unaffordable for general researchers or small companies. Fortunately, the newly released NVIDIA Hopper H100 GPU \cite{choquette2023nvidia} supports FP8 format for accelerating compute-intensive Transformer models (around 9 times faster than previous A100 GPU for training), making trillion-parameter models within the reach of all researchers. While the inference speedup of H100 compared with A100 can be 30 times faster, making tuning a promising direction. 

\item Hyperparameter configuration: In machine learning, hyperparameters such as initial learning rate, batch size, and task-specific parameters often considerably impact performance. To avoid the manual process of trial-and-error, hyperparameter optimization is a sub-field of automated machine learning, which aims to identify a well-performing combination of hyperparameters. Simple techniques are grid or random search. Recent advances in hyperparameter optimization are  evolution strategies, Bayesian optimization, Hyperband, etc. \cite{bischl2021hyperparameter}.

\item Algorithm: The combination of back-propagation and stochastic gradient descent remains the mainstream algorithm to make foundation models optimize towards some statistical goals (e.g., the probability that a picture is identified as a cat). Meanwhile, reinforcement learning with human feedback brings more raw human opinions, which can be some kind of human-machine interactions that align the pre-trained large models to more specific human desired tasks. 

\end{itemize}

\subsection{Tuning Techinques}
As introduced in Section \ref{sec:petl_techniques}, recent developments in visual tuning techniques can be regarded as originating from the prompt tuning of the NLP domain and working towards the PETL direction. Then a couple of adapter methods are proposed, showing better performance than visual prompt methods but lacking interpretability. (It is expected that visual tuning techniques will be implemented on more existing benchmarks, their reformed versions, and the emerging brand-new benchmarks, which will not be listed in this survey. Readers are recommended to refer to the benchmarks of their targeting domains.) The bias-tuning and LoRA methods further reduced the number of parameters, leading to direct parameter tuning methods via addition or decomposition. More recent works are grouped as remapping tuning, among which NAS-based methods \cite{xie2021weight,han2021dynamic} show an even more aggressive PETL manner. These techniques provide exciting research foundations for developing future prompts, leading to 
 better use of language and visual knowledge stored in large models via guidance and interaction, respectively.
We discuss three core progressive interaction aspects: interpretable prompt, conversational guidance, and diversified interactions, that researchers will discern explosive development as follows.

\subsubsection{Interpretable Prompt}
Prompt engineering will work from intuitive design to more understandable and interpretable directions. 
Existing text or visual prompts are more like implicit guidance at the high level, describing what is the downstream visual task. As introduced in Section \ref{sec:prompt_tuning}, many works attempted to learn prompts to facilitate visual downstream tasks. Despite some progress, they suffered from poor interpretability, i.e., it remains difficult to understand what prompts the network has learned. For example, some works (e.g., VPT) learn unordered token-based prompts, which can not be visualized into an understandable prompt. Chen \textit{et al.} \cite{chen2022good} attempted to learn understandable prompts. Regarding other tuning techniques such as adapter, parameter-based, and remapping ones, they are also faced with the interpretability issue, as they intrinsically aim to reduce the number of tuned parameters for adapting the downstream tasks to the large model. Hence, future research should answer questions such as what are good text and vision 
prompts, and how to evaluate them throughout the learning pipeline (from the input side to the output side); the relationship between vision and text prompts, and in what situations visual and text prompts can be mutually replaced; How to design explicit, consistent, and logical prompts that enable a large model to adapt efficiently. 

\subsubsection{Conversational Guidance} 
We observe the development of visual tuning will lead to new jobs such as prompt engineers who have expertise in providing guidance to large-scale visual-language models such as Sora \cite{liu2024sora}. 
Multi-round conversational systems can provide a natural platform that guides models to adapt toward the desired task goals \cite{wu2023visual}. 
It is generally expected that vision models will homogenize with language models \cite{radford2021learning,jia2021scaling,yu2022coca,singh2022flava,alayrac2022flamingo,singh2022revisiting,wu2023visual}. However, due to the fact that “a picture is worth a thousand words”, the development of visual tuning is somehow behind the success of large language models (detailed in Section \ref{sec:background_theory_model}). 
Specifically, concerning data complexity and scenario diversity, the industrial applications of the vision domain (not common application scenarios such as autonomous vehicles, transportation recommendation \cite{liu2021exploring}, whether prediction \cite{nguyen2023climax}, protein design \cite{verkuil2022language}, etc.) are highly demanding for customization based on the specific task requirements. 
Given a tumor detection task, a prompt engineer will select or design good segmentation samples in multiple rounds of conversation with the large models to improve some core steps of the task by referring to various agents \cite{durante2024agent} or tools \cite{qin2023tool} and eventually achieve acceptable results for production.

\subsubsection{Diversified Interactions} 
In addition to the interaction in a conversational system with text and visual prompts, interactions in vision can be more diversified. Humans can gradually build up the evaluation standard themselves and then practice (i.e., learn or train for a model) towards higher standards. 
Existing self-learning models have not set up mechanisms with progressively improving goals. We expect, in the long run, universal AI or strong AI in a specific domain will evolve in the form of prompts, guidance, and diversified interactions. 
In recent works \cite{wang2022images,zhang2023makes}, a segmentation sample is also used as a prompt to tell the model what task will be performed. Currently, interactions in image synthesis use text prompts and sketch images \cite{rombach2022high}, enabling everyone to become a visual content creator. These visual interactions can be regarded as a kind of visual prompt based on the image. Image can represent a limited part of visual interaction scenarios, which can be regarded as tasks with static viewpoints but provide basic conditions for more plentiful visual interaction. Current image-based interactions via prompts are also known as in-context learning, which aims to mimic the efficient visual understanding of the human brain and intrinsically narrows the searching space of foundation models \cite{wang2022images}. 
In addition to the simple aspect of navigating large models to downstream tasks, there are more diversified interaction scenarios that provide plenty of egocentric visual interactions such as robots, drones, and bionic robot dogs \cite{driess2023palm,vemprala2023chatgpt}. These video data provide interactive ecological environments that enable the development of human vision mentioned in Section \ref{sec:background_theory_biological}.
Although tuning foundation models pre-trained via self-supervised learning indicates a promising future direction, future visual interactions will rely on advanced pre-training techniques (knowledge accumulation techniques) that are beyond currently tested ones that are based on generating masked pixels or contrastive learning. Promising long-term directions that enable diversified interactions can involve emerging technologies such as brain-computer interface, quantum computing, event cameras, etc. This will lead to new generalization capabilities on top of the future “collection-labeling-training-feedback” cycle system.

\section{Conclusion}\label{sec:conclusion}
This survey summarized visual tuning techniques, particularly focusing on the recent state of visual tuning in the coming regime of large models. Starting from fine-tuning, existing states of prompt tuning, adapter tuning, parameter tuning, and remapping tuning are systematically investigated and compared based on a comprehensive understanding of their technical details. Based on the expected emerging large models, future visual tuning directions are discussed from perspectives of prompt, guidance, interaction, and optimization. We hope this first survey on the latest state of visual tuning will offer a new perspective to researchers in the era of large models, facilitating their research based on a better understanding of the current state and grasping the future core research challenges. 

\section{Acknowledgments}
This work is supported by the National Key R\&D Program of China (2022ZD0160300), the National Natural Science Foundation of China (NO. 62276004 and NO. 61932020), Huawei Technologies under Grant No.: P0038941, and the major key project of PCL, China (No. PCL2021A12). The authors would like to thank the editors and anonymous reviewers who helped improve this manuscript.

\bibliographystyle{ACM-Reference-Format}
\bibliography{sample-base-bruce}


\begin{thebibliography}{301}


\ifx \showCODEN    \undefined \def \showCODEN     #1{\unskip}     \fi
\ifx \showDOI      \undefined \def \showDOI       #1{#1}\fi
\ifx \showISBNx    \undefined \def \showISBNx     #1{\unskip}     \fi
\ifx \showISBNxiii \undefined \def \showISBNxiii  #1{\unskip}     \fi
\ifx \showISSN     \undefined \def \showISSN      #1{\unskip}     \fi
\ifx \showLCCN     \undefined \def \showLCCN      #1{\unskip}     \fi
\ifx \shownote     \undefined \def \shownote      #1{#1}          \fi
\ifx \showarticletitle \undefined \def \showarticletitle #1{#1}   \fi
\ifx \showURL      \undefined \def \showURL       {\relax}        \fi
\providecommand\bibfield[2]{#2}
\providecommand\bibinfo[2]{#2}
\providecommand\natexlab[1]{#1}
\providecommand\showeprint[2][]{arXiv:#2}

\bibitem[Alayrac et~al\mbox{.}(2022)]%
        {alayrac2022flamingo}
\bibfield{author}{\bibinfo{person}{Jean-Baptiste Alayrac}, \bibinfo{person}{Jeff Donahue}, \bibinfo{person}{Pauline Luc}, \bibinfo{person}{Antoine Miech}, \bibinfo{person}{Iain Barr}, \bibinfo{person}{Yana Hasson}, \bibinfo{person}{Karel Lenc}, \bibinfo{person}{Arthur Mensch}, \bibinfo{person}{Katie Millican}, \bibinfo{person}{Malcolm Reynolds}, {et~al\mbox{.}}} \bibinfo{year}{2022}\natexlab{}.
\newblock \showarticletitle{Flamingo: a visual language model for few-shot learning}.
\newblock \bibinfo{journal}{\emph{Preprint at https://arxiv.org/abs/2204.14198}} (\bibinfo{year}{2022}).
\newblock


\bibitem[Albawi et~al\mbox{.}(2017)]%
        {albawi2017understanding}
\bibfield{author}{\bibinfo{person}{Saad Albawi}, \bibinfo{person}{Tareq~Abed Mohammed}, {and} \bibinfo{person}{Saad Al-Zawi}.} \bibinfo{year}{2017}\natexlab{}.
\newblock \showarticletitle{Understanding of a convolutional neural network}. In \bibinfo{booktitle}{\emph{Proc. ICET}}. IEEE, \bibinfo{pages}{1--6}.
\newblock


\bibitem[Amplayo et~al\mbox{.}(2022)]%
        {amplayo2022attribute}
\bibfield{author}{\bibinfo{person}{Reinald~Kim Amplayo}, \bibinfo{person}{Kang~Min Yoo}, {and} \bibinfo{person}{Sang-Woo Lee}.} \bibinfo{year}{2022}\natexlab{}.
\newblock \showarticletitle{Attribute Injection for Pretrained Language Models: A New Benchmark and An Efficient Method}. In \bibinfo{booktitle}{\emph{Proc. COLING}}. \bibinfo{pages}{1051--1064}.
\newblock


\bibitem[Angarano et~al\mbox{.}(2022)]%
        {angarano2022generative}
\bibfield{author}{\bibinfo{person}{Simone Angarano}, \bibinfo{person}{Francesco Salvetti}, \bibinfo{person}{Mauro Martini}, {and} \bibinfo{person}{Marcello Chiaberge}.} \bibinfo{year}{2022}\natexlab{}.
\newblock \showarticletitle{Generative Adversarial Super-Resolution at the Edge with Knowledge Distillation}.
\newblock \bibinfo{journal}{\emph{Preprint at https://arxiv.org/abs/2209.03355}} (\bibinfo{year}{2022}).
\newblock


\bibitem[Ashok et~al\mbox{.}(2018)]%
        {ashok2018n2n}
\bibfield{author}{\bibinfo{person}{Anubhav Ashok}, \bibinfo{person}{Nicholas Rhinehart}, \bibinfo{person}{Fares Beainy}, {and} \bibinfo{person}{Kris~M Kitani}.} \bibinfo{year}{2018}\natexlab{}.
\newblock \showarticletitle{N2n learning: Network to network compression via policy gradient reinforcement learning}. In \bibinfo{booktitle}{\emph{Proc. ICLR}}.
\newblock


\bibitem[Ba and Caruana(2014)]%
        {ba2014deep}
\bibfield{author}{\bibinfo{person}{Jimmy Ba} {and} \bibinfo{person}{Rich Caruana}.} \bibinfo{year}{2014}\natexlab{}.
\newblock \showarticletitle{Do deep nets really need to be deep?}. In \bibinfo{booktitle}{\emph{Proc. NeurIPS}}, Vol.~\bibinfo{volume}{27}.
\newblock


\bibitem[Bengio et~al\mbox{.}(2013)]%
        {bengio2013representation}
\bibfield{author}{\bibinfo{person}{Yoshua Bengio}, \bibinfo{person}{Aaron Courville}, {and} \bibinfo{person}{Pascal Vincent}.} \bibinfo{year}{2013}\natexlab{}.
\newblock \showarticletitle{Representation learning: A review and new perspectives}.
\newblock \bibinfo{journal}{\emph{IEEE Trans. Patt. Anal. Mach. Intell.}} \bibinfo{volume}{35}, \bibinfo{number}{8} (\bibinfo{year}{2013}), \bibinfo{pages}{1798--1828}.
\newblock


\bibitem[Berg et~al\mbox{.}(2022)]%
        {berg2022prompt}
\bibfield{author}{\bibinfo{person}{Hugo Berg}, \bibinfo{person}{Siobhan~Mackenzie Hall}, \bibinfo{person}{Yash Bhalgat}, \bibinfo{person}{Wonsuk Yang}, \bibinfo{person}{Hannah~Rose Kirk}, \bibinfo{person}{Aleksandar Shtedritski}, {and} \bibinfo{person}{Max Bain}.} \bibinfo{year}{2022}\natexlab{}.
\newblock \showarticletitle{A prompt array keeps the bias away: Debiasing vision-language models with adversarial learning}.
\newblock \bibinfo{journal}{\emph{Preprint at https://arxiv.org/abs/2203.11933}} (\bibinfo{year}{2022}).
\newblock


\bibitem[Bischl et~al\mbox{.}(2021)]%
        {bischl2021hyperparameter}
\bibfield{author}{\bibinfo{person}{Bernd Bischl}, \bibinfo{person}{Martin Binder}, \bibinfo{person}{Michel Lang}, \bibinfo{person}{Tobias Pielok}, \bibinfo{person}{Jakob Richter}, \bibinfo{person}{Stefan Coors}, \bibinfo{person}{Janek Thomas}, \bibinfo{person}{Theresa Ullmann}, \bibinfo{person}{Marc Becker}, \bibinfo{person}{Anne-Laure Boulesteix}, {et~al\mbox{.}}} \bibinfo{year}{2021}\natexlab{}.
\newblock \showarticletitle{Hyperparameter optimization: Foundations, algorithms, best practices, and open challenges}.
\newblock \bibinfo{journal}{\emph{Wiley Interdisciplinary Reviews: Data Mining and Knowledge Discovery}} (\bibinfo{year}{2021}), \bibinfo{pages}{e1484}.
\newblock


\bibitem[Bommasani et~al\mbox{.}(2021)]%
        {bommasani2021opportunities}
\bibfield{author}{\bibinfo{person}{Rishi Bommasani}, \bibinfo{person}{Drew~A Hudson}, \bibinfo{person}{Ehsan Adeli}, \bibinfo{person}{Russ Altman}, \bibinfo{person}{Simran Arora}, \bibinfo{person}{Sydney von Arx}, \bibinfo{person}{Michael~S Bernstein}, \bibinfo{person}{Jeannette Bohg}, \bibinfo{person}{Antoine Bosselut}, \bibinfo{person}{Emma Brunskill}, {et~al\mbox{.}}} \bibinfo{year}{2021}\natexlab{}.
\newblock \showarticletitle{On the opportunities and risks of foundation models}.
\newblock \bibinfo{journal}{\emph{Preprint at https://arxiv.org/abs/2108.07258}} (\bibinfo{year}{2021}).
\newblock


\bibitem[Bowman et~al\mbox{.}(2023)]%
        {bowman2023carte}
\bibfield{author}{\bibinfo{person}{Benjamin Bowman}, \bibinfo{person}{Alessandro Achille}, \bibinfo{person}{Luca Zancato}, \bibinfo{person}{Matthew Trager}, \bibinfo{person}{Pramuditha Perera}, \bibinfo{person}{Giovanni Paolini}, {and} \bibinfo{person}{Stefano Soatto}.} \bibinfo{year}{2023}\natexlab{}.
\newblock \showarticletitle{$\backslash$A-la-carte Prompt Tuning (APT): Combining Distinct Data Via Composable Prompting}.
\newblock \bibinfo{journal}{\emph{Preprint at https://arxiv.org/abs/2302.07994}} (\bibinfo{year}{2023}).
\newblock


\bibitem[Brown et~al\mbox{.}(2020)]%
        {brown2020language}
\bibfield{author}{\bibinfo{person}{Tom Brown}, \bibinfo{person}{Benjamin Mann}, \bibinfo{person}{Nick Ryder}, \bibinfo{person}{Melanie Subbiah}, \bibinfo{person}{Jared~D Kaplan}, \bibinfo{person}{Prafulla Dhariwal}, \bibinfo{person}{Arvind Neelakantan}, \bibinfo{person}{Pranav Shyam}, \bibinfo{person}{Girish Sastry}, \bibinfo{person}{Amanda Askell}, {et~al\mbox{.}}} \bibinfo{year}{2020}\natexlab{}.
\newblock \showarticletitle{Language models are few-shot learners}. In \bibinfo{booktitle}{\emph{Proc. NeurIPS}}, Vol.~\bibinfo{volume}{33}. \bibinfo{pages}{1877--1901}.
\newblock


\bibitem[Bruce et~al\mbox{.}(2024)]%
        {bruce2024egcn}
\bibfield{author}{\bibinfo{person}{XB Bruce}, \bibinfo{person}{Yan Liu}, \bibinfo{person}{Keith~CC Chan}, {and} \bibinfo{person}{Chang~Wen Chen}.} \bibinfo{year}{2024}\natexlab{}.
\newblock \showarticletitle{EGCN++: A New Fusion Strategy for Ensemble Learning in Skeleton-Based Rehabilitation Exercise Assessment}.
\newblock \bibinfo{journal}{\emph{IEEE Trans. Patt. Anal. Mach. Intell.}} \bibinfo{number}{01} (\bibinfo{year}{2024}), \bibinfo{pages}{1--16}.
\newblock


\bibitem[Bruce et~al\mbox{.}(2022)]%
        {bruce2022mmnet}
\bibfield{author}{\bibinfo{person}{XB Bruce}, \bibinfo{person}{Yan Liu}, \bibinfo{person}{Xiang Zhang}, \bibinfo{person}{Sheng-hua Zhong}, {and} \bibinfo{person}{Keith~CC Chan}.} \bibinfo{year}{2022}\natexlab{}.
\newblock \showarticletitle{Mmnet: A model-based multimodal network for human action recognition in rgb-d videos}.
\newblock \bibinfo{journal}{\emph{IEEE Trans. Patt. Anal. Mach. Intell.}} (\bibinfo{year}{2022}).
\newblock


\bibitem[Bu et~al\mbox{.}(2022)]%
        {bu2022differentially}
\bibfield{author}{\bibinfo{person}{Zhiqi Bu}, \bibinfo{person}{Yu-Xiang Wang}, \bibinfo{person}{Sheng Zha}, {and} \bibinfo{person}{George Karypis}.} \bibinfo{year}{2022}\natexlab{}.
\newblock \showarticletitle{Differentially Private Bias-Term only Fine-tuning of Foundation Models}.
\newblock \bibinfo{journal}{\emph{Preprint at https://arxiv.org/abs/2210.00036}} (\bibinfo{year}{2022}).
\newblock


\bibitem[Buciluǎ et~al\mbox{.}(2006)]%
        {buciluǎ2006model}
\bibfield{author}{\bibinfo{person}{Cristian Buciluǎ}, \bibinfo{person}{Rich Caruana}, {and} \bibinfo{person}{Alexandru Niculescu-Mizil}.} \bibinfo{year}{2006}\natexlab{}.
\newblock \showarticletitle{Model compression}. In \bibinfo{booktitle}{\emph{ACM SIGKDD}}. \bibinfo{pages}{535--541}.
\newblock


\bibitem[Bulat and Tzimiropoulos(2022)]%
        {bulat2022language}
\bibfield{author}{\bibinfo{person}{Adrian Bulat} {and} \bibinfo{person}{Georgios Tzimiropoulos}.} \bibinfo{year}{2022}\natexlab{}.
\newblock \showarticletitle{Language-Aware Soft Prompting for Vision \& Language Foundation Models}.
\newblock \bibinfo{journal}{\emph{Preprint at https://arxiv.org/abs/2210.01115}} (\bibinfo{year}{2022}).
\newblock


\bibitem[Cai et~al\mbox{.}(2018a)]%
        {cai2018efficient}
\bibfield{author}{\bibinfo{person}{Han Cai}, \bibinfo{person}{Tianyao Chen}, \bibinfo{person}{Weinan Zhang}, \bibinfo{person}{Yong Yu}, {and} \bibinfo{person}{Jun Wang}.} \bibinfo{year}{2018}\natexlab{a}.
\newblock \showarticletitle{Efficient architecture search by network transformation}. In \bibinfo{booktitle}{\emph{Proc. AAAI}}, Vol.~\bibinfo{volume}{32}.
\newblock


\bibitem[Cai et~al\mbox{.}(2018b)]%
        {cai2018path}
\bibfield{author}{\bibinfo{person}{Han Cai}, \bibinfo{person}{Jiacheng Yang}, \bibinfo{person}{Weinan Zhang}, \bibinfo{person}{Song Han}, {and} \bibinfo{person}{Yong Yu}.} \bibinfo{year}{2018}\natexlab{b}.
\newblock \showarticletitle{Path-level network transformation for efficient architecture search}. In \bibinfo{booktitle}{\emph{Proc. ICML}}. PMLR, \bibinfo{pages}{678--687}.
\newblock


\bibitem[Calonder et~al\mbox{.}(2010)]%
        {calonder2010brief}
\bibfield{author}{\bibinfo{person}{Michael Calonder}, \bibinfo{person}{Vincent Lepetit}, \bibinfo{person}{Christoph Strecha}, {and} \bibinfo{person}{Pascal Fua}.} \bibinfo{year}{2010}\natexlab{}.
\newblock \showarticletitle{Brief: Binary robust independent elementary features}. In \bibinfo{booktitle}{\emph{Proc. ECCV}}. Springer, \bibinfo{pages}{778--792}.
\newblock


\bibitem[Carreira and Zisserman(2017)]%
        {carreira2017quo}
\bibfield{author}{\bibinfo{person}{Joao Carreira} {and} \bibinfo{person}{Andrew Zisserman}.} \bibinfo{year}{2017}\natexlab{}.
\newblock \showarticletitle{Quo vadis, action recognition? a new model and the kinetics dataset}. In \bibinfo{booktitle}{\emph{Proc. CVPR}}. \bibinfo{pages}{6299--6308}.
\newblock


\bibitem[Chang et~al\mbox{.}(2019)]%
        {DBLP:conf/nips/ChangZGMXP19}
\bibfield{author}{\bibinfo{person}{Jianlong Chang}, \bibinfo{person}{Xinbang Zhang}, \bibinfo{person}{Yiwen Guo}, \bibinfo{person}{Gaofeng Meng}, \bibinfo{person}{Shiming Xiang}, {and} \bibinfo{person}{Chunhong Pan}.} \bibinfo{year}{2019}\natexlab{}.
\newblock \showarticletitle{{DATA:} Differentiable ArchiTecture Approximation}. In \bibinfo{booktitle}{\emph{Proc. NeurIPS}}. \bibinfo{pages}{874--884}.
\newblock


\bibitem[Chen et~al\mbox{.}(2021b)]%
        {chen2021cross}
\bibfield{author}{\bibinfo{person}{Defang Chen}, \bibinfo{person}{Jian-Ping Mei}, \bibinfo{person}{Yuan Zhang}, \bibinfo{person}{Can Wang}, \bibinfo{person}{Zhe Wang}, \bibinfo{person}{Yan Feng}, {and} \bibinfo{person}{Chun Chen}.} \bibinfo{year}{2021}\natexlab{b}.
\newblock \showarticletitle{Cross-layer distillation with semantic calibration}. In \bibinfo{booktitle}{\emph{Proc. AAAI}}, Vol.~\bibinfo{volume}{35}. \bibinfo{pages}{7028--7036}.
\newblock


\bibitem[Chen and Zhao(2012)]%
        {chen2012data}
\bibfield{author}{\bibinfo{person}{Deyan Chen} {and} \bibinfo{person}{Hong Zhao}.} \bibinfo{year}{2012}\natexlab{}.
\newblock \showarticletitle{Data security and privacy protection issues in cloud computing}. In \bibinfo{booktitle}{\emph{2012 international conference on computer science and electronics engineering}}, Vol.~\bibinfo{volume}{1}. IEEE, \bibinfo{pages}{647--651}.
\newblock


\bibitem[Chen et~al\mbox{.}(2022e)]%
        {chen2022prompt}
\bibfield{author}{\bibinfo{person}{Guangyi Chen}, \bibinfo{person}{Weiran Yao}, \bibinfo{person}{Xiangchen Song}, \bibinfo{person}{Xinyue Li}, \bibinfo{person}{Yongming Rao}, {and} \bibinfo{person}{Kun Zhang}.} \bibinfo{year}{2022}\natexlab{e}.
\newblock \showarticletitle{Prompt Learning with Optimal Transport for Vision-Language Models}.
\newblock \bibinfo{journal}{\emph{Preprint at https://arxiv.org/abs/2210.01253}} (\bibinfo{year}{2022}).
\newblock


\bibitem[Chen et~al\mbox{.}(2022c)]%
        {chen2022conv}
\bibfield{author}{\bibinfo{person}{Hao Chen}, \bibinfo{person}{Ran Tao}, \bibinfo{person}{Han Zhang}, \bibinfo{person}{Yidong Wang}, \bibinfo{person}{Wei Ye}, \bibinfo{person}{Jindong Wang}, \bibinfo{person}{Guosheng Hu}, {and} \bibinfo{person}{Marios Savvides}.} \bibinfo{year}{2022}\natexlab{c}.
\newblock \showarticletitle{Conv-Adapter: Exploring Parameter Efficient Transfer Learning for ConvNets}.
\newblock \bibinfo{journal}{\emph{Preprint at https://arxiv.org/abs/2208.07463}} (\bibinfo{year}{2022}).
\newblock


\bibitem[Chen et~al\mbox{.}(2020b)]%
        {chen2020learning}
\bibfield{author}{\bibinfo{person}{Hanting Chen}, \bibinfo{person}{Yunhe Wang}, \bibinfo{person}{Chang Xu}, \bibinfo{person}{Chao Xu}, {and} \bibinfo{person}{Dacheng Tao}.} \bibinfo{year}{2020}\natexlab{b}.
\newblock \showarticletitle{Learning student networks via feature embedding}.
\newblock \bibinfo{journal}{\emph{IEEE Trans. Neur. Netw. Learn. Syst.}} \bibinfo{volume}{32}, \bibinfo{number}{1} (\bibinfo{year}{2020}), \bibinfo{pages}{25--35}.
\newblock


\bibitem[Chen et~al\mbox{.}(2022d)]%
        {chen2022multi}
\bibfield{author}{\bibinfo{person}{Haoran Chen}, \bibinfo{person}{Zuxuan Wu}, {and} \bibinfo{person}{Yu-Gang Jiang}.} \bibinfo{year}{2022}\natexlab{d}.
\newblock \showarticletitle{Multi-Prompt Alignment for Multi-source Unsupervised Domain Adaptation}.
\newblock \bibinfo{journal}{\emph{Preprint at https://arxiv.org/abs/2209.15210}} (\bibinfo{year}{2022}).
\newblock


\bibitem[Chen et~al\mbox{.}(2020a)]%
        {chen2020generative}
\bibfield{author}{\bibinfo{person}{Mark Chen}, \bibinfo{person}{Alec Radford}, \bibinfo{person}{Rewon Child}, \bibinfo{person}{Jeffrey Wu}, \bibinfo{person}{Heewoo Jun}, \bibinfo{person}{David Luan}, {and} \bibinfo{person}{Ilya Sutskever}.} \bibinfo{year}{2020}\natexlab{a}.
\newblock \showarticletitle{Generative pretraining from pixels}. In \bibinfo{booktitle}{\emph{Proc. ICML}}. PMLR, \bibinfo{pages}{1691--1703}.
\newblock


\bibitem[Chen et~al\mbox{.}(2021a)]%
        {chen2021distilling}
\bibfield{author}{\bibinfo{person}{Pengguang Chen}, \bibinfo{person}{Shu Liu}, \bibinfo{person}{Hengshuang Zhao}, {and} \bibinfo{person}{Jiaya Jia}.} \bibinfo{year}{2021}\natexlab{a}.
\newblock \showarticletitle{Distilling knowledge via knowledge review}. In \bibinfo{booktitle}{\emph{Proc. CVPR}}. \bibinfo{pages}{5008--5017}.
\newblock


\bibitem[Chen et~al\mbox{.}(2022b)]%
        {chen2022adaptformer}
\bibfield{author}{\bibinfo{person}{Shoufa Chen}, \bibinfo{person}{Chongjian Ge}, \bibinfo{person}{Zhan Tong}, \bibinfo{person}{Jiangliu Wang}, \bibinfo{person}{Yibing Song}, \bibinfo{person}{Jue Wang}, {and} \bibinfo{person}{Ping Luo}.} \bibinfo{year}{2022}\natexlab{b}.
\newblock \showarticletitle{AdaptFormer: Adapting Vision Transformers for Scalable Visual Recognition}.
\newblock \bibinfo{journal}{\emph{Preprint at https://arxiv.org/abs/2205.13535}} (\bibinfo{year}{2022}).
\newblock


\bibitem[Chen et~al\mbox{.}(2016)]%
        {chen2016net2net}
\bibfield{author}{\bibinfo{person}{Tianqi Chen}, \bibinfo{person}{Ian Goodfellow}, {and} \bibinfo{person}{Jonathon Shlens}.} \bibinfo{year}{2016}\natexlab{}.
\newblock \showarticletitle{Net2net: Accelerating learning via knowledge transfer}. In \bibinfo{booktitle}{\emph{Proc. ICLR}}.
\newblock


\bibitem[Chen et~al\mbox{.}(2022f)]%
        {chen2022principle}
\bibfield{author}{\bibinfo{person}{Tianlong Chen}, \bibinfo{person}{Zhenyu Zhang}, \bibinfo{person}{Yu Cheng}, \bibinfo{person}{Ahmed Awadallah}, {and} \bibinfo{person}{Zhangyang Wang}.} \bibinfo{year}{2022}\natexlab{f}.
\newblock \showarticletitle{The principle of diversity: Training stronger vision transformers calls for reducing all levels of redundancy}. In \bibinfo{booktitle}{\emph{Proc. CVPR}}. \bibinfo{pages}{12020--12030}.
\newblock


\bibitem[Chen et~al\mbox{.}(2019)]%
        {chen2019progressive}
\bibfield{author}{\bibinfo{person}{Xin Chen}, \bibinfo{person}{Lingxi Xie}, \bibinfo{person}{Jun Wu}, {and} \bibinfo{person}{Qi Tian}.} \bibinfo{year}{2019}\natexlab{}.
\newblock \showarticletitle{Progressive differentiable architecture search: Bridging the depth gap between search and evaluation}. In \bibinfo{booktitle}{\emph{Proc. CVPR}}. \bibinfo{pages}{1294--1303}.
\newblock


\bibitem[Chen et~al\mbox{.}(2022g)]%
        {chen2022good}
\bibfield{author}{\bibinfo{person}{Xiang Chen}, \bibinfo{person}{Ningyu Zhang}, \bibinfo{person}{Lei Li}, \bibinfo{person}{Yunzhi Yao}, \bibinfo{person}{Shumin Deng}, \bibinfo{person}{Chuanqi Tan}, \bibinfo{person}{Fei Huang}, \bibinfo{person}{Luo Si}, {and} \bibinfo{person}{Huajun Chen}.} \bibinfo{year}{2022}\natexlab{g}.
\newblock \showarticletitle{Good Visual Guidance Makes A Better Extractor: Hierarchical Visual Prefix for Multimodal Entity and Relation Extraction}.
\newblock \bibinfo{journal}{\emph{North American Chapter of the Association for Computational Linguistics}} (\bibinfo{year}{2022}).
\newblock


\bibitem[Chen et~al\mbox{.}(2022a)]%
        {chen2022vision}
\bibfield{author}{\bibinfo{person}{Zhe Chen}, \bibinfo{person}{Yuchen Duan}, \bibinfo{person}{Wenhai Wang}, \bibinfo{person}{Junjun He}, \bibinfo{person}{Tong Lu}, \bibinfo{person}{Jifeng Dai}, {and} \bibinfo{person}{Yu Qiao}.} \bibinfo{year}{2022}\natexlab{a}.
\newblock \showarticletitle{Vision transformer adapter for dense predictions}.
\newblock \bibinfo{journal}{\emph{Preprint at https://arxiv.org/abs/2205.08534}} (\bibinfo{year}{2022}).
\newblock


\bibitem[Choquette(2023)]%
        {choquette2023nvidia}
\bibfield{author}{\bibinfo{person}{Jack Choquette}.} \bibinfo{year}{2023}\natexlab{}.
\newblock \showarticletitle{NVIDIA Hopper H100 GPU: Scaling Performance}.
\newblock \bibinfo{journal}{\emph{IEEE Micro}} (\bibinfo{year}{2023}).
\newblock


\bibitem[Chu et~al\mbox{.}(2021a)]%
        {chu2021twins}
\bibfield{author}{\bibinfo{person}{Xiangxiang Chu}, \bibinfo{person}{Zhi Tian}, \bibinfo{person}{Yuqing Wang}, \bibinfo{person}{Bo Zhang}, \bibinfo{person}{Haibing Ren}, \bibinfo{person}{Xiaolin Wei}, \bibinfo{person}{Huaxia Xia}, {and} \bibinfo{person}{Chunhua Shen}.} \bibinfo{year}{2021}\natexlab{a}.
\newblock \showarticletitle{Twins: Revisiting the design of spatial attention in vision transformers}. In \bibinfo{booktitle}{\emph{Proc. NeurIPS}}, Vol.~\bibinfo{volume}{34}. \bibinfo{pages}{9355--9366}.
\newblock


\bibitem[Chu et~al\mbox{.}(2021b)]%
        {chu2021conditional}
\bibfield{author}{\bibinfo{person}{Xiangxiang Chu}, \bibinfo{person}{Zhi Tian}, \bibinfo{person}{Bo Zhang}, \bibinfo{person}{Xinlong Wang}, \bibinfo{person}{Xiaolin Wei}, \bibinfo{person}{Huaxia Xia}, {and} \bibinfo{person}{Chunhua Shen}.} \bibinfo{year}{2021}\natexlab{b}.
\newblock \showarticletitle{Conditional positional encodings for vision transformers}.
\newblock \bibinfo{journal}{\emph{Preprint at https://arxiv.org/abs/2102.10882}} (\bibinfo{year}{2021}).
\newblock


\bibitem[Chu et~al\mbox{.}(2021c)]%
        {chu2021darts}
\bibfield{author}{\bibinfo{person}{Xiangxiang Chu}, \bibinfo{person}{Xiaoxing Wang}, \bibinfo{person}{Bo Zhang}, \bibinfo{person}{Shun Lu}, \bibinfo{person}{Xiaolin Wei}, {and} \bibinfo{person}{Junchi Yan}.} \bibinfo{year}{2021}\natexlab{c}.
\newblock \showarticletitle{Darts-: robustly stepping out of performance collapse without indicators}. In \bibinfo{booktitle}{\emph{Proc. ICLR}}.
\newblock


\bibitem[Chu et~al\mbox{.}(2021d)]%
        {chu2021fairnas}
\bibfield{author}{\bibinfo{person}{Xiangxiang Chu}, \bibinfo{person}{Bo Zhang}, {and} \bibinfo{person}{Ruijun Xu}.} \bibinfo{year}{2021}\natexlab{d}.
\newblock \showarticletitle{Fairnas: Rethinking evaluation fairness of weight sharing neural architecture search}. In \bibinfo{booktitle}{\emph{Proc. CVPR}}. \bibinfo{pages}{12239--12248}.
\newblock


\bibitem[De~Lathauwer et~al\mbox{.}(2000)]%
        {de2000multilinear}
\bibfield{author}{\bibinfo{person}{Lieven De~Lathauwer}, \bibinfo{person}{Bart De~Moor}, {and} \bibinfo{person}{Joos Vandewalle}.} \bibinfo{year}{2000}\natexlab{}.
\newblock \showarticletitle{A multilinear singular value decomposition}.
\newblock \bibinfo{journal}{\emph{SIAM journal on Matrix Analysis and Applications}} \bibinfo{volume}{21}, \bibinfo{number}{4} (\bibinfo{year}{2000}), \bibinfo{pages}{1253--1278}.
\newblock


\bibitem[Dehghani et~al\mbox{.}(2023)]%
        {dehghani2023scaling}
\bibfield{author}{\bibinfo{person}{Mostafa Dehghani}, \bibinfo{person}{Josip Djolonga}, \bibinfo{person}{Basil Mustafa}, \bibinfo{person}{Piotr Padlewski}, \bibinfo{person}{Jonathan Heek}, \bibinfo{person}{Justin Gilmer}, \bibinfo{person}{Andreas Steiner}, \bibinfo{person}{Mathilde Caron}, \bibinfo{person}{Robert Geirhos}, \bibinfo{person}{Ibrahim Alabdulmohsin}, {et~al\mbox{.}}} \bibinfo{year}{2023}\natexlab{}.
\newblock \showarticletitle{Scaling Vision Transformers to 22 Billion Parameters}.
\newblock \bibinfo{journal}{\emph{Preprint at https://arxiv.org/abs/2302.05442}} (\bibinfo{year}{2023}).
\newblock


\bibitem[Dekking et~al\mbox{.}(2005)]%
        {dekking2005modern}
\bibfield{author}{\bibinfo{person}{Frederik~Michel Dekking}, \bibinfo{person}{Cornelis Kraaikamp}, \bibinfo{person}{Hendrik~Paul Lopuha{\"a}}, {and} \bibinfo{person}{Ludolf~Erwin Meester}.} \bibinfo{year}{2005}\natexlab{}.
\newblock \bibinfo{booktitle}{\emph{A Modern Introduction to Probability and Statistics: Understanding why and how}}. Vol.~\bibinfo{volume}{488}.
\newblock \bibinfo{publisher}{Springer}.
\newblock


\bibitem[Deng et~al\mbox{.}(2009)]%
        {deng2009imagenet}
\bibfield{author}{\bibinfo{person}{Jia Deng}, \bibinfo{person}{Wei Dong}, \bibinfo{person}{Richard Socher}, \bibinfo{person}{Li-Jia Li}, \bibinfo{person}{Kai Li}, {and} \bibinfo{person}{Li Fei-Fei}.} \bibinfo{year}{2009}\natexlab{}.
\newblock \showarticletitle{Imagenet: A large-scale hierarchical image database}. In \bibinfo{booktitle}{\emph{2009 CVPR}}. IEEE, \bibinfo{pages}{248--255}.
\newblock


\bibitem[Ding et~al\mbox{.}(2023)]%
        {ding2023parameter}
\bibfield{author}{\bibinfo{person}{Ning Ding}, \bibinfo{person}{Yujia Qin}, \bibinfo{person}{Guang Yang}, \bibinfo{person}{Fuchao Wei}, \bibinfo{person}{Zonghan Yang}, \bibinfo{person}{Yusheng Su}, \bibinfo{person}{Shengding Hu}, \bibinfo{person}{Yulin Chen}, \bibinfo{person}{Chi-Min Chan}, \bibinfo{person}{Weize Chen}, {et~al\mbox{.}}} \bibinfo{year}{2023}\natexlab{}.
\newblock \showarticletitle{Parameter-efficient fine-tuning of large-scale pre-trained language models}.
\newblock \bibinfo{journal}{\emph{Nature Machine Intelligence}} (\bibinfo{year}{2023}), \bibinfo{pages}{1--16}.
\newblock


\bibitem[Dong et~al\mbox{.}(2022b)]%
        {dong2022lpt}
\bibfield{author}{\bibinfo{person}{Bowen Dong}, \bibinfo{person}{Pan Zhou}, \bibinfo{person}{Shuicheng Yan}, {and} \bibinfo{person}{Wangmeng Zuo}.} \bibinfo{year}{2022}\natexlab{b}.
\newblock \showarticletitle{LPT: Long-tailed Prompt Tuning for Image Classification}.
\newblock \bibinfo{journal}{\emph{Preprint at https://arxiv.org/abs/2210.01033}} (\bibinfo{year}{2022}).
\newblock


\bibitem[Dong et~al\mbox{.}(2022a)]%
        {dong2022autoencoders}
\bibfield{author}{\bibinfo{person}{Runpei Dong}, \bibinfo{person}{Zekun Qi}, \bibinfo{person}{Linfeng Zhang}, \bibinfo{person}{Junbo Zhang}, \bibinfo{person}{Jianjian Sun}, \bibinfo{person}{Zheng Ge}, \bibinfo{person}{Li Yi}, {and} \bibinfo{person}{Kaisheng Ma}.} \bibinfo{year}{2022}\natexlab{a}.
\newblock \showarticletitle{Autoencoders as Cross-Modal Teachers: Can Pretrained 2D Image Transformers Help 3D Representation Learning?}
\newblock \bibinfo{journal}{\emph{Preprint at https://arxiv.org/abs/2212.08320}} (\bibinfo{year}{2022}).
\newblock


\bibitem[Dosovitskiy et~al\mbox{.}(2021)]%
        {dosovitskiy2020image}
\bibfield{author}{\bibinfo{person}{Alexey Dosovitskiy}, \bibinfo{person}{Lucas Beyer}, \bibinfo{person}{Alexander Kolesnikov}, \bibinfo{person}{Dirk Weissenborn}, \bibinfo{person}{Xiaohua Zhai}, \bibinfo{person}{Thomas Unterthiner}, \bibinfo{person}{Mostafa Dehghani}, \bibinfo{person}{Matthias Minderer}, \bibinfo{person}{Georg Heigold}, \bibinfo{person}{Sylvain Gelly}, {et~al\mbox{.}}} \bibinfo{year}{2021}\natexlab{}.
\newblock \showarticletitle{An image is worth 16x16 words: Transformers for image recognition at scale}. In \bibinfo{booktitle}{\emph{Proc. ICLR}}.
\newblock


\bibitem[Driess et~al\mbox{.}(2023)]%
        {driess2023palm}
\bibfield{author}{\bibinfo{person}{Danny Driess}, \bibinfo{person}{Fei Xia}, \bibinfo{person}{Mehdi~SM Sajjadi}, \bibinfo{person}{Corey Lynch}, \bibinfo{person}{Aakanksha Chowdhery}, \bibinfo{person}{Brian Ichter}, \bibinfo{person}{Ayzaan Wahid}, \bibinfo{person}{Jonathan Tompson}, \bibinfo{person}{Quan Vuong}, \bibinfo{person}{Tianhe Yu}, {et~al\mbox{.}}} \bibinfo{year}{2023}\natexlab{}.
\newblock \showarticletitle{Palm-e: An embodied multimodal language model}.
\newblock \bibinfo{journal}{\emph{arXiv preprint arXiv:2303.03378}} (\bibinfo{year}{2023}).
\newblock


\bibitem[Durante et~al\mbox{.}(2024)]%
        {durante2024agent}
\bibfield{author}{\bibinfo{person}{Zane Durante}, \bibinfo{person}{Qiuyuan Huang}, \bibinfo{person}{Naoki Wake}, \bibinfo{person}{Ran Gong}, \bibinfo{person}{Jae~Sung Park}, \bibinfo{person}{Bidipta Sarkar}, \bibinfo{person}{Rohan Taori}, \bibinfo{person}{Yusuke Noda}, \bibinfo{person}{Demetri Terzopoulos}, \bibinfo{person}{Yejin Choi}, {et~al\mbox{.}}} \bibinfo{year}{2024}\natexlab{}.
\newblock \showarticletitle{Agent ai: Surveying the horizons of multimodal interaction}.
\newblock \bibinfo{journal}{\emph{arXiv preprint arXiv:2401.03568}} (\bibinfo{year}{2024}).
\newblock


\bibitem[Edalati et~al\mbox{.}(2022)]%
        {edalati2022krona}
\bibfield{author}{\bibinfo{person}{Ali Edalati}, \bibinfo{person}{Marzieh Tahaei}, \bibinfo{person}{Ivan Kobyzev}, \bibinfo{person}{Vahid~Partovi Nia}, \bibinfo{person}{James~J Clark}, {and} \bibinfo{person}{Mehdi Rezagholizadeh}.} \bibinfo{year}{2022}\natexlab{}.
\newblock \showarticletitle{KronA: Parameter Efficient Tuning with Kronecker Adapter}.
\newblock \bibinfo{journal}{\emph{Preprint at https://arxiv.org/abs/2212.10650}} (\bibinfo{year}{2022}).
\newblock


\bibitem[Eichenberg et~al\mbox{.}(2021)]%
        {eichenberg2021magma}
\bibfield{author}{\bibinfo{person}{Constantin Eichenberg}, \bibinfo{person}{Sidney Black}, \bibinfo{person}{Samuel Weinbach}, \bibinfo{person}{Letitia Parcalabescu}, {and} \bibinfo{person}{Anette Frank}.} \bibinfo{year}{2021}\natexlab{}.
\newblock \showarticletitle{MAGMA--Multimodal Augmentation of Generative Models through Adapter-based Finetuning}.
\newblock \bibinfo{journal}{\emph{Preprint at https://arxiv.org/abs/2112.05253}} (\bibinfo{year}{2021}).
\newblock


\bibitem[Elsken et~al\mbox{.}(2018)]%
        {elsken2018simple}
\bibfield{author}{\bibinfo{person}{Thomas Elsken}, \bibinfo{person}{Jan-Hendrik Metzen}, {and} \bibinfo{person}{Frank Hutter}.} \bibinfo{year}{2018}\natexlab{}.
\newblock \showarticletitle{Simple and efficient architecture search for convolutional neural networks}. In \bibinfo{booktitle}{\emph{Proc. ICLR, Workshop Track}}.
\newblock


\bibitem[Ermis et~al\mbox{.}(2022)]%
        {Ermis_2022_CVPR}
\bibfield{author}{\bibinfo{person}{Beyza Ermis}, \bibinfo{person}{Giovanni Zappella}, \bibinfo{person}{Martin Wistuba}, \bibinfo{person}{Aditya Rawal}, {and} \bibinfo{person}{C\'edric Archambeau}.} \bibinfo{year}{2022}\natexlab{}.
\newblock \showarticletitle{Continual Learning With Transformers for Image Classification}. In \bibinfo{booktitle}{\emph{Proc. CVPR Workshops}}. \bibinfo{pages}{3774--3781}.
\newblock


\bibitem[Evans(2006)]%
        {evans2006scaling}
\bibfield{author}{\bibinfo{person}{Philip Evans}.} \bibinfo{year}{2006}\natexlab{}.
\newblock \showarticletitle{Scaling and assessment of data quality}.
\newblock \bibinfo{journal}{\emph{Acta Crystallographica Section D: Biological Crystallography}} \bibinfo{volume}{62}, \bibinfo{number}{1} (\bibinfo{year}{2006}), \bibinfo{pages}{72--82}.
\newblock


\bibitem[Fang et~al\mbox{.}(2020a)]%
        {fang2020fast}
\bibfield{author}{\bibinfo{person}{Jiemin Fang}, \bibinfo{person}{Yuzhu Sun}, \bibinfo{person}{Kangjian Peng}, \bibinfo{person}{Qian Zhang}, \bibinfo{person}{Yuan Li}, \bibinfo{person}{Wenyu Liu}, {and} \bibinfo{person}{Xinggang Wang}.} \bibinfo{year}{2020}\natexlab{a}.
\newblock \showarticletitle{Fast neural network adaptation via parameter remapping and architecture search}. In \bibinfo{booktitle}{\emph{Proc. ICLR}}.
\newblock


\bibitem[Fang et~al\mbox{.}(2020b)]%
        {fang2020fna++}
\bibfield{author}{\bibinfo{person}{Jiemin Fang}, \bibinfo{person}{Yuzhu Sun}, \bibinfo{person}{Qian Zhang}, \bibinfo{person}{Kangjian Peng}, \bibinfo{person}{Yuan Li}, \bibinfo{person}{Wenyu Liu}, {and} \bibinfo{person}{Xinggang Wang}.} \bibinfo{year}{2020}\natexlab{b}.
\newblock \showarticletitle{FNA++: Fast network adaptation via parameter remapping and architecture search}.
\newblock \bibinfo{journal}{\emph{IEEE Trans. Patt. Anal. Mach. Intell.}} \bibinfo{volume}{43}, \bibinfo{number}{9} (\bibinfo{year}{2020}), \bibinfo{pages}{2990--3004}.
\newblock


\bibitem[Feichtenhofer(2020)]%
        {feichtenhofer2020x3d}
\bibfield{author}{\bibinfo{person}{Christoph Feichtenhofer}.} \bibinfo{year}{2020}\natexlab{}.
\newblock \showarticletitle{X3d: Expanding architectures for efficient video recognition}. In \bibinfo{booktitle}{\emph{Proc. CVPR}}. \bibinfo{pages}{203--213}.
\newblock


\bibitem[Fu et~al\mbox{.}(2022)]%
        {fu2022adapterbias}
\bibfield{author}{\bibinfo{person}{Chin-Lun Fu}, \bibinfo{person}{Zih-Ching Chen}, \bibinfo{person}{Yun-Ru Lee}, {and} \bibinfo{person}{Hung-yi Lee}.} \bibinfo{year}{2022}\natexlab{}.
\newblock \showarticletitle{AdapterBias: Parameter-efficient Token-dependent Representation Shift for Adapters in NLP Tasks}.
\newblock \bibinfo{journal}{\emph{Preprint at https://arxiv.org/abs/2205.00305}} (\bibinfo{year}{2022}).
\newblock


\bibitem[Gan et~al\mbox{.}(2022)]%
        {gan2022decorate}
\bibfield{author}{\bibinfo{person}{Yulu Gan}, \bibinfo{person}{Xianzheng Ma}, \bibinfo{person}{Yihang Lou}, \bibinfo{person}{Yan Bai}, \bibinfo{person}{Renrui Zhang}, \bibinfo{person}{Nian Shi}, {and} \bibinfo{person}{Lin Luo}.} \bibinfo{year}{2022}\natexlab{}.
\newblock \showarticletitle{Decorate the Newcomers: Visual Domain Prompt for Continual Test Time Adaptation}.
\newblock \bibinfo{journal}{\emph{Preprint at https://arxiv.org/abs/2212.04145}} (\bibinfo{year}{2022}).
\newblock


\bibitem[Gao et~al\mbox{.}(2023)]%
        {gao2023compositional}
\bibfield{author}{\bibinfo{person}{Kaifeng Gao}, \bibinfo{person}{Long Chen}, \bibinfo{person}{Hanwang Zhang}, \bibinfo{person}{Jun Xiao}, {and} \bibinfo{person}{Qianru Sun}.} \bibinfo{year}{2023}\natexlab{}.
\newblock \showarticletitle{Compositional Prompt Tuning with Motion Cues for Open-vocabulary Video Relation Detection}.
\newblock \bibinfo{journal}{\emph{Preprint at https://arxiv.org/abs/2302.00268}} (\bibinfo{year}{2023}).
\newblock


\bibitem[Gao et~al\mbox{.}(2021)]%
        {gao2021clip}
\bibfield{author}{\bibinfo{person}{Peng Gao}, \bibinfo{person}{Shijie Geng}, \bibinfo{person}{Renrui Zhang}, \bibinfo{person}{Teli Ma}, \bibinfo{person}{Rongyao Fang}, \bibinfo{person}{Yongfeng Zhang}, \bibinfo{person}{Hongsheng Li}, {and} \bibinfo{person}{Yu Qiao}.} \bibinfo{year}{2021}\natexlab{}.
\newblock \showarticletitle{Clip-adapter: Better vision-language models with feature adapters}.
\newblock \bibinfo{journal}{\emph{Preprint at https://arxiv.org/abs/2110.04544}} (\bibinfo{year}{2021}).
\newblock


\bibitem[Gao et~al\mbox{.}(2022)]%
        {gao2022visual}
\bibfield{author}{\bibinfo{person}{Yunhe Gao}, \bibinfo{person}{Xingjian Shi}, \bibinfo{person}{Yi Zhu}, \bibinfo{person}{Hao Wang}, \bibinfo{person}{Zhiqiang Tang}, \bibinfo{person}{Xiong Zhou}, \bibinfo{person}{Mu Li}, {and} \bibinfo{person}{Dimitris~N Metaxas}.} \bibinfo{year}{2022}\natexlab{}.
\newblock \showarticletitle{Visual Prompt Tuning for Test-time Domain Adaptation}.
\newblock \bibinfo{journal}{\emph{Preprint at https://arxiv.org/abs/2210.04831}} (\bibinfo{year}{2022}).
\newblock


\bibitem[Gibson(2014)]%
        {gibson2014ecological}
\bibfield{author}{\bibinfo{person}{James~J Gibson}.} \bibinfo{year}{2014}\natexlab{}.
\newblock \bibinfo{booktitle}{\emph{The ecological approach to visual perception: classic edition}}.
\newblock \bibinfo{publisher}{Psychology press}.
\newblock


\bibitem[Gong et~al\mbox{.}(2014)]%
        {gong2014compressing}
\bibfield{author}{\bibinfo{person}{Yunchao Gong}, \bibinfo{person}{Liu Liu}, \bibinfo{person}{Ming Yang}, {and} \bibinfo{person}{Lubomir Bourdev}.} \bibinfo{year}{2014}\natexlab{}.
\newblock \showarticletitle{Compressing deep convolutional networks using vector quantization}.
\newblock \bibinfo{journal}{\emph{Preprint at https://arxiv.org/abs/1412.6115}} (\bibinfo{year}{2014}).
\newblock


\bibitem[Grassucci et~al\mbox{.}(2022)]%
        {grassucci2022phnns}
\bibfield{author}{\bibinfo{person}{Eleonora Grassucci}, \bibinfo{person}{Aston Zhang}, {and} \bibinfo{person}{Danilo Comminiello}.} \bibinfo{year}{2022}\natexlab{}.
\newblock \showarticletitle{PHNNs: Lightweight neural networks via parameterized hypercomplex convolutions}.
\newblock \bibinfo{journal}{\emph{IEEE Trans. Neur. Netw. Learn. Syst.}} (\bibinfo{year}{2022}).
\newblock


\bibitem[Gu et~al\mbox{.}(2021)]%
        {gu2021open}
\bibfield{author}{\bibinfo{person}{Xiuye Gu}, \bibinfo{person}{Tsung-Yi Lin}, \bibinfo{person}{Weicheng Kuo}, {and} \bibinfo{person}{Yin Cui}.} \bibinfo{year}{2021}\natexlab{}.
\newblock \showarticletitle{Open-vocabulary object detection via vision and language knowledge distillation}.
\newblock \bibinfo{journal}{\emph{Preprint at https://arxiv.org/abs/2104.13921}} (\bibinfo{year}{2021}).
\newblock


\bibitem[Guan et~al\mbox{.}(2020)]%
        {guan2020differentiable}
\bibfield{author}{\bibinfo{person}{Yushuo Guan}, \bibinfo{person}{Pengyu Zhao}, \bibinfo{person}{Bingxuan Wang}, \bibinfo{person}{Yuanxing Zhang}, \bibinfo{person}{Cong Yao}, \bibinfo{person}{Kaigui Bian}, {and} \bibinfo{person}{Jian Tang}.} \bibinfo{year}{2020}\natexlab{}.
\newblock \showarticletitle{Differentiable feature aggregation search for knowledge distillation}. In \bibinfo{booktitle}{\emph{ECCV 16}}. Springer, \bibinfo{pages}{469--484}.
\newblock


\bibitem[Guo et~al\mbox{.}(2021)]%
        {guo2021distilling}
\bibfield{author}{\bibinfo{person}{Jianyuan Guo}, \bibinfo{person}{Kai Han}, \bibinfo{person}{Yunhe Wang}, \bibinfo{person}{Han Wu}, \bibinfo{person}{Xinghao Chen}, \bibinfo{person}{Chunjing Xu}, {and} \bibinfo{person}{Chang Xu}.} \bibinfo{year}{2021}\natexlab{}.
\newblock \showarticletitle{Distilling object detectors via decoupled features}. In \bibinfo{booktitle}{\emph{Proc. CVPR}}. \bibinfo{pages}{2154--2164}.
\newblock


\bibitem[Guo et~al\mbox{.}(2022)]%
        {guo2022cmt}
\bibfield{author}{\bibinfo{person}{Jianyuan Guo}, \bibinfo{person}{Kai Han}, \bibinfo{person}{Han Wu}, \bibinfo{person}{Yehui Tang}, \bibinfo{person}{Xinghao Chen}, \bibinfo{person}{Yunhe Wang}, {and} \bibinfo{person}{Chang Xu}.} \bibinfo{year}{2022}\natexlab{}.
\newblock \showarticletitle{Cmt: Convolutional neural networks meet vision transformers}. In \bibinfo{booktitle}{\emph{Proc. CVPR}}. \bibinfo{pages}{12175--12185}.
\newblock


\bibitem[Han et~al\mbox{.}(2022)]%
        {han2022survey}
\bibfield{author}{\bibinfo{person}{Kai Han}, \bibinfo{person}{Yunhe Wang}, \bibinfo{person}{Hanting Chen}, \bibinfo{person}{Xinghao Chen}, \bibinfo{person}{Jianyuan Guo}, \bibinfo{person}{Zhenhua Liu}, \bibinfo{person}{Yehui Tang}, \bibinfo{person}{An Xiao}, \bibinfo{person}{Chunjing Xu}, \bibinfo{person}{Yixing Xu}, {et~al\mbox{.}}} \bibinfo{year}{2022}\natexlab{}.
\newblock \showarticletitle{A survey on vision transformer}.
\newblock \bibinfo{journal}{\emph{IEEE Trans. Patt. Anal. Mach. Intell.}} (\bibinfo{year}{2022}).
\newblock


\bibitem[Han et~al\mbox{.}(2021b)]%
        {han2021transformer}
\bibfield{author}{\bibinfo{person}{Kai Han}, \bibinfo{person}{An Xiao}, \bibinfo{person}{Enhua Wu}, \bibinfo{person}{Jianyuan Guo}, \bibinfo{person}{Chunjing Xu}, {and} \bibinfo{person}{Yunhe Wang}.} \bibinfo{year}{2021}\natexlab{b}.
\newblock \showarticletitle{Transformer in transformer}. In \bibinfo{booktitle}{\emph{Proc. NeurIPS}}, Vol.~\bibinfo{volume}{34}. \bibinfo{pages}{15908--15919}.
\newblock


\bibitem[Han et~al\mbox{.}(2021a)]%
        {han2021dynamic}
\bibfield{author}{\bibinfo{person}{Yizeng Han}, \bibinfo{person}{Gao Huang}, \bibinfo{person}{Shiji Song}, \bibinfo{person}{Le Yang}, \bibinfo{person}{Honghui Wang}, {and} \bibinfo{person}{Yulin Wang}.} \bibinfo{year}{2021}\natexlab{a}.
\newblock \showarticletitle{Dynamic neural networks: A survey}.
\newblock \bibinfo{journal}{\emph{IEEE Trans. Patt. Anal. Mach. Intell.}} (\bibinfo{year}{2021}).
\newblock


\bibitem[Hao et~al\mbox{.}(2023)]%
        {hao2023consolidator}
\bibfield{author}{\bibinfo{person}{Tianxiang Hao}, \bibinfo{person}{Hui Chen}, \bibinfo{person}{Yuchen Guo}, {and} \bibinfo{person}{Guiguang Ding}.} \bibinfo{year}{2023}\natexlab{}.
\newblock \showarticletitle{Consolidator: Mergable Adapter with Group Connections for Visual Adaptation}. In \bibinfo{booktitle}{\emph{Proc. ICLR}}.
\newblock
\urldef\tempurl%
\url{https://openreview.net/forum?id=J_Cja7cpgW}
\showURL{%
\tempurl}


\bibitem[Hao et~al\mbox{.}(2022)]%
        {jia2022efficient}
\bibfield{author}{\bibinfo{person}{Zhiwei Hao}, \bibinfo{person}{Jianyuan Guo}, \bibinfo{person}{Ding Jia}, \bibinfo{person}{Kai Han}, \bibinfo{person}{Yehui Tang}, \bibinfo{person}{Chao Zhang}, \bibinfo{person}{Han Hu}, {and} \bibinfo{person}{Yunhe Wang}.} \bibinfo{year}{2022}\natexlab{}.
\newblock \showarticletitle{Learning Efficient Vision Transformers via Fine-Grained Manifold Distillation}. In \bibinfo{booktitle}{\emph{Proc. NeurIPS}}.
\newblock


\bibitem[He et~al\mbox{.}(2020)]%
        {he2020milenas}
\bibfield{author}{\bibinfo{person}{Chaoyang He}, \bibinfo{person}{Haishan Ye}, \bibinfo{person}{Li Shen}, {and} \bibinfo{person}{Tong Zhang}.} \bibinfo{year}{2020}\natexlab{}.
\newblock \showarticletitle{Milenas: Efficient neural architecture search via mixed-level reformulation}. In \bibinfo{booktitle}{\emph{Proc. CVPR}}. \bibinfo{pages}{11993--12002}.
\newblock


\bibitem[He et~al\mbox{.}(2022d)]%
        {he2022towards}
\bibfield{author}{\bibinfo{person}{Junxian He}, \bibinfo{person}{Chunting Zhou}, \bibinfo{person}{Xuezhe Ma}, \bibinfo{person}{Taylor Berg-Kirkpatrick}, {and} \bibinfo{person}{Graham Neubig}.} \bibinfo{year}{2022}\natexlab{d}.
\newblock \showarticletitle{Towards a Unified View of Parameter-Efficient Transfer Learning}. In \bibinfo{booktitle}{\emph{Proc. ICLR}}.
\newblock
\urldef\tempurl%
\url{https://openreview.net/forum?id=0RDcd5Axok}
\showURL{%
\tempurl}


\bibitem[He et~al\mbox{.}(2022a)]%
        {he2022masked}
\bibfield{author}{\bibinfo{person}{Kaiming He}, \bibinfo{person}{Xinlei Chen}, \bibinfo{person}{Saining Xie}, \bibinfo{person}{Yanghao Li}, \bibinfo{person}{Piotr Doll{\'a}r}, {and} \bibinfo{person}{Ross Girshick}.} \bibinfo{year}{2022}\natexlab{a}.
\newblock \showarticletitle{Masked autoencoders are scalable vision learners}. In \bibinfo{booktitle}{\emph{Proc. CVPR}}. \bibinfo{pages}{16000--16009}.
\newblock


\bibitem[He et~al\mbox{.}(2016)]%
        {he2016deep}
\bibfield{author}{\bibinfo{person}{Kaiming He}, \bibinfo{person}{Xiangyu Zhang}, \bibinfo{person}{Shaoqing Ren}, {and} \bibinfo{person}{Jian Sun}.} \bibinfo{year}{2016}\natexlab{}.
\newblock \showarticletitle{Deep residual learning for image recognition}. In \bibinfo{booktitle}{\emph{Proc. CVPR}}. \bibinfo{pages}{770--778}.
\newblock


\bibitem[He et~al\mbox{.}(2022b)]%
        {he2022parameter}
\bibfield{author}{\bibinfo{person}{Xuehai He}, \bibinfo{person}{Chunyuan Li}, \bibinfo{person}{Pengchuan Zhang}, \bibinfo{person}{Jianwei Yang}, {and} \bibinfo{person}{Xin~Eric Wang}.} \bibinfo{year}{2022}\natexlab{b}.
\newblock \showarticletitle{Parameter-efficient fine-tuning for vision transformers}.
\newblock \bibinfo{journal}{\emph{Preprint at https://arxiv.org/abs/2203.16329}} (\bibinfo{year}{2022}).
\newblock


\bibitem[He et~al\mbox{.}(2022c)]%
        {he2022cpl}
\bibfield{author}{\bibinfo{person}{Xuehai He}, \bibinfo{person}{Diji Yang}, \bibinfo{person}{Weixi Feng}, \bibinfo{person}{Tsu-Jui Fu}, \bibinfo{person}{Arjun Akula}, \bibinfo{person}{Varun Jampani}, \bibinfo{person}{Pradyumna Narayana}, \bibinfo{person}{Sugato Basu}, \bibinfo{person}{William~Yang Wang}, {and} \bibinfo{person}{Xin~Eric Wang}.} \bibinfo{year}{2022}\natexlab{c}.
\newblock \showarticletitle{CPL: Counterfactual Prompt Learning for Vision and Language Models}.
\newblock \bibinfo{journal}{\emph{Preprint at https://arxiv.org/abs/2210.10362}} (\bibinfo{year}{2022}).
\newblock


\bibitem[Herzig et~al\mbox{.}(2022)]%
        {herzig2022promptonomyvit}
\bibfield{author}{\bibinfo{person}{Roei Herzig}, \bibinfo{person}{Ofir Abramovich}, \bibinfo{person}{Elad Ben-Avraham}, \bibinfo{person}{Assaf Arbelle}, \bibinfo{person}{Leonid Karlinsky}, \bibinfo{person}{Ariel Shamir}, \bibinfo{person}{Trevor Darrell}, {and} \bibinfo{person}{Amir Globerson}.} \bibinfo{year}{2022}\natexlab{}.
\newblock \showarticletitle{PromptonomyViT: Multi-Task Prompt Learning Improves Video Transformers using Synthetic Scene Data}.
\newblock \bibinfo{journal}{\emph{Preprint at https://arxiv.org/abs/2212.04821}} (\bibinfo{year}{2022}).
\newblock


\bibitem[Hinton et~al\mbox{.}(2015)]%
        {hinton2015distilling}
\bibfield{author}{\bibinfo{person}{Geoffrey Hinton}, \bibinfo{person}{Oriol Vinyals}, {and} \bibinfo{person}{Jeff Dean}.} \bibinfo{year}{2015}\natexlab{}.
\newblock \showarticletitle{Distilling the knowledge in a neural network}.
\newblock \bibinfo{journal}{\emph{Preprint at https://arxiv.org/abs/1503.02531}} (\bibinfo{year}{2015}).
\newblock


\bibitem[Hou et~al\mbox{.}(2022)]%
        {hou2022meta}
\bibfield{author}{\bibinfo{person}{Zejiang Hou}, \bibinfo{person}{Julian Salazar}, {and} \bibinfo{person}{George Polovets}.} \bibinfo{year}{2022}\natexlab{}.
\newblock \showarticletitle{Meta-learning the difference: Preparing large language models for efficient adaptation}.
\newblock \bibinfo{journal}{\emph{Transactions of the Association for Computational Linguistics}}  \bibinfo{volume}{10} (\bibinfo{year}{2022}), \bibinfo{pages}{1249--1265}.
\newblock


\bibitem[Houlsby et~al\mbox{.}(2019)]%
        {houlsby2019parameter}
\bibfield{author}{\bibinfo{person}{Neil Houlsby}, \bibinfo{person}{Andrei Giurgiu}, \bibinfo{person}{Stanislaw Jastrzebski}, \bibinfo{person}{Bruna Morrone}, \bibinfo{person}{Quentin De~Laroussilhe}, \bibinfo{person}{Andrea Gesmundo}, \bibinfo{person}{Mona Attariyan}, {and} \bibinfo{person}{Sylvain Gelly}.} \bibinfo{year}{2019}\natexlab{}.
\newblock \showarticletitle{Parameter-efficient transfer learning for NLP}. In \bibinfo{booktitle}{\emph{Proc. ICML}}. PMLR, \bibinfo{pages}{2790--2799}.
\newblock


\bibitem[Hu et~al\mbox{.}(2022b)]%
        {hu2022lora}
\bibfield{author}{\bibinfo{person}{Edward~J Hu}, \bibinfo{person}{yelong shen}, \bibinfo{person}{Phillip Wallis}, \bibinfo{person}{Zeyuan Allen-Zhu}, \bibinfo{person}{Yuanzhi Li}, \bibinfo{person}{Shean Wang}, \bibinfo{person}{Lu Wang}, {and} \bibinfo{person}{Weizhu Chen}.} \bibinfo{year}{2022}\natexlab{b}.
\newblock \showarticletitle{Lo{RA}: Low-Rank Adaptation of Large Language Models}. In \bibinfo{booktitle}{\emph{Proc. ICLR}}.
\newblock
\urldef\tempurl%
\url{https://openreview.net/forum?id=nZeVKeeFYf9}
\showURL{%
\tempurl}


\bibitem[Hu et~al\mbox{.}(2021)]%
        {hu2021boosting}
\bibfield{author}{\bibinfo{person}{Junjie Hu}, \bibinfo{person}{Chenyou Fan}, \bibinfo{person}{Hualie Jiang}, \bibinfo{person}{Xiyue Guo}, \bibinfo{person}{Yuan Gao}, \bibinfo{person}{Xiangyong Lu}, {and} \bibinfo{person}{Tin~Lun Lam}.} \bibinfo{year}{2021}\natexlab{}.
\newblock \showarticletitle{Boosting light-weight depth estimation via knowledge distillation}.
\newblock \bibinfo{journal}{\emph{Preprint at https://arxiv.org/abs/2105.06143}} (\bibinfo{year}{2021}).
\newblock


\bibitem[Hu et~al\mbox{.}(2022a)]%
        {hu2022prosfda}
\bibfield{author}{\bibinfo{person}{Shishuai Hu}, \bibinfo{person}{Zehui Liao}, {and} \bibinfo{person}{Yong Xia}.} \bibinfo{year}{2022}\natexlab{a}.
\newblock \showarticletitle{ProSFDA: Prompt Learning based Source-free Domain Adaptation for Medical Image Segmentation}.
\newblock \bibinfo{journal}{\emph{Preprint at https://arxiv.org/abs/2211.11514}} (\bibinfo{year}{2022}).
\newblock


\bibitem[Huang et~al\mbox{.}(2021)]%
        {huang2021shuffle}
\bibfield{author}{\bibinfo{person}{Zilong Huang}, \bibinfo{person}{Youcheng Ben}, \bibinfo{person}{Guozhong Luo}, \bibinfo{person}{Pei Cheng}, \bibinfo{person}{Gang Yu}, {and} \bibinfo{person}{Bin Fu}.} \bibinfo{year}{2021}\natexlab{}.
\newblock \showarticletitle{Shuffle transformer: Rethinking spatial shuffle for vision transformer}.
\newblock \bibinfo{journal}{\emph{Preprint at https://arxiv.org/abs/2106.03650}} (\bibinfo{year}{2021}).
\newblock


\bibitem[Hubel and Wiesel(1959)]%
        {hubel1959receptive}
\bibfield{author}{\bibinfo{person}{David~H Hubel} {and} \bibinfo{person}{Torsten~N Wiesel}.} \bibinfo{year}{1959}\natexlab{}.
\newblock \showarticletitle{Receptive fields of single neurones in the cat's striate cortex}.
\newblock \bibinfo{journal}{\emph{The Journal of physiology}} \bibinfo{volume}{148}, \bibinfo{number}{3} (\bibinfo{year}{1959}), \bibinfo{pages}{574}.
\newblock


\bibitem[Ionescu et~al\mbox{.}(2013)]%
        {ionescu2013human3}
\bibfield{author}{\bibinfo{person}{Catalin Ionescu}, \bibinfo{person}{Dragos Papava}, \bibinfo{person}{Vlad Olaru}, {and} \bibinfo{person}{Cristian Sminchisescu}.} \bibinfo{year}{2013}\natexlab{}.
\newblock \showarticletitle{Human3. 6m: Large scale datasets and predictive methods for 3d human sensing in natural environments}.
\newblock \bibinfo{journal}{\emph{IEEE Trans. Patt. Anal. Mach. Intell.}} \bibinfo{volume}{36}, \bibinfo{number}{7} (\bibinfo{year}{2013}), \bibinfo{pages}{1325--1339}.
\newblock


\bibitem[Jia et~al\mbox{.}(2021)]%
        {jia2021scaling}
\bibfield{author}{\bibinfo{person}{Chao Jia}, \bibinfo{person}{Yinfei Yang}, \bibinfo{person}{Ye Xia}, \bibinfo{person}{Yi-Ting Chen}, \bibinfo{person}{Zarana Parekh}, \bibinfo{person}{Hieu Pham}, \bibinfo{person}{Quoc Le}, \bibinfo{person}{Yun-Hsuan Sung}, \bibinfo{person}{Zhen Li}, {and} \bibinfo{person}{Tom Duerig}.} \bibinfo{year}{2021}\natexlab{}.
\newblock \showarticletitle{Scaling up visual and vision-language representation learning with noisy text supervision}. In \bibinfo{booktitle}{\emph{Proc. ICML}}. PMLR, \bibinfo{pages}{4904--4916}.
\newblock


\bibitem[Jia et~al\mbox{.}(2022)]%
        {jia2022visual}
\bibfield{author}{\bibinfo{person}{Menglin Jia}, \bibinfo{person}{Luming Tang}, \bibinfo{person}{Bor-Chun Chen}, \bibinfo{person}{Claire Cardie}, \bibinfo{person}{Serge Belongie}, \bibinfo{person}{Bharath Hariharan}, {and} \bibinfo{person}{Ser-Nam Lim}.} \bibinfo{year}{2022}\natexlab{}.
\newblock \showarticletitle{Visual prompt tuning}. In \bibinfo{booktitle}{\emph{Proc. ECCV}}. Springer, \bibinfo{pages}{709--727}.
\newblock


\bibitem[Jiang et~al\mbox{.}(2022c)]%
        {jiang2022cross}
\bibfield{author}{\bibinfo{person}{Haojun Jiang}, \bibinfo{person}{Jianke Zhang}, \bibinfo{person}{Rui Huang}, \bibinfo{person}{Chunjiang Ge}, \bibinfo{person}{Zanlin Ni}, \bibinfo{person}{Jiwen Lu}, \bibinfo{person}{Jie Zhou}, \bibinfo{person}{Shiji Song}, {and} \bibinfo{person}{Gao Huang}.} \bibinfo{year}{2022}\natexlab{c}.
\newblock \showarticletitle{Cross-Modal Adapter for Text-Video Retrieval}.
\newblock \bibinfo{journal}{\emph{Preprint at https://arxiv.org/abs/2211.09623}} (\bibinfo{year}{2022}).
\newblock


\bibitem[Jiang et~al\mbox{.}(2022a)]%
        {jiang2022dna}
\bibfield{author}{\bibinfo{person}{Ziyu Jiang}, \bibinfo{person}{Tianlong Chen}, \bibinfo{person}{Xuxi Chen}, \bibinfo{person}{Yu Cheng}, \bibinfo{person}{Luowei Zhou}, \bibinfo{person}{Lu Yuan}, \bibinfo{person}{Ahmed Awadallah}, {and} \bibinfo{person}{Zhangyang Wang}.} \bibinfo{year}{2022}\natexlab{a}.
\newblock \showarticletitle{DnA: Improving Few-Shot Transfer Learning with Low-Rank Decomposition and Alignment}. In \bibinfo{booktitle}{\emph{Proc. ECCV}}. Springer, \bibinfo{pages}{239--256}.
\newblock


\bibitem[Jiang et~al\mbox{.}(2022b)]%
        {jiangback}
\bibfield{author}{\bibinfo{person}{Ziyu Jiang}, \bibinfo{person}{Xuxi Chen}, \bibinfo{person}{Xueqin Huang}, \bibinfo{person}{Xianzhi Du}, \bibinfo{person}{Denny Zhou}, {and} \bibinfo{person}{Zhangyang Wang}.} \bibinfo{year}{2022}\natexlab{b}.
\newblock \showarticletitle{Back Razor: Memory-Efficient Transfer Learning by Self-Sparsified Backpropagation}. In \bibinfo{booktitle}{\emph{Proc. NeurIPS}}.
\newblock


\bibitem[Jie and Deng(2022a)]%
        {jie2022convolutional}
\bibfield{author}{\bibinfo{person}{Shibo Jie} {and} \bibinfo{person}{Zhi-Hong Deng}.} \bibinfo{year}{2022}\natexlab{a}.
\newblock \showarticletitle{Convolutional bypasses are better vision transformer adapters}.
\newblock \bibinfo{journal}{\emph{Preprint at https://arxiv.org/abs/2207.07039}} (\bibinfo{year}{2022}).
\newblock


\bibitem[Jie and Deng(2022b)]%
        {jie2022fact}
\bibfield{author}{\bibinfo{person}{Shibo Jie} {and} \bibinfo{person}{Zhi-Hong Deng}.} \bibinfo{year}{2022}\natexlab{b}.
\newblock \showarticletitle{FacT: Factor-Tuning for Lightweight Adaptation on Vision Transformer}.
\newblock \bibinfo{journal}{\emph{Preprint at https://arxiv.org/abs/2212.03145}} (\bibinfo{year}{2022}).
\newblock


\bibitem[Ju et~al\mbox{.}(2022)]%
        {ju2022prompting}
\bibfield{author}{\bibinfo{person}{Chen Ju}, \bibinfo{person}{Tengda Han}, \bibinfo{person}{Kunhao Zheng}, \bibinfo{person}{Ya Zhang}, {and} \bibinfo{person}{Weidi Xie}.} \bibinfo{year}{2022}\natexlab{}.
\newblock \showarticletitle{Prompting visual-language models for efficient video understanding}. In \bibinfo{booktitle}{\emph{Proc. ECCV}}. Springer, \bibinfo{pages}{105--124}.
\newblock


\bibitem[Kaissis et~al\mbox{.}(2020)]%
        {kaissis2020secure}
\bibfield{author}{\bibinfo{person}{Georgios~A Kaissis}, \bibinfo{person}{Marcus~R Makowski}, \bibinfo{person}{Daniel R{\"u}ckert}, {and} \bibinfo{person}{Rickmer~F Braren}.} \bibinfo{year}{2020}\natexlab{}.
\newblock \showarticletitle{Secure, privacy-preserving and federated machine learning in medical imaging}.
\newblock \bibinfo{journal}{\emph{Nature Machine Intelligence}} \bibinfo{volume}{2}, \bibinfo{number}{6} (\bibinfo{year}{2020}), \bibinfo{pages}{305--311}.
\newblock


\bibitem[Karimi~Mahabadi et~al\mbox{.}(2021)]%
        {karimi2021compacter}
\bibfield{author}{\bibinfo{person}{Rabeeh Karimi~Mahabadi}, \bibinfo{person}{James Henderson}, {and} \bibinfo{person}{Sebastian Ruder}.} \bibinfo{year}{2021}\natexlab{}.
\newblock \showarticletitle{Compacter: Efficient low-rank hypercomplex adapter layers}. In \bibinfo{booktitle}{\emph{Proc. NeurIPS}}, Vol.~\bibinfo{volume}{34}. \bibinfo{pages}{1022--1035}.
\newblock


\bibitem[Kay et~al\mbox{.}(2017)]%
        {kay2017kinetics}
\bibfield{author}{\bibinfo{person}{Will Kay}, \bibinfo{person}{Joao Carreira}, \bibinfo{person}{Karen Simonyan}, \bibinfo{person}{Brian Zhang}, \bibinfo{person}{Chloe Hillier}, \bibinfo{person}{Sudheendra Vijayanarasimhan}, \bibinfo{person}{Fabio Viola}, \bibinfo{person}{Tim Green}, \bibinfo{person}{Trevor Back}, \bibinfo{person}{Paul Natsev}, {et~al\mbox{.}}} \bibinfo{year}{2017}\natexlab{}.
\newblock \showarticletitle{The kinetics human action video dataset}.
\newblock \bibinfo{journal}{\emph{Preprint at https://arxiv.org/abs/1705.06950}} (\bibinfo{year}{2017}).
\newblock


\bibitem[Ke and Sukthankar(2004)]%
        {ke2004pca}
\bibfield{author}{\bibinfo{person}{Yan Ke} {and} \bibinfo{person}{Rahul Sukthankar}.} \bibinfo{year}{2004}\natexlab{}.
\newblock \showarticletitle{PCA-SIFT: A more distinctive representation for local image descriptors}. In \bibinfo{booktitle}{\emph{Proc. CVPR}}, Vol.~\bibinfo{volume}{2}. IEEE, \bibinfo{pages}{II--II}.
\newblock


\bibitem[Khan et~al\mbox{.}(2022)]%
        {khan2022transformers}
\bibfield{author}{\bibinfo{person}{Salman Khan}, \bibinfo{person}{Muzammal Naseer}, \bibinfo{person}{Munawar Hayat}, \bibinfo{person}{Syed~Waqas Zamir}, \bibinfo{person}{Fahad~Shahbaz Khan}, {and} \bibinfo{person}{Mubarak Shah}.} \bibinfo{year}{2022}\natexlab{}.
\newblock \showarticletitle{Transformers in vision: A survey}.
\newblock \bibinfo{journal}{\emph{ACM computing surveys (CSUR)}} \bibinfo{volume}{54}, \bibinfo{number}{10s} (\bibinfo{year}{2022}), \bibinfo{pages}{1--41}.
\newblock


\bibitem[Khattak et~al\mbox{.}(2022)]%
        {khattak2022maple}
\bibfield{author}{\bibinfo{person}{Muhammad~Uzair Khattak}, \bibinfo{person}{Hanoona Rasheed}, \bibinfo{person}{Muhammad Maaz}, \bibinfo{person}{Salman Khan}, {and} \bibinfo{person}{Fahad~Shahbaz Khan}.} \bibinfo{year}{2022}\natexlab{}.
\newblock \showarticletitle{Maple: Multi-modal prompt learning}.
\newblock \bibinfo{journal}{\emph{Preprint at https://arxiv.org/abs/2210.03117}} (\bibinfo{year}{2022}).
\newblock


\bibitem[Kim et~al\mbox{.}(2018)]%
        {kim2018paraphrasing}
\bibfield{author}{\bibinfo{person}{Jangho Kim}, \bibinfo{person}{SeongUk Park}, {and} \bibinfo{person}{Nojun Kwak}.} \bibinfo{year}{2018}\natexlab{}.
\newblock \showarticletitle{Paraphrasing complex network: Network compression via factor transfer}. In \bibinfo{booktitle}{\emph{Proc. NeurIPS}}, Vol.~\bibinfo{volume}{31}.
\newblock


\bibitem[Kim et~al\mbox{.}(2021)]%
        {kim2021adapt}
\bibfield{author}{\bibinfo{person}{Konwoo Kim}, \bibinfo{person}{Michael Laskin}, \bibinfo{person}{Igor Mordatch}, {and} \bibinfo{person}{Deepak Pathak}.} \bibinfo{year}{2021}\natexlab{}.
\newblock \showarticletitle{How to Adapt Your Large-Scale Vision-and-Language Model}.
\newblock \bibinfo{journal}{\emph{Preprint at https://openreview.net/forum?id=EhwEUb2ynIa}} (\bibinfo{year}{2021}).
\newblock


\bibitem[Kim et~al\mbox{.}(2023b)]%
        {kim2023zegot}
\bibfield{author}{\bibinfo{person}{Kwanyoung Kim}, \bibinfo{person}{Yujin Oh}, {and} \bibinfo{person}{Jong~Chul Ye}.} \bibinfo{year}{2023}\natexlab{b}.
\newblock \showarticletitle{ZegOT: Zero-shot Segmentation Through Optimal Transport of Text Prompts}.
\newblock \bibinfo{journal}{\emph{Preprint at https://arxiv.org/abs/2301.12171}} (\bibinfo{year}{2023}).
\newblock


\bibitem[Kim et~al\mbox{.}(2023a)]%
        {kim2023prompt}
\bibfield{author}{\bibinfo{person}{Minsu Kim}, \bibinfo{person}{Hyung-Il Kim}, {and} \bibinfo{person}{Yong~Man Ro}.} \bibinfo{year}{2023}\natexlab{a}.
\newblock \showarticletitle{Prompt Tuning of Deep Neural Networks for Speaker-adaptive Visual Speech Recognition}.
\newblock \bibinfo{journal}{\emph{Preprint at https://arxiv.org/abs/2302.08102}} (\bibinfo{year}{2023}).
\newblock


\bibitem[Kothari et~al\mbox{.}(2022)]%
        {kothari2022motion}
\bibfield{author}{\bibinfo{person}{Parth Kothari}, \bibinfo{person}{Danya Li}, \bibinfo{person}{Yuejiang Liu}, {and} \bibinfo{person}{Alexandre Alahi}.} \bibinfo{year}{2022}\natexlab{}.
\newblock \showarticletitle{Motion style transfer: Modular low-rank adaptation for deep motion forecasting}.
\newblock \bibinfo{journal}{\emph{Preprint at https://arxiv.org/abs/2211.03165}} (\bibinfo{year}{2022}).
\newblock


\bibitem[Krizhevsky et~al\mbox{.}(2017)]%
        {krizhevsky2017imagenet}
\bibfield{author}{\bibinfo{person}{Alex Krizhevsky}, \bibinfo{person}{Ilya Sutskever}, {and} \bibinfo{person}{Geoffrey~E Hinton}.} \bibinfo{year}{2017}\natexlab{}.
\newblock \showarticletitle{Imagenet classification with deep convolutional neural networks}.
\newblock \bibinfo{journal}{\emph{Commun. ACM}} \bibinfo{volume}{60}, \bibinfo{number}{6} (\bibinfo{year}{2017}), \bibinfo{pages}{84--90}.
\newblock


\bibitem[Kumar et~al\mbox{.}(2021)]%
        {kumar2021fine}
\bibfield{author}{\bibinfo{person}{Ananya Kumar}, \bibinfo{person}{Aditi Raghunathan}, \bibinfo{person}{Robbie~Matthew Jones}, \bibinfo{person}{Tengyu Ma}, {and} \bibinfo{person}{Percy Liang}.} \bibinfo{year}{2021}\natexlab{}.
\newblock \showarticletitle{Fine-Tuning can Distort Pretrained Features and Underperform Out-of-Distribution}. In \bibinfo{booktitle}{\emph{Proc. ICLR}}.
\newblock


\bibitem[Lake et~al\mbox{.}(2015)]%
        {lake2015human}
\bibfield{author}{\bibinfo{person}{Brenden~M Lake}, \bibinfo{person}{Ruslan Salakhutdinov}, {and} \bibinfo{person}{Joshua~B Tenenbaum}.} \bibinfo{year}{2015}\natexlab{}.
\newblock \showarticletitle{Human-level concept learning through probabilistic program induction}.
\newblock \bibinfo{journal}{\emph{Science}} \bibinfo{volume}{350}, \bibinfo{number}{6266} (\bibinfo{year}{2015}), \bibinfo{pages}{1332--1338}.
\newblock


\bibitem[Lavin and Gray(2016)]%
        {lavin2016fast}
\bibfield{author}{\bibinfo{person}{Andrew Lavin} {and} \bibinfo{person}{Scott Gray}.} \bibinfo{year}{2016}\natexlab{}.
\newblock \showarticletitle{Fast algorithms for convolutional neural networks}. In \bibinfo{booktitle}{\emph{Proc. CVPR}}. \bibinfo{pages}{4013--4021}.
\newblock


\bibitem[LeCun et~al\mbox{.}(2015)]%
        {lecun2015deep}
\bibfield{author}{\bibinfo{person}{Yann LeCun}, \bibinfo{person}{Yoshua Bengio}, {and} \bibinfo{person}{Geoffrey Hinton}.} \bibinfo{year}{2015}\natexlab{}.
\newblock \showarticletitle{Deep learning}.
\newblock \bibinfo{journal}{\emph{nature}} \bibinfo{volume}{521}, \bibinfo{number}{7553} (\bibinfo{year}{2015}), \bibinfo{pages}{436--444}.
\newblock


\bibitem[Li et~al\mbox{.}(2022c)]%
        {li2021efficient}
\bibfield{author}{\bibinfo{person}{Chunyuan Li}, \bibinfo{person}{Jianwei Yang}, \bibinfo{person}{Pengchuan Zhang}, \bibinfo{person}{Mei Gao}, \bibinfo{person}{Bin Xiao}, \bibinfo{person}{Xiyang Dai}, \bibinfo{person}{Lu Yuan}, {and} \bibinfo{person}{Jianfeng Gao}.} \bibinfo{year}{2022}\natexlab{c}.
\newblock \showarticletitle{Efficient self-supervised vision transformers for representation learning}. In \bibinfo{booktitle}{\emph{Proc. ICLR}}.
\newblock


\bibitem[Li et~al\mbox{.}(2020)]%
        {li2020sgas}
\bibfield{author}{\bibinfo{person}{Guohao Li}, \bibinfo{person}{Guocheng Qian}, \bibinfo{person}{Itzel~C Delgadillo}, \bibinfo{person}{Matthias Muller}, \bibinfo{person}{Ali Thabet}, {and} \bibinfo{person}{Bernard Ghanem}.} \bibinfo{year}{2020}\natexlab{}.
\newblock \showarticletitle{Sgas: Sequential greedy architecture search}. In \bibinfo{booktitle}{\emph{Proc. CVPR}}. \bibinfo{pages}{1620--1630}.
\newblock


\bibitem[Li et~al\mbox{.}(2017)]%
        {li2017mimicking}
\bibfield{author}{\bibinfo{person}{Quanquan Li}, \bibinfo{person}{Shengying Jin}, {and} \bibinfo{person}{Junjie Yan}.} \bibinfo{year}{2017}\natexlab{}.
\newblock \showarticletitle{Mimicking very efficient network for object detection}. In \bibinfo{booktitle}{\emph{Proc. CVPR}}. \bibinfo{pages}{6356--6364}.
\newblock


\bibitem[Li et~al\mbox{.}(2021a)]%
        {li2021else}
\bibfield{author}{\bibinfo{person}{Tianjiao Li}, \bibinfo{person}{Qiuhong Ke}, \bibinfo{person}{Hossein Rahmani}, \bibinfo{person}{Rui~En Ho}, \bibinfo{person}{Henghui Ding}, {and} \bibinfo{person}{Jun Liu}.} \bibinfo{year}{2021}\natexlab{a}.
\newblock \showarticletitle{Else-net: Elastic semantic network for continual action recognition from skeleton data}. In \bibinfo{booktitle}{\emph{Proc. CVPR}}. \bibinfo{pages}{13434--13443}.
\newblock


\bibitem[Li et~al\mbox{.}(2022b)]%
        {li2022low}
\bibfield{author}{\bibinfo{person}{Tao Li}, \bibinfo{person}{Lei Tan}, \bibinfo{person}{Zhehao Huang}, \bibinfo{person}{Qinghua Tao}, \bibinfo{person}{Yipeng Liu}, {and} \bibinfo{person}{Xiaolin Huang}.} \bibinfo{year}{2022}\natexlab{b}.
\newblock \showarticletitle{Low dimensional trajectory hypothesis is true: Dnns can be trained in tiny subspaces}.
\newblock \bibinfo{journal}{\emph{IEEE Trans. Patt. Anal. Mach. Intell.}} (\bibinfo{year}{2022}).
\newblock


\bibitem[Li et~al\mbox{.}(2022a)]%
        {li2022exploring}
\bibfield{author}{\bibinfo{person}{Yanghao Li}, \bibinfo{person}{Hanzi Mao}, \bibinfo{person}{Ross Girshick}, {and} \bibinfo{person}{Kaiming He}.} \bibinfo{year}{2022}\natexlab{a}.
\newblock \showarticletitle{Exploring plain vision transformer backbones for object detection}. In \bibinfo{booktitle}{\emph{Proc. ECCV}}. Springer, \bibinfo{pages}{280--296}.
\newblock


\bibitem[Li et~al\mbox{.}(2021c)]%
        {li2021benchmarking}
\bibfield{author}{\bibinfo{person}{Yanghao Li}, \bibinfo{person}{Saining Xie}, \bibinfo{person}{Xinlei Chen}, \bibinfo{person}{Piotr Dollar}, \bibinfo{person}{Kaiming He}, {and} \bibinfo{person}{Ross Girshick}.} \bibinfo{year}{2021}\natexlab{c}.
\newblock \showarticletitle{Benchmarking detection transfer learning with vision transformers}.
\newblock \bibinfo{journal}{\emph{Preprint at https://arxiv.org/abs/2111.11429}} (\bibinfo{year}{2021}).
\newblock


\bibitem[Li et~al\mbox{.}(2021b)]%
        {li2021survey}
\bibfield{author}{\bibinfo{person}{Zewen Li}, \bibinfo{person}{Fan Liu}, \bibinfo{person}{Wenjie Yang}, \bibinfo{person}{Shouheng Peng}, {and} \bibinfo{person}{Jun Zhou}.} \bibinfo{year}{2021}\natexlab{b}.
\newblock \showarticletitle{A survey of convolutional neural networks: analysis, applications, and prospects}.
\newblock \bibinfo{journal}{\emph{IEEE Trans. Neur. Netw. Learn. Syst.}} (\bibinfo{year}{2021}).
\newblock


\bibitem[Lian et~al\mbox{.}(2022)]%
        {lian2022scaling}
\bibfield{author}{\bibinfo{person}{Dongze Lian}, \bibinfo{person}{Zhou Daquan}, \bibinfo{person}{Jiashi Feng}, {and} \bibinfo{person}{Xinchao Wang}.} \bibinfo{year}{2022}\natexlab{}.
\newblock \showarticletitle{Scaling \& Shifting Your Features: A New Baseline for Efficient Model Tuning}. In \bibinfo{booktitle}{\emph{Proc. NeurIPS}}, \bibfield{editor}{\bibinfo{person}{Alice~H. Oh}, \bibinfo{person}{Alekh Agarwal}, \bibinfo{person}{Danielle Belgrave}, {and} \bibinfo{person}{Kyunghyun Cho}} (Eds.).
\newblock
\urldef\tempurl%
\url{https://openreview.net/forum?id=XtyeppctGgc}
\showURL{%
\tempurl}


\bibitem[Liang et~al\mbox{.}(2019)]%
        {liang2019darts+}
\bibfield{author}{\bibinfo{person}{Hanwen Liang}, \bibinfo{person}{Shifeng Zhang}, \bibinfo{person}{Jiacheng Sun}, \bibinfo{person}{Xingqiu He}, \bibinfo{person}{Weiran Huang}, \bibinfo{person}{Kechen Zhuang}, {and} \bibinfo{person}{Zhenguo Li}.} \bibinfo{year}{2019}\natexlab{}.
\newblock \showarticletitle{Darts+: Improved differentiable architecture search with early stopping}.
\newblock \bibinfo{journal}{\emph{Preprint at https://arxiv.org/abs/1909.06035}} (\bibinfo{year}{2019}).
\newblock


\bibitem[Lin et~al\mbox{.}(2023)]%
        {lin2023comes}
\bibfield{author}{\bibinfo{person}{Junfan Lin}, \bibinfo{person}{Jianlong Chang}, \bibinfo{person}{Lingbo Liu}, \bibinfo{person}{Guanbin Li}, \bibinfo{person}{Liang Lin}, \bibinfo{person}{Qi Tian}, {and} \bibinfo{person}{Chang-wen Chen}.} \bibinfo{year}{2023}\natexlab{}.
\newblock \showarticletitle{Being Comes from Not-being: Open-vocabulary Text-to-Motion Generation with Wordless Training}. In \bibinfo{booktitle}{\emph{Proc. CVPR}}.
\newblock


\bibitem[Lin et~al\mbox{.}(2014)]%
        {lin2014microsoft}
\bibfield{author}{\bibinfo{person}{Tsung-Yi Lin}, \bibinfo{person}{Michael Maire}, \bibinfo{person}{Serge Belongie}, \bibinfo{person}{James Hays}, \bibinfo{person}{Pietro Perona}, \bibinfo{person}{Deva Ramanan}, \bibinfo{person}{Piotr Doll{\'a}r}, {and} \bibinfo{person}{C~Lawrence Zitnick}.} \bibinfo{year}{2014}\natexlab{}.
\newblock \showarticletitle{Microsoft coco: Common objects in context}. In \bibinfo{booktitle}{\emph{Proc. ECCV}}. Springer, \bibinfo{pages}{740--755}.
\newblock


\bibitem[Lin et~al\mbox{.}(2022)]%
        {lin2022vision}
\bibfield{author}{\bibinfo{person}{Yan-Bo Lin}, \bibinfo{person}{Yi-Lin Sung}, \bibinfo{person}{Jie Lei}, \bibinfo{person}{Mohit Bansal}, {and} \bibinfo{person}{Gedas Bertasius}.} \bibinfo{year}{2022}\natexlab{}.
\newblock \showarticletitle{Vision Transformers are Parameter-Efficient Audio-Visual Learners}.
\newblock \bibinfo{journal}{\emph{Preprint at https://arxiv.org/abs/2212.07983}} (\bibinfo{year}{2022}).
\newblock


\bibitem[Lin et~al\mbox{.}(2020)]%
        {lin2020exploring}
\bibfield{author}{\bibinfo{person}{Zhaojiang Lin}, \bibinfo{person}{Andrea Madotto}, {and} \bibinfo{person}{Pascale Fung}.} \bibinfo{year}{2020}\natexlab{}.
\newblock \showarticletitle{Exploring Versatile Generative Language Model Via Parameter-Efficient Transfer Learning}. In \bibinfo{booktitle}{\emph{Findings of the Association for Computational Linguistics: EMNLP 2020}}. \bibinfo{pages}{441--459}.
\newblock


\bibitem[Liu et~al\mbox{.}(2019d)]%
        {liu2019darts}
\bibfield{author}{\bibinfo{person}{Hanxiao Liu}, \bibinfo{person}{Karen Simonyan}, {and} \bibinfo{person}{Yiming Yang}.} \bibinfo{year}{2019}\natexlab{d}.
\newblock \showarticletitle{Darts: Differentiable architecture search}. In \bibinfo{booktitle}{\emph{Proc. ICLR}}.
\newblock


\bibitem[Liu et~al\mbox{.}(2019c)]%
        {liu2019ntu}
\bibfield{author}{\bibinfo{person}{Jun Liu}, \bibinfo{person}{Amir Shahroudy}, \bibinfo{person}{Mauricio Perez}, \bibinfo{person}{Gang Wang}, \bibinfo{person}{Ling-Yu Duan}, {and} \bibinfo{person}{Alex~C Kot}.} \bibinfo{year}{2019}\natexlab{c}.
\newblock \showarticletitle{Ntu rgb+ d 120: A large-scale benchmark for 3d human activity understanding}.
\newblock \bibinfo{journal}{\emph{IEEE Trans. Patt. Anal. Mach. Intell.}} \bibinfo{volume}{42}, \bibinfo{number}{10} (\bibinfo{year}{2019}), \bibinfo{pages}{2684--2701}.
\newblock


\bibitem[Liu et~al\mbox{.}(2020a)]%
        {liu2020efficient}
\bibfield{author}{\bibinfo{person}{Lingbo Liu}, \bibinfo{person}{Jiaqi Chen}, \bibinfo{person}{Hefeng Wu}, \bibinfo{person}{Tianshui Chen}, \bibinfo{person}{Guanbin Li}, {and} \bibinfo{person}{Liang Lin}.} \bibinfo{year}{2020}\natexlab{a}.
\newblock \showarticletitle{Efficient crowd counting via structured knowledge transfer}. In \bibinfo{booktitle}{\emph{ACM MM}}. \bibinfo{pages}{2645--2654}.
\newblock


\bibitem[Liu et~al\mbox{.}(2021a)]%
        {liu2021cross}
\bibfield{author}{\bibinfo{person}{Lingbo Liu}, \bibinfo{person}{Jiaqi Chen}, \bibinfo{person}{Hefeng Wu}, \bibinfo{person}{Guanbin Li}, \bibinfo{person}{Chenglong Li}, {and} \bibinfo{person}{Liang Lin}.} \bibinfo{year}{2021}\natexlab{a}.
\newblock \showarticletitle{Cross-modal collaborative representation learning and a large-scale rgbt benchmark for crowd counting}. In \bibinfo{booktitle}{\emph{Proc. CVPR}}. \bibinfo{pages}{4823--4833}.
\newblock


\bibitem[Liu et~al\mbox{.}(2022c)]%
        {liu2022prompt}
\bibfield{author}{\bibinfo{person}{Lingbo Liu}, \bibinfo{person}{Bruce~XB Yu}, \bibinfo{person}{Jianlong Chang}, \bibinfo{person}{Qi Tian}, {and} \bibinfo{person}{Chang-Wen Chen}.} \bibinfo{year}{2022}\natexlab{c}.
\newblock \showarticletitle{Prompt-matched semantic segmentation}.
\newblock \bibinfo{journal}{\emph{Preprint at https://arxiv.org/abs/2208.10159}} (\bibinfo{year}{2022}).
\newblock


\bibitem[Liu et~al\mbox{.}(2021d)]%
        {liu2021pre}
\bibfield{author}{\bibinfo{person}{Pengfei Liu}, \bibinfo{person}{Weizhe Yuan}, \bibinfo{person}{Jinlan Fu}, \bibinfo{person}{Zhengbao Jiang}, \bibinfo{person}{Hiroaki Hayashi}, {and} \bibinfo{person}{Graham Neubig}.} \bibinfo{year}{2021}\natexlab{d}.
\newblock \showarticletitle{Pre-train, prompt, and predict: A systematic survey of prompting methods in natural language processing}.
\newblock \bibinfo{journal}{\emph{Preprint at https://arxiv.org/abs/2107.13586}} (\bibinfo{year}{2021}).
\newblock


\bibitem[Liu and Wang(2023)]%
        {liu2023ten}
\bibfield{author}{\bibinfo{person}{Shiwei Liu} {and} \bibinfo{person}{Zhangyang Wang}.} \bibinfo{year}{2023}\natexlab{}.
\newblock \showarticletitle{Ten Lessons We Have Learned in the New" Sparseland": A Short Handbook for Sparse Neural Network Researchers}.
\newblock \bibinfo{journal}{\emph{Preprint at https://arxiv.org/abs/2302.02596}} (\bibinfo{year}{2023}).
\newblock


\bibitem[Liu and Zhang(2022)]%
        {liu2022closer}
\bibfield{author}{\bibinfo{person}{Xiaobin Liu} {and} \bibinfo{person}{Shiliang Zhang}.} \bibinfo{year}{2022}\natexlab{}.
\newblock \showarticletitle{Who is closer: A computational method for domain gap evaluation}.
\newblock \bibinfo{journal}{\emph{Patt. Recog.}}  \bibinfo{volume}{122} (\bibinfo{year}{2022}), \bibinfo{pages}{108293}.
\newblock


\bibitem[Liu et~al\mbox{.}(2019a)]%
        {liu2019knowledge}
\bibfield{author}{\bibinfo{person}{Yufan Liu}, \bibinfo{person}{Jiajiong Cao}, \bibinfo{person}{Bing Li}, \bibinfo{person}{Chunfeng Yuan}, \bibinfo{person}{Weiming Hu}, \bibinfo{person}{Yangxi Li}, {and} \bibinfo{person}{Yunqiang Duan}.} \bibinfo{year}{2019}\natexlab{a}.
\newblock \showarticletitle{Knowledge distillation via instance relationship graph}. In \bibinfo{booktitle}{\emph{Proc. CVPR}}. \bibinfo{pages}{7096--7104}.
\newblock


\bibitem[Liu et~al\mbox{.}(2019b)]%
        {liu2019structured}
\bibfield{author}{\bibinfo{person}{Yifan Liu}, \bibinfo{person}{Ke Chen}, \bibinfo{person}{Chris Liu}, \bibinfo{person}{Zengchang Qin}, \bibinfo{person}{Zhenbo Luo}, {and} \bibinfo{person}{Jingdong Wang}.} \bibinfo{year}{2019}\natexlab{b}.
\newblock \showarticletitle{Structured knowledge distillation for semantic segmentation}. In \bibinfo{booktitle}{\emph{Proc. CVPR}}. \bibinfo{pages}{2604--2613}.
\newblock


\bibitem[Liu et~al\mbox{.}(2020b)]%
        {liu2020search}
\bibfield{author}{\bibinfo{person}{Yu Liu}, \bibinfo{person}{Xuhui Jia}, \bibinfo{person}{Mingxing Tan}, \bibinfo{person}{Raviteja Vemulapalli}, \bibinfo{person}{Yukun Zhu}, \bibinfo{person}{Bradley Green}, {and} \bibinfo{person}{Xiaogang Wang}.} \bibinfo{year}{2020}\natexlab{b}.
\newblock \showarticletitle{Search to distill: Pearls are everywhere but not the eyes}. In \bibinfo{booktitle}{\emph{Proc. CVPR}}. \bibinfo{pages}{7539--7548}.
\newblock


\bibitem[Liu et~al\mbox{.}(2021c)]%
        {liu2021exploring}
\bibfield{author}{\bibinfo{person}{Yang Liu}, \bibinfo{person}{Cheng Lyu}, \bibinfo{person}{Zhiyuan Liu}, {and} \bibinfo{person}{Jinde Cao}.} \bibinfo{year}{2021}\natexlab{c}.
\newblock \showarticletitle{Exploring a large-scale multi-modal transportation recommendation system}.
\newblock \bibinfo{journal}{\emph{Transportation Research Part C: Emerging Technologies}}  \bibinfo{volume}{126} (\bibinfo{year}{2021}), \bibinfo{pages}{103070}.
\newblock


\bibitem[Liu et~al\mbox{.}(2020c)]%
        {liu2020structured}
\bibfield{author}{\bibinfo{person}{Yifan Liu}, \bibinfo{person}{Changyong Shu}, \bibinfo{person}{Jingdong Wang}, {and} \bibinfo{person}{Chunhua Shen}.} \bibinfo{year}{2020}\natexlab{c}.
\newblock \showarticletitle{Structured knowledge distillation for dense prediction}.
\newblock \bibinfo{journal}{\emph{IEEE Trans. Patt. Anal. Mach. Intell.}} (\bibinfo{year}{2020}).
\newblock


\bibitem[Liu et~al\mbox{.}(2024)]%
        {liu2024sora}
\bibfield{author}{\bibinfo{person}{Yixin Liu}, \bibinfo{person}{Kai Zhang}, \bibinfo{person}{Yuan Li}, \bibinfo{person}{Zhiling Yan}, \bibinfo{person}{Chujie Gao}, \bibinfo{person}{Ruoxi Chen}, \bibinfo{person}{Zhengqing Yuan}, \bibinfo{person}{Yue Huang}, \bibinfo{person}{Hanchi Sun}, \bibinfo{person}{Jianfeng Gao}, {et~al\mbox{.}}} \bibinfo{year}{2024}\natexlab{}.
\newblock \showarticletitle{Sora: A Review on Background, Technology, Limitations, and Opportunities of Large Vision Models}.
\newblock \bibinfo{journal}{\emph{arXiv preprint arXiv:2402.17177}} (\bibinfo{year}{2024}).
\newblock


\bibitem[Liu et~al\mbox{.}(2022a)]%
        {liu2022polyhistor}
\bibfield{author}{\bibinfo{person}{Yen-Cheng Liu}, \bibinfo{person}{Chih-Yao Ma}, \bibinfo{person}{Junjiao Tian}, \bibinfo{person}{Zijian He}, {and} \bibinfo{person}{Zsolt Kira}.} \bibinfo{year}{2022}\natexlab{a}.
\newblock \showarticletitle{Polyhistor: Parameter-Efficient Multi-Task Adaptation for Dense Vision Tasks}.
\newblock \bibinfo{journal}{\emph{Preprint at https://arxiv.org/abs/2210.03265}} (\bibinfo{year}{2022}).
\newblock


\bibitem[Liu et~al\mbox{.}(2021b)]%
        {liu2021swin}
\bibfield{author}{\bibinfo{person}{Ze Liu}, \bibinfo{person}{Yutong Lin}, \bibinfo{person}{Yue Cao}, \bibinfo{person}{Han Hu}, \bibinfo{person}{Yixuan Wei}, \bibinfo{person}{Zheng Zhang}, \bibinfo{person}{Stephen Lin}, {and} \bibinfo{person}{Baining Guo}.} \bibinfo{year}{2021}\natexlab{b}.
\newblock \showarticletitle{Swin transformer: Hierarchical vision transformer using shifted windows}. In \bibinfo{booktitle}{\emph{Proc. CVPR}}. \bibinfo{pages}{10012--10022}.
\newblock


\bibitem[Liu et~al\mbox{.}(2022b)]%
        {liu2022video}
\bibfield{author}{\bibinfo{person}{Ze Liu}, \bibinfo{person}{Jia Ning}, \bibinfo{person}{Yue Cao}, \bibinfo{person}{Yixuan Wei}, \bibinfo{person}{Zheng Zhang}, \bibinfo{person}{Stephen Lin}, {and} \bibinfo{person}{Han Hu}.} \bibinfo{year}{2022}\natexlab{b}.
\newblock \showarticletitle{Video swin transformer}. In \bibinfo{booktitle}{\emph{Proc. CVPR}}. \bibinfo{pages}{3202--3211}.
\newblock


\bibitem[Loedeman et~al\mbox{.}(2022)]%
        {loedeman2022prompt}
\bibfield{author}{\bibinfo{person}{Jochem Loedeman}, \bibinfo{person}{Maarten~C Stol}, \bibinfo{person}{Tengda Han}, {and} \bibinfo{person}{Yuki~M Asano}.} \bibinfo{year}{2022}\natexlab{}.
\newblock \showarticletitle{Prompt Generation Networks for Efficient Adaptation of Frozen Vision Transformers}.
\newblock \bibinfo{journal}{\emph{Preprint at https://arxiv.org/abs/2210.06466}} (\bibinfo{year}{2022}).
\newblock


\bibitem[Lu et~al\mbox{.}(2023)]%
        {lu2023uniadapter}
\bibfield{author}{\bibinfo{person}{Haoyu Lu}, \bibinfo{person}{Mingyu Ding}, \bibinfo{person}{Yuqi Huo}, \bibinfo{person}{Guoxing Yang}, \bibinfo{person}{Zhiwu Lu}, \bibinfo{person}{Masayoshi Tomizuka}, {and} \bibinfo{person}{Wei Zhan}.} \bibinfo{year}{2023}\natexlab{}.
\newblock \showarticletitle{UniAdapter: Unified Parameter-Efficient Transfer Learning for Cross-modal Modeling}.
\newblock \bibinfo{journal}{\emph{Preprint at https://arxiv.org/abs/2302.06605}} (\bibinfo{year}{2023}).
\newblock


\bibitem[Luo et~al\mbox{.}(2023a)]%
        {luo2023towards}
\bibfield{author}{\bibinfo{person}{Gen Luo}, \bibinfo{person}{Minglang Huang}, \bibinfo{person}{Yiyi Zhou}, \bibinfo{person}{Xiaoshuai Sun}, \bibinfo{person}{Guannan Jiang}, \bibinfo{person}{Zhiyu Wang}, {and} \bibinfo{person}{Rongrong Ji}.} \bibinfo{year}{2023}\natexlab{a}.
\newblock \showarticletitle{Towards Efficient Visual Adaption via Structural Re-parameterization}.
\newblock \bibinfo{journal}{\emph{Preprint at https://arxiv.org/abs/2302.08106}} (\bibinfo{year}{2023}).
\newblock


\bibitem[Luo et~al\mbox{.}(2022)]%
        {luo2022towards}
\bibfield{author}{\bibinfo{person}{Gen Luo}, \bibinfo{person}{Yiyi Zhou}, \bibinfo{person}{Xiaoshuai Sun}, \bibinfo{person}{Yan Wang}, \bibinfo{person}{Liujuan Cao}, \bibinfo{person}{Yongjian Wu}, \bibinfo{person}{Feiyue Huang}, {and} \bibinfo{person}{Rongrong Ji}.} \bibinfo{year}{2022}\natexlab{}.
\newblock \showarticletitle{Towards lightweight transformer via group-wise transformation for vision-and-language tasks}.
\newblock \bibinfo{journal}{\emph{TIP}}  \bibinfo{volume}{31} (\bibinfo{year}{2022}), \bibinfo{pages}{3386--3398}.
\newblock


\bibitem[Luo et~al\mbox{.}(2023b)]%
        {luo2023survey}
\bibfield{author}{\bibinfo{person}{Xiao Luo}, \bibinfo{person}{Haixin Wang}, \bibinfo{person}{Daqing Wu}, \bibinfo{person}{Chong Chen}, \bibinfo{person}{Minghua Deng}, \bibinfo{person}{Jianqiang Huang}, {and} \bibinfo{person}{Xian-Sheng Hua}.} \bibinfo{year}{2023}\natexlab{b}.
\newblock \showarticletitle{A survey on deep hashing methods}.
\newblock \bibinfo{journal}{\emph{ACM Transactions on Knowledge Discovery from Data}} \bibinfo{volume}{17}, \bibinfo{number}{1} (\bibinfo{year}{2023}), \bibinfo{pages}{1--50}.
\newblock


\bibitem[Ma et~al\mbox{.}(2023)]%
        {ma2023understanding}
\bibfield{author}{\bibinfo{person}{Chengcheng Ma}, \bibinfo{person}{Yang Liu}, \bibinfo{person}{Jiankang Deng}, \bibinfo{person}{Lingxi Xie}, \bibinfo{person}{Weiming Dong}, {and} \bibinfo{person}{Changsheng Xu}.} \bibinfo{year}{2023}\natexlab{}.
\newblock \showarticletitle{Understanding and Mitigating Overfitting in Prompt Tuning for Vision-Language Models}.
\newblock \bibinfo{journal}{\emph{TCSVT}} (\bibinfo{year}{2023}).
\newblock


\bibitem[Ma et~al\mbox{.}(2021)]%
        {ma2021simple}
\bibfield{author}{\bibinfo{person}{Teli Ma}, \bibinfo{person}{Shijie Geng}, \bibinfo{person}{Mengmeng Wang}, \bibinfo{person}{Jing Shao}, \bibinfo{person}{Jiasen Lu}, \bibinfo{person}{Hongsheng Li}, \bibinfo{person}{Peng Gao}, {and} \bibinfo{person}{Yu Qiao}.} \bibinfo{year}{2021}\natexlab{}.
\newblock \showarticletitle{A simple long-tailed recognition baseline via vision-language model}.
\newblock \bibinfo{journal}{\emph{Preprint at https://arxiv.org/abs/2111.14745}} (\bibinfo{year}{2021}).
\newblock


\bibitem[Ma et~al\mbox{.}(2022)]%
        {ma2022dhwp}
\bibfield{author}{\bibinfo{person}{Zeyu Ma}, \bibinfo{person}{Yuhang Guo}, \bibinfo{person}{Xiao Luo}, \bibinfo{person}{Chong Chen}, \bibinfo{person}{Minghua Deng}, \bibinfo{person}{Wei Cheng}, {and} \bibinfo{person}{Guangming Lu}.} \bibinfo{year}{2022}\natexlab{}.
\newblock \showarticletitle{DHWP: Learning High-Quality Short Hash Codes Via Weight Pruning}. In \bibinfo{booktitle}{\emph{ICASSP}}. IEEE, \bibinfo{pages}{4783--4787}.
\newblock


\bibitem[Mao et~al\mbox{.}(2022)]%
        {mao2022understanding}
\bibfield{author}{\bibinfo{person}{Chengzhi Mao}, \bibinfo{person}{Scott Geng}, \bibinfo{person}{Junfeng Yang}, \bibinfo{person}{Xin Wang}, {and} \bibinfo{person}{Carl Vondrick}.} \bibinfo{year}{2022}\natexlab{}.
\newblock \showarticletitle{Understanding Zero-Shot Adversarial Robustness for Large-Scale Models}.
\newblock \bibinfo{journal}{\emph{Preprint at https://arxiv.org/abs/2212.07016}} (\bibinfo{year}{2022}).
\newblock


\bibitem[Mao et~al\mbox{.}(2021)]%
        {mao2021unipelt}
\bibfield{author}{\bibinfo{person}{Yuning Mao}, \bibinfo{person}{Lambert Mathias}, \bibinfo{person}{Rui Hou}, \bibinfo{person}{Amjad Almahairi}, \bibinfo{person}{Hao Ma}, \bibinfo{person}{Jiawei Han}, \bibinfo{person}{Wen-tau Yih}, {and} \bibinfo{person}{Madian Khabsa}.} \bibinfo{year}{2021}\natexlab{}.
\newblock \showarticletitle{Unipelt: A unified framework for parameter-efficient language model tuning}.
\newblock \bibinfo{journal}{\emph{Preprint at https://arxiv.org/abs/2110.07577}} (\bibinfo{year}{2021}).
\newblock


\bibitem[MAROUF et~al\mbox{.}(2023)]%
        {marouf2023tiny}
\bibfield{author}{\bibinfo{person}{Imad~Eddine MAROUF}, \bibinfo{person}{Enzo Tartaglione}, {and} \bibinfo{person}{St{\'e}phane Lathuili{\`e}re}.} \bibinfo{year}{2023}\natexlab{}.
\newblock \bibinfo{title}{Tiny Adapters for Vision Transformers}.
\newblock
\newblock
\urldef\tempurl%
\url{https://openreview.net/forum?id=V0Vo9eW2nzL}
\showURL{%
\tempurl}


\bibitem[Marr(2010)]%
        {marr2010vision}
\bibfield{author}{\bibinfo{person}{David Marr}.} \bibinfo{year}{2010}\natexlab{}.
\newblock \bibinfo{booktitle}{\emph{Vision: A computational investigation into the human representation and processing of visual information}}.
\newblock \bibinfo{publisher}{MIT press}.
\newblock


\bibitem[Metzer et~al\mbox{.}(2023)]%
        {metzer2023latent}
\bibfield{author}{\bibinfo{person}{Gal Metzer}, \bibinfo{person}{Elad Richardson}, \bibinfo{person}{Or Patashnik}, \bibinfo{person}{Raja Giryes}, {and} \bibinfo{person}{Daniel Cohen-Or}.} \bibinfo{year}{2023}\natexlab{}.
\newblock \showarticletitle{Latent-nerf for shape-guided generation of 3d shapes and textures}. In \bibinfo{booktitle}{\emph{Proc. CVPR}}. \bibinfo{pages}{12663--12673}.
\newblock


\bibitem[Mildenhall et~al\mbox{.}(2021)]%
        {mildenhall2021nerf}
\bibfield{author}{\bibinfo{person}{Ben Mildenhall}, \bibinfo{person}{Pratul~P Srinivasan}, \bibinfo{person}{Matthew Tancik}, \bibinfo{person}{Jonathan~T Barron}, \bibinfo{person}{Ravi Ramamoorthi}, {and} \bibinfo{person}{Ren Ng}.} \bibinfo{year}{2021}\natexlab{}.
\newblock \showarticletitle{Nerf: Representing scenes as neural radiance fields for view synthesis}.
\newblock \bibinfo{journal}{\emph{Commun. ACM}} \bibinfo{volume}{65}, \bibinfo{number}{1} (\bibinfo{year}{2021}), \bibinfo{pages}{99--106}.
\newblock


\bibitem[Namazifar et~al\mbox{.}(2023)]%
        {namazifar2023role}
\bibfield{author}{\bibinfo{person}{Mahdi Namazifar}, \bibinfo{person}{Devamanyu Hazarika}, {and} \bibinfo{person}{Dilek Hakkani-Tur}.} \bibinfo{year}{2023}\natexlab{}.
\newblock \showarticletitle{Role of Bias Terms in Dot-Product Attention}.
\newblock \bibinfo{journal}{\emph{Preprint at https://arxiv.org/abs/2302.08626}} (\bibinfo{year}{2023}).
\newblock


\bibitem[Nguyen et~al\mbox{.}(2023)]%
        {nguyen2023climax}
\bibfield{author}{\bibinfo{person}{Tung Nguyen}, \bibinfo{person}{Johannes Brandstetter}, \bibinfo{person}{Ashish Kapoor}, \bibinfo{person}{Jayesh~K Gupta}, {and} \bibinfo{person}{Aditya Grover}.} \bibinfo{year}{2023}\natexlab{}.
\newblock \showarticletitle{ClimaX: A foundation model for weather and climate}.
\newblock \bibinfo{journal}{\emph{Preprint at https://arxiv.org/abs/2301.10343}} (\bibinfo{year}{2023}).
\newblock


\bibitem[Ni et~al\mbox{.}(2022)]%
        {ni2022expanding}
\bibfield{author}{\bibinfo{person}{Bolin Ni}, \bibinfo{person}{Houwen Peng}, \bibinfo{person}{Minghao Chen}, \bibinfo{person}{Songyang Zhang}, \bibinfo{person}{Gaofeng Meng}, \bibinfo{person}{Jianlong Fu}, \bibinfo{person}{Shiming Xiang}, {and} \bibinfo{person}{Haibin Ling}.} \bibinfo{year}{2022}\natexlab{}.
\newblock \showarticletitle{Expanding language-image pretrained models for general video recognition}. In \bibinfo{booktitle}{\emph{Computer Vision--ECCV 2022: 17th European Conference, Tel Aviv, Israel, October 23--27, 2022, Proceedings, Part IV}}. Springer, \bibinfo{pages}{1--18}.
\newblock


\bibitem[Nie et~al\mbox{.}(2022)]%
        {nie2022pro}
\bibfield{author}{\bibinfo{person}{Xing Nie}, \bibinfo{person}{Bolin Ni}, \bibinfo{person}{Jianlong Chang}, \bibinfo{person}{Gaomeng Meng}, \bibinfo{person}{Chunlei Huo}, \bibinfo{person}{Zhaoxiang Zhang}, \bibinfo{person}{Shiming Xiang}, \bibinfo{person}{Qi Tian}, {and} \bibinfo{person}{Chunhong Pan}.} \bibinfo{year}{2022}\natexlab{}.
\newblock \showarticletitle{Pro-tuning: Unified prompt tuning for vision tasks}.
\newblock \bibinfo{journal}{\emph{Preprint at https://arxiv.org/abs/2207.14381}} (\bibinfo{year}{2022}).
\newblock


\bibitem[Pan et~al\mbox{.}(2022a)]%
        {panst}
\bibfield{author}{\bibinfo{person}{Junting Pan}, \bibinfo{person}{Ziyi Lin}, \bibinfo{person}{Xiatian Zhu}, \bibinfo{person}{Jing Shao}, {and} \bibinfo{person}{Hongsheng Li}.} \bibinfo{year}{2022}\natexlab{a}.
\newblock \showarticletitle{ST-Adapter: Parameter-Efficient Image-to-Video Transfer Learning}. In \bibinfo{booktitle}{\emph{Proc. NeurIPS}}.
\newblock


\bibitem[Pan et~al\mbox{.}(2022b)]%
        {pan2022st}
\bibfield{author}{\bibinfo{person}{Junting Pan}, \bibinfo{person}{Ziyi Lin}, \bibinfo{person}{Xiatian Zhu}, \bibinfo{person}{Jing Shao}, {and} \bibinfo{person}{Hongsheng Li}.} \bibinfo{year}{2022}\natexlab{b}.
\newblock \showarticletitle{ST-Adapter: Parameter-Efficient Image-to-Video Transfer Learning for Action Recognition}.
\newblock \bibinfo{journal}{\emph{Preprint at https://arxiv.org/abs/2206.13559}} (\bibinfo{year}{2022}).
\newblock


\bibitem[Pantazis et~al\mbox{.}(2022)]%
        {pantazis2022svl}
\bibfield{author}{\bibinfo{person}{Omiros Pantazis}, \bibinfo{person}{Gabriel Brostow}, \bibinfo{person}{Kate Jones}, {and} \bibinfo{person}{Oisin Mac~Aodha}.} \bibinfo{year}{2022}\natexlab{}.
\newblock \showarticletitle{SVL-Adapter: Self-Supervised Adapter for Vision-Language Pretrained Models}.
\newblock \bibinfo{journal}{\emph{Preprint at https://arxiv.org/abs/2210.03794}} (\bibinfo{year}{2022}).
\newblock


\bibitem[Papalampidi and Lapata(2022)]%
        {papalampidi2022hierarchical3d}
\bibfield{author}{\bibinfo{person}{Pinelopi Papalampidi} {and} \bibinfo{person}{Mirella Lapata}.} \bibinfo{year}{2022}\natexlab{}.
\newblock \showarticletitle{Hierarchical3D Adapters for Long Video-to-text Summarization}.
\newblock \bibinfo{journal}{\emph{Preprint at https://arxiv.org/abs/2210.04829}} (\bibinfo{year}{2022}).
\newblock


\bibitem[Park et~al\mbox{.}(2022)]%
        {park2022lanit}
\bibfield{author}{\bibinfo{person}{Jihye Park}, \bibinfo{person}{Soohyun Kim}, \bibinfo{person}{Sunwoo Kim}, \bibinfo{person}{Jaejun Yoo}, \bibinfo{person}{Youngjung Uh}, {and} \bibinfo{person}{Seungryong Kim}.} \bibinfo{year}{2022}\natexlab{}.
\newblock \showarticletitle{LANIT: Language-Driven Image-to-Image Translation for Unlabeled Data}.
\newblock \bibinfo{journal}{\emph{Preprint at https://arxiv.org/abs/2208.14889}} (\bibinfo{year}{2022}).
\newblock


\bibitem[Park et~al\mbox{.}(2019)]%
        {park2019relational}
\bibfield{author}{\bibinfo{person}{Wonpyo Park}, \bibinfo{person}{Dongju Kim}, \bibinfo{person}{Yan Lu}, {and} \bibinfo{person}{Minsu Cho}.} \bibinfo{year}{2019}\natexlab{}.
\newblock \showarticletitle{Relational knowledge distillation}. In \bibinfo{booktitle}{\emph{Proc. CVPR}}. \bibinfo{pages}{3967--3976}.
\newblock


\bibitem[Passalis et~al\mbox{.}(2020)]%
        {passalis2020heterogeneous}
\bibfield{author}{\bibinfo{person}{Nikolaos Passalis}, \bibinfo{person}{Maria Tzelepi}, {and} \bibinfo{person}{Anastasios Tefas}.} \bibinfo{year}{2020}\natexlab{}.
\newblock \showarticletitle{Heterogeneous knowledge distillation using information flow modeling}. In \bibinfo{booktitle}{\emph{Proc. CVPR}}. \bibinfo{pages}{2339--2348}.
\newblock


\bibitem[Peebles and Xie(2023)]%
        {peebles2023scalable}
\bibfield{author}{\bibinfo{person}{William Peebles} {and} \bibinfo{person}{Saining Xie}.} \bibinfo{year}{2023}\natexlab{}.
\newblock \showarticletitle{Scalable diffusion models with transformers}. In \bibinfo{booktitle}{\emph{Proc. CVPR}}. \bibinfo{pages}{4195--4205}.
\newblock


\bibitem[Peng et~al\mbox{.}(2019)]%
        {peng2019correlation}
\bibfield{author}{\bibinfo{person}{Baoyun Peng}, \bibinfo{person}{Xiao Jin}, \bibinfo{person}{Jiaheng Liu}, \bibinfo{person}{Dongsheng Li}, \bibinfo{person}{Yichao Wu}, \bibinfo{person}{Yu Liu}, \bibinfo{person}{Shunfeng Zhou}, {and} \bibinfo{person}{Zhaoning Zhang}.} \bibinfo{year}{2019}\natexlab{}.
\newblock \showarticletitle{Correlation congruence for knowledge distillation}. In \bibinfo{booktitle}{\emph{Proc. CVPR}}. \bibinfo{pages}{5007--5016}.
\newblock


\bibitem[Peng et~al\mbox{.}(2022)]%
        {peng2022sgva}
\bibfield{author}{\bibinfo{person}{Fang Peng}, \bibinfo{person}{Xiaoshan Yang}, {and} \bibinfo{person}{Changsheng Xu}.} \bibinfo{year}{2022}\natexlab{}.
\newblock \showarticletitle{SgVA-CLIP: Semantic-guided Visual Adapting of Vision-Language Models for Few-shot Image Classification}.
\newblock \bibinfo{journal}{\emph{Preprint at https://arxiv.org/abs/2211.16191}} (\bibinfo{year}{2022}).
\newblock


\bibitem[Pfeiffer et~al\mbox{.}(2020)]%
        {pfeiffer2020adapterhub}
\bibfield{author}{\bibinfo{person}{Jonas Pfeiffer}, \bibinfo{person}{Andreas R{\"u}ckl{\'e}}, \bibinfo{person}{Clifton Poth}, \bibinfo{person}{Aishwarya Kamath}, \bibinfo{person}{Ivan Vuli{\'c}}, \bibinfo{person}{Sebastian Ruder}, \bibinfo{person}{Kyunghyun Cho}, {and} \bibinfo{person}{Iryna Gurevych}.} \bibinfo{year}{2020}\natexlab{}.
\newblock \showarticletitle{Adapterhub: A framework for adapting transformers}.
\newblock \bibinfo{journal}{\emph{arXiv preprint arXiv:2007.07779}} (\bibinfo{year}{2020}).
\newblock


\bibitem[Pham et~al\mbox{.}(2021)]%
        {pham2021meta}
\bibfield{author}{\bibinfo{person}{Hieu Pham}, \bibinfo{person}{Zihang Dai}, \bibinfo{person}{Qizhe Xie}, {and} \bibinfo{person}{Quoc~V Le}.} \bibinfo{year}{2021}\natexlab{}.
\newblock \showarticletitle{Meta pseudo labels}. In \bibinfo{booktitle}{\emph{Proc. CVPR}}. \bibinfo{pages}{11557--11568}.
\newblock


\bibitem[Pham et~al\mbox{.}(2018)]%
        {pham2018efficient}
\bibfield{author}{\bibinfo{person}{Hieu Pham}, \bibinfo{person}{Melody Guan}, \bibinfo{person}{Barret Zoph}, \bibinfo{person}{Quoc Le}, {and} \bibinfo{person}{Jeff Dean}.} \bibinfo{year}{2018}\natexlab{}.
\newblock \showarticletitle{Efficient neural architecture search via parameters sharing}. In \bibinfo{booktitle}{\emph{Proc. ICML}}. PMLR, \bibinfo{pages}{4095--4104}.
\newblock


\bibitem[Porrello et~al\mbox{.}(2020)]%
        {porrello2020robust}
\bibfield{author}{\bibinfo{person}{Angelo Porrello}, \bibinfo{person}{Luca Bergamini}, {and} \bibinfo{person}{Simone Calderara}.} \bibinfo{year}{2020}\natexlab{}.
\newblock \showarticletitle{Robust re-identification by multiple views knowledge distillation}. In \bibinfo{booktitle}{\emph{Proc. ECCV}}. Springer, \bibinfo{pages}{93--110}.
\newblock


\bibitem[Qin et~al\mbox{.}(2023a)]%
        {qin2023tool}
\bibfield{author}{\bibinfo{person}{Yujia Qin}, \bibinfo{person}{Shengding Hu}, \bibinfo{person}{Yankai Lin}, \bibinfo{person}{Weize Chen}, \bibinfo{person}{Ning Ding}, \bibinfo{person}{Ganqu Cui}, \bibinfo{person}{Zheni Zeng}, \bibinfo{person}{Yufei Huang}, \bibinfo{person}{Chaojun Xiao}, \bibinfo{person}{Chi Han}, {et~al\mbox{.}}} \bibinfo{year}{2023}\natexlab{a}.
\newblock \showarticletitle{Tool learning with foundation models}.
\newblock \bibinfo{journal}{\emph{arXiv preprint arXiv:2304.08354}} (\bibinfo{year}{2023}).
\newblock


\bibitem[Qin et~al\mbox{.}(2023b)]%
        {qin2023low}
\bibfield{author}{\bibinfo{person}{Ziran Qin}, \bibinfo{person}{Mingbao Lin}, {and} \bibinfo{person}{Weiyao Lin}.} \bibinfo{year}{2023}\natexlab{b}.
\newblock \showarticletitle{Low-Rank Winograd Transformation for 3D Convolutional Neural Networks}.
\newblock \bibinfo{journal}{\emph{Preprint at https://arxiv.org/abs/2301.11180}} (\bibinfo{year}{2023}).
\newblock


\bibitem[Qu et~al\mbox{.}(2021)]%
        {qu2021recent}
\bibfield{author}{\bibinfo{person}{Haoxuan Qu}, \bibinfo{person}{Hossein Rahmani}, \bibinfo{person}{Li Xu}, \bibinfo{person}{Bryan Williams}, {and} \bibinfo{person}{Jun Liu}.} \bibinfo{year}{2021}\natexlab{}.
\newblock \showarticletitle{Recent advances of continual learning in computer vision: An overview}.
\newblock \bibinfo{journal}{\emph{Preprint at https://arxiv.org/abs/2109.11369}} (\bibinfo{year}{2021}).
\newblock


\bibitem[Radford et~al\mbox{.}(2021)]%
        {radford2021learning}
\bibfield{author}{\bibinfo{person}{Alec Radford}, \bibinfo{person}{Jong~Wook Kim}, \bibinfo{person}{Chris Hallacy}, \bibinfo{person}{Aditya Ramesh}, \bibinfo{person}{Gabriel Goh}, \bibinfo{person}{Sandhini Agarwal}, \bibinfo{person}{Girish Sastry}, \bibinfo{person}{Amanda Askell}, \bibinfo{person}{Pamela Mishkin}, \bibinfo{person}{Jack Clark}, {et~al\mbox{.}}} \bibinfo{year}{2021}\natexlab{}.
\newblock \showarticletitle{Learning transferable visual models from natural language supervision}. In \bibinfo{booktitle}{\emph{Proc. ICML}}. PMLR, \bibinfo{pages}{8748--8763}.
\newblock


\bibitem[Rao et~al\mbox{.}(2022a)]%
        {rao2022parameter}
\bibfield{author}{\bibinfo{person}{Jun Rao}, \bibinfo{person}{Xv Meng}, \bibinfo{person}{Liang Ding}, \bibinfo{person}{Shuhan Qi}, {and} \bibinfo{person}{Dacheng Tao}.} \bibinfo{year}{2022}\natexlab{a}.
\newblock \showarticletitle{Parameter-efficient and student-friendly knowledge distillation}.
\newblock \bibinfo{journal}{\emph{Preprint at https://arxiv.org/abs/2205.15308}} (\bibinfo{year}{2022}).
\newblock


\bibitem[Rao et~al\mbox{.}(2022b)]%
        {rao2022amixer}
\bibfield{author}{\bibinfo{person}{Yongming Rao}, \bibinfo{person}{Wenliang Zhao}, \bibinfo{person}{Jie Zhou}, {and} \bibinfo{person}{Jiwen Lu}.} \bibinfo{year}{2022}\natexlab{b}.
\newblock \showarticletitle{AMixer: Adaptive Weight Mixing for Self-attention Free Vision Transformers}. In \bibinfo{booktitle}{\emph{Proc. ECCV}}. Springer, \bibinfo{pages}{50--67}.
\newblock


\bibitem[Rebuffi et~al\mbox{.}(2017)]%
        {rebuffi2017learning}
\bibfield{author}{\bibinfo{person}{Alvise Rebuffi}, \bibinfo{person}{Hakan Bilen}, {and} \bibinfo{person}{Andrea Vedaldi}.} \bibinfo{year}{2017}\natexlab{}.
\newblock \showarticletitle{Learning multiple visual domains with residual adapters}. In \bibinfo{booktitle}{\emph{Proc. NeurIPS}}.
\newblock


\bibitem[Rebuffi et~al\mbox{.}(2018)]%
        {rebuffi2018efficient}
\bibfield{author}{\bibinfo{person}{Sylvestre-Alvise Rebuffi}, \bibinfo{person}{Hakan Bilen}, {and} \bibinfo{person}{Andrea Vedaldi}.} \bibinfo{year}{2018}\natexlab{}.
\newblock \showarticletitle{Efficient parametrization of multi-domain deep neural networks}. In \bibinfo{booktitle}{\emph{Proc. CVPR}}. \bibinfo{pages}{8119--8127}.
\newblock


\bibitem[Reed et~al\mbox{.}(2022)]%
        {reed2022generalist}
\bibfield{author}{\bibinfo{person}{Scott Reed}, \bibinfo{person}{Konrad Zolna}, \bibinfo{person}{Emilio Parisotto}, \bibinfo{person}{Sergio~Gomez Colmenarejo}, \bibinfo{person}{Alexander Novikov}, \bibinfo{person}{Gabriel Barth-Maron}, \bibinfo{person}{Mai Gimenez}, \bibinfo{person}{Yury Sulsky}, \bibinfo{person}{Jackie Kay}, \bibinfo{person}{Jost~Tobias Springenberg}, {et~al\mbox{.}}} \bibinfo{year}{2022}\natexlab{}.
\newblock \showarticletitle{A generalist agent}.
\newblock \bibinfo{journal}{\emph{Preprint at https://arxiv.org/abs/2205.06175}} (\bibinfo{year}{2022}).
\newblock


\bibitem[Remigereau et~al\mbox{.}(2022)]%
        {remigereau2022knowledge}
\bibfield{author}{\bibinfo{person}{F{\'e}lix Remigereau}, \bibinfo{person}{Djebril Mekhazni}, \bibinfo{person}{Sajjad Abdoli}, \bibinfo{person}{Rafael~MO Cruz}, \bibinfo{person}{Eric Granger}, {et~al\mbox{.}}} \bibinfo{year}{2022}\natexlab{}.
\newblock \showarticletitle{Knowledge distillation for multi-target domain adaptation in real-time person re-identification}. In \bibinfo{booktitle}{\emph{ICIP}}. IEEE, \bibinfo{pages}{3853--3557}.
\newblock


\bibitem[Rombach et~al\mbox{.}(2022)]%
        {rombach2022high}
\bibfield{author}{\bibinfo{person}{Robin Rombach}, \bibinfo{person}{Andreas Blattmann}, \bibinfo{person}{Dominik Lorenz}, \bibinfo{person}{Patrick Esser}, {and} \bibinfo{person}{Bj{\"o}rn Ommer}.} \bibinfo{year}{2022}\natexlab{}.
\newblock \showarticletitle{High-resolution image synthesis with latent diffusion models}. In \bibinfo{booktitle}{\emph{Proc. CVPR}}. \bibinfo{pages}{10684--10695}.
\newblock


\bibitem[Romero et~al\mbox{.}(2015)]%
        {romero2015fitnets}
\bibfield{author}{\bibinfo{person}{Adriana Romero}, \bibinfo{person}{Nicolas Ballas}, \bibinfo{person}{Samira~Ebrahimi Kahou}, \bibinfo{person}{Antoine Chassang}, \bibinfo{person}{Carlo Gatta}, {and} \bibinfo{person}{Yoshua Bengio}.} \bibinfo{year}{2015}\natexlab{}.
\newblock \showarticletitle{Fitnets: Hints for thin deep nets}. In \bibinfo{booktitle}{\emph{Proc. ICLR}}.
\newblock


\bibitem[Rosenfeld and Tsotsos(2018)]%
        {rosenfeld2018incremental}
\bibfield{author}{\bibinfo{person}{Amir Rosenfeld} {and} \bibinfo{person}{John~K Tsotsos}.} \bibinfo{year}{2018}\natexlab{}.
\newblock \showarticletitle{Incremental learning through deep adaptation}.
\newblock \bibinfo{journal}{\emph{IEEE Trans. Patt. Anal. Mach. Intell.}} \bibinfo{volume}{42}, \bibinfo{number}{3} (\bibinfo{year}{2018}), \bibinfo{pages}{651--663}.
\newblock


\bibitem[Rublee et~al\mbox{.}(2011)]%
        {rublee2011orb}
\bibfield{author}{\bibinfo{person}{Ethan Rublee}, \bibinfo{person}{Vincent Rabaud}, \bibinfo{person}{Kurt Konolige}, {and} \bibinfo{person}{Gary Bradski}.} \bibinfo{year}{2011}\natexlab{}.
\newblock \showarticletitle{ORB: An efficient alternative to SIFT or SURF}. In \bibinfo{booktitle}{\emph{Proc. ICCV}}. IEEE, \bibinfo{pages}{2564--2571}.
\newblock


\bibitem[Ruiz et~al\mbox{.}(2023)]%
        {ruiz2023dreambooth}
\bibfield{author}{\bibinfo{person}{Nataniel Ruiz}, \bibinfo{person}{Yuanzhen Li}, \bibinfo{person}{Varun Jampani}, \bibinfo{person}{Yael Pritch}, \bibinfo{person}{Michael Rubinstein}, {and} \bibinfo{person}{Kfir Aberman}.} \bibinfo{year}{2023}\natexlab{}.
\newblock \showarticletitle{Dreambooth: Fine tuning text-to-image diffusion models for subject-driven generation}. In \bibinfo{booktitle}{\emph{Proc. CVPR}}. \bibinfo{pages}{22500--22510}.
\newblock


\bibitem[Russell(2010)]%
        {russell2010artificial}
\bibfield{author}{\bibinfo{person}{Stuart~J Russell}.} \bibinfo{year}{2010}\natexlab{}.
\newblock \bibinfo{booktitle}{\emph{Artificial intelligence a modern approach}}.
\newblock \bibinfo{publisher}{Pearson Education, Inc.}
\newblock


\bibitem[Saito et~al\mbox{.}(2022)]%
        {saito2022prefix}
\bibfield{author}{\bibinfo{person}{Kuniaki Saito}, \bibinfo{person}{Kihyuk Sohn}, \bibinfo{person}{Xiang Zhang}, \bibinfo{person}{Chun-Liang Li}, \bibinfo{person}{Chen-Yu Lee}, \bibinfo{person}{Kate Saenko}, {and} \bibinfo{person}{Tomas Pfister}.} \bibinfo{year}{2022}\natexlab{}.
\newblock \showarticletitle{Prefix conditioning unifies language and label supervision}.
\newblock \bibinfo{journal}{\emph{Preprint at https://arxiv.org/abs/2206.01125}} (\bibinfo{year}{2022}).
\newblock


\bibitem[Sandler et~al\mbox{.}(2018)]%
        {sandler2018mobilenetv2}
\bibfield{author}{\bibinfo{person}{Mark Sandler}, \bibinfo{person}{Andrew Howard}, \bibinfo{person}{Menglong Zhu}, \bibinfo{person}{Andrey Zhmoginov}, {and} \bibinfo{person}{Liang-Chieh Chen}.} \bibinfo{year}{2018}\natexlab{}.
\newblock \showarticletitle{Mobilenetv2: Inverted residuals and linear bottlenecks}. In \bibinfo{booktitle}{\emph{Proc. CVPR}}. \bibinfo{pages}{4510--4520}.
\newblock


\bibitem[Schuster and Paliwal(1997)]%
        {schuster1997bidirectional}
\bibfield{author}{\bibinfo{person}{Mike Schuster} {and} \bibinfo{person}{Kuldip~K Paliwal}.} \bibinfo{year}{1997}\natexlab{}.
\newblock \showarticletitle{Bidirectional recurrent neural networks}.
\newblock \bibinfo{journal}{\emph{IEEE Trans. Sign. Proces.}} \bibinfo{volume}{45}, \bibinfo{number}{11} (\bibinfo{year}{1997}), \bibinfo{pages}{2673--2681}.
\newblock


\bibitem[Sharma et~al\mbox{.}(2023)]%
        {sharma2023lossless}
\bibfield{author}{\bibinfo{person}{Mohit Sharma}, \bibinfo{person}{Claudio Fantacci}, \bibinfo{person}{Yuxiang Zhou}, \bibinfo{person}{Skanda Koppula}, \bibinfo{person}{Nicolas Heess}, \bibinfo{person}{Jon Scholz}, {and} \bibinfo{person}{Yusuf Aytar}.} \bibinfo{year}{2023}\natexlab{}.
\newblock \showarticletitle{Lossless Adaptation of Pretrained Vision Models For Robotic Manipulation}. In \bibinfo{booktitle}{\emph{Proc. ICLR}}.
\newblock
\urldef\tempurl%
\url{https://openreview.net/forum?id=5IND3TXJRb-}
\showURL{%
\tempurl}


\bibitem[Shen et~al\mbox{.}(2022)]%
        {shen2022multitask}
\bibfield{author}{\bibinfo{person}{Sheng Shen}, \bibinfo{person}{Shijia Yang}, \bibinfo{person}{Tianjun Zhang}, \bibinfo{person}{Bohan Zhai}, \bibinfo{person}{Joseph~E Gonzalez}, \bibinfo{person}{Kurt Keutzer}, {and} \bibinfo{person}{Trevor Darrell}.} \bibinfo{year}{2022}\natexlab{}.
\newblock \showarticletitle{Multitask Vision-Language Prompt Tuning}.
\newblock \bibinfo{journal}{\emph{Preprint at https://arxiv.org/abs/2211.11720}} (\bibinfo{year}{2022}).
\newblock


\bibitem[Shi et~al\mbox{.}(2023)]%
        {shi2023open}
\bibfield{author}{\bibinfo{person}{Yifeng Shi}, \bibinfo{person}{Feng Lv}, \bibinfo{person}{Xinliang Wang}, \bibinfo{person}{Chunlong Xia}, \bibinfo{person}{Shaojie Li}, \bibinfo{person}{Shujie Yang}, \bibinfo{person}{Teng Xi}, {and} \bibinfo{person}{Gang Zhang}.} \bibinfo{year}{2023}\natexlab{}.
\newblock \showarticletitle{Open-transmind: A new baseline and benchmark for 1st foundation model challenge of intelligent transportation}. In \bibinfo{booktitle}{\emph{Proc. CVPR Workshop}}. \bibinfo{pages}{6327--6334}.
\newblock


\bibitem[Shimomoto et~al\mbox{.}(2022)]%
        {shimomoto2022towards}
\bibfield{author}{\bibinfo{person}{Erica~K Shimomoto}, \bibinfo{person}{Edison Marrese-Taylor}, \bibinfo{person}{Hiroya Takamura}, \bibinfo{person}{Ichiro Kobayashi}, \bibinfo{person}{Hideki Nakayama}, {and} \bibinfo{person}{Yusuke Miyao}.} \bibinfo{year}{2022}\natexlab{}.
\newblock \showarticletitle{Towards Parameter-Efficient Integration of Pre-Trained Language Models In Temporal Video Grounding}.
\newblock \bibinfo{journal}{\emph{Preprint at https://arxiv.org/abs/2209.13359}} (\bibinfo{year}{2022}).
\newblock


\bibitem[Shu et~al\mbox{.}(2022)]%
        {shu2022test}
\bibfield{author}{\bibinfo{person}{Manli Shu}, \bibinfo{person}{Weili Nie}, \bibinfo{person}{De-An Huang}, \bibinfo{person}{Zhiding Yu}, \bibinfo{person}{Tom Goldstein}, \bibinfo{person}{Anima Anandkumar}, {and} \bibinfo{person}{Chaowei Xiao}.} \bibinfo{year}{2022}\natexlab{}.
\newblock \showarticletitle{Test-time prompt tuning for zero-shot generalization in vision-language models}.
\newblock \bibinfo{journal}{\emph{Preprint at https://arxiv.org/abs/2209.07511}} (\bibinfo{year}{2022}).
\newblock


\bibitem[Shysheya et~al\mbox{.}(2023)]%
        {shysheya2023fit}
\bibfield{author}{\bibinfo{person}{Aliaksandra Shysheya}, \bibinfo{person}{John~F Bronskill}, \bibinfo{person}{Massimiliano Patacchiola}, \bibinfo{person}{Sebastian Nowozin}, {and} \bibinfo{person}{Richard~E Turner}.} \bibinfo{year}{2023}\natexlab{}.
\newblock \showarticletitle{FiT: Parameter Efficient Few-shot Transfer Learning for Personalized and Federated Image Classification}. In \bibinfo{booktitle}{\emph{Proc. ICLR}}.
\newblock
\urldef\tempurl%
\url{https://openreview.net/forum?id=9aokcgBVIj1}
\showURL{%
\tempurl}


\bibitem[Simonyan and Zisserman(2014)]%
        {simonyan2014very}
\bibfield{author}{\bibinfo{person}{Karen Simonyan} {and} \bibinfo{person}{Andrew Zisserman}.} \bibinfo{year}{2014}\natexlab{}.
\newblock \showarticletitle{Very deep convolutional networks for large-scale image recognition}.
\newblock \bibinfo{journal}{\emph{Preprint at https://arxiv.org/abs/1409.1556}} (\bibinfo{year}{2014}).
\newblock


\bibitem[Singh et~al\mbox{.}(2022b)]%
        {singh2022flava}
\bibfield{author}{\bibinfo{person}{Amanpreet Singh}, \bibinfo{person}{Ronghang Hu}, \bibinfo{person}{Vedanuj Goswami}, \bibinfo{person}{Guillaume Couairon}, \bibinfo{person}{Wojciech Galuba}, \bibinfo{person}{Marcus Rohrbach}, {and} \bibinfo{person}{Douwe Kiela}.} \bibinfo{year}{2022}\natexlab{b}.
\newblock \showarticletitle{Flava: A foundational language and vision alignment model}. In \bibinfo{booktitle}{\emph{Proc. CVPR}}. \bibinfo{pages}{15638--15650}.
\newblock


\bibitem[Singh et~al\mbox{.}(2022a)]%
        {singh2022revisiting}
\bibfield{author}{\bibinfo{person}{Mannat Singh}, \bibinfo{person}{Laura Gustafson}, \bibinfo{person}{Aaron Adcock}, \bibinfo{person}{Vinicius de Freitas~Reis}, \bibinfo{person}{Bugra Gedik}, \bibinfo{person}{Raj~Prateek Kosaraju}, \bibinfo{person}{Dhruv Mahajan}, \bibinfo{person}{Ross Girshick}, \bibinfo{person}{Piotr Doll{\'a}r}, {and} \bibinfo{person}{Laurens van~der Maaten}.} \bibinfo{year}{2022}\natexlab{a}.
\newblock \showarticletitle{Revisiting Weakly Supervised Pre-Training of Visual Perception Models}. In \bibinfo{booktitle}{\emph{Proc. CVPR}}. \bibinfo{pages}{804--814}.
\newblock


\bibitem[Sohn et~al\mbox{.}(2022)]%
        {sohn2022visual}
\bibfield{author}{\bibinfo{person}{Kihyuk Sohn}, \bibinfo{person}{Yuan Hao}, \bibinfo{person}{Jos{\'e} Lezama}, \bibinfo{person}{Luisa Polania}, \bibinfo{person}{Huiwen Chang}, \bibinfo{person}{Han Zhang}, \bibinfo{person}{Irfan Essa}, {and} \bibinfo{person}{Lu Jiang}.} \bibinfo{year}{2022}\natexlab{}.
\newblock \showarticletitle{Visual Prompt Tuning for Generative Transfer Learning}.
\newblock \bibinfo{journal}{\emph{Preprint at https://arxiv.org/abs/2210.00990}} (\bibinfo{year}{2022}).
\newblock


\bibitem[Sun et~al\mbox{.}(2017)]%
        {sun2017revisiting}
\bibfield{author}{\bibinfo{person}{Chen Sun}, \bibinfo{person}{Abhinav Shrivastava}, \bibinfo{person}{Saurabh Singh}, {and} \bibinfo{person}{Abhinav Gupta}.} \bibinfo{year}{2017}\natexlab{}.
\newblock \showarticletitle{Revisiting unreasonable effectiveness of data in deep learning era}. In \bibinfo{booktitle}{\emph{Proc. ICCV}}. \bibinfo{pages}{843--852}.
\newblock


\bibitem[Sun et~al\mbox{.}(2022)]%
        {sun2022dualcoop}
\bibfield{author}{\bibinfo{person}{Ximeng Sun}, \bibinfo{person}{Ping Hu}, {and} \bibinfo{person}{Kate Saenko}.} \bibinfo{year}{2022}\natexlab{}.
\newblock \showarticletitle{Dualcoop: Fast adaptation to multi-label recognition with limited annotations}.
\newblock \bibinfo{journal}{\emph{Preprint at https://arxiv.org/abs/2206.09541}} (\bibinfo{year}{2022}).
\newblock


\bibitem[Sun et~al\mbox{.}(2014)]%
        {sun2014data}
\bibfield{author}{\bibinfo{person}{Yunchuan Sun}, \bibinfo{person}{Junsheng Zhang}, \bibinfo{person}{Yongping Xiong}, {and} \bibinfo{person}{Guangyu Zhu}.} \bibinfo{year}{2014}\natexlab{}.
\newblock \showarticletitle{Data security and privacy in cloud computing}.
\newblock \bibinfo{journal}{\emph{International Journal of Distributed Sensor Networks}} \bibinfo{volume}{10}, \bibinfo{number}{7} (\bibinfo{year}{2014}), \bibinfo{pages}{190903}.
\newblock


\bibitem[Sung et~al\mbox{.}(2022a)]%
        {sung2022lst}
\bibfield{author}{\bibinfo{person}{Yi-Lin Sung}, \bibinfo{person}{Jaemin Cho}, {and} \bibinfo{person}{Mohit Bansal}.} \bibinfo{year}{2022}\natexlab{a}.
\newblock \showarticletitle{Lst: Ladder side-tuning for parameter and memory efficient transfer learning}. In \bibinfo{booktitle}{\emph{Proc. NeurIPS}}.
\newblock


\bibitem[Sung et~al\mbox{.}(2022b)]%
        {sung2022vl}
\bibfield{author}{\bibinfo{person}{Yi-Lin Sung}, \bibinfo{person}{Jaemin Cho}, {and} \bibinfo{person}{Mohit Bansal}.} \bibinfo{year}{2022}\natexlab{b}.
\newblock \showarticletitle{Vl-adapter: Parameter-efficient transfer learning for vision-and-language tasks}. In \bibinfo{booktitle}{\emph{Proc. CVPR}}. \bibinfo{pages}{5227--5237}.
\newblock


\bibitem[Szegedy et~al\mbox{.}(2015)]%
        {szegedy2015going}
\bibfield{author}{\bibinfo{person}{Christian Szegedy}, \bibinfo{person}{Wei Liu}, \bibinfo{person}{Yangqing Jia}, \bibinfo{person}{Pierre Sermanet}, \bibinfo{person}{Scott Reed}, \bibinfo{person}{Dragomir Anguelov}, \bibinfo{person}{Dumitru Erhan}, \bibinfo{person}{Vincent Vanhoucke}, {and} \bibinfo{person}{Andrew Rabinovich}.} \bibinfo{year}{2015}\natexlab{}.
\newblock \showarticletitle{Going deeper with convolutions}. In \bibinfo{booktitle}{\emph{Proc. CVPR}}. \bibinfo{pages}{1--9}.
\newblock


\bibitem[Szegedy et~al\mbox{.}(2016)]%
        {szegedy2016rethinking}
\bibfield{author}{\bibinfo{person}{Christian Szegedy}, \bibinfo{person}{Vincent Vanhoucke}, \bibinfo{person}{Sergey Ioffe}, \bibinfo{person}{Jon Shlens}, {and} \bibinfo{person}{Zbigniew Wojna}.} \bibinfo{year}{2016}\natexlab{}.
\newblock \showarticletitle{Rethinking the inception architecture for computer vision}. In \bibinfo{booktitle}{\emph{Proc. CVPR}}. \bibinfo{pages}{2818--2826}.
\newblock


\bibitem[Tan et~al\mbox{.}(2018)]%
        {tan2018survey}
\bibfield{author}{\bibinfo{person}{Chuanqi Tan}, \bibinfo{person}{Fuchun Sun}, \bibinfo{person}{Tao Kong}, \bibinfo{person}{Wenchang Zhang}, \bibinfo{person}{Chao Yang}, {and} \bibinfo{person}{Chunfang Liu}.} \bibinfo{year}{2018}\natexlab{}.
\newblock \showarticletitle{A survey on deep transfer learning}. In \bibinfo{booktitle}{\emph{International conference on artificial neural networks}}. Springer, \bibinfo{pages}{270--279}.
\newblock


\bibitem[Tan and Le(2019)]%
        {tan2019efficientnet}
\bibfield{author}{\bibinfo{person}{Mingxing Tan} {and} \bibinfo{person}{Quoc Le}.} \bibinfo{year}{2019}\natexlab{}.
\newblock \showarticletitle{Efficientnet: Rethinking model scaling for convolutional neural networks}. In \bibinfo{booktitle}{\emph{Proc. ICML}}. PMLR, \bibinfo{pages}{6105--6114}.
\newblock


\bibitem[Tao et~al\mbox{.}(2023)]%
        {tao2023galip}
\bibfield{author}{\bibinfo{person}{Ming Tao}, \bibinfo{person}{Bing-Kun Bao}, \bibinfo{person}{Hao Tang}, {and} \bibinfo{person}{Changsheng Xu}.} \bibinfo{year}{2023}\natexlab{}.
\newblock \showarticletitle{GALIP: Generative Adversarial CLIPs for Text-to-Image Synthesis}.
\newblock \bibinfo{journal}{\emph{Preprint at https://arxiv.org/abs/2301.12959}} (\bibinfo{year}{2023}).
\newblock


\bibitem[Tay et~al\mbox{.}(2022)]%
        {tay2022efficient}
\bibfield{author}{\bibinfo{person}{Yi Tay}, \bibinfo{person}{Mostafa Dehghani}, \bibinfo{person}{Dara Bahri}, {and} \bibinfo{person}{Donald Metzler}.} \bibinfo{year}{2022}\natexlab{}.
\newblock \showarticletitle{Efficient transformers: A survey}.
\newblock \bibinfo{journal}{\emph{Comput. Surveys}} \bibinfo{volume}{55}, \bibinfo{number}{6} (\bibinfo{year}{2022}), \bibinfo{pages}{1--28}.
\newblock


\bibitem[Tong et~al\mbox{.}(2022)]%
        {tong2022videomae}
\bibfield{author}{\bibinfo{person}{Zhan Tong}, \bibinfo{person}{Yibing Song}, \bibinfo{person}{Jue Wang}, {and} \bibinfo{person}{Limin Wang}.} \bibinfo{year}{2022}\natexlab{}.
\newblock \showarticletitle{Videomae: Masked autoencoders are data-efficient learners for self-supervised video pre-training}.
\newblock \bibinfo{journal}{\emph{Preprint at https://arxiv.org/abs/2203.12602}} (\bibinfo{year}{2022}).
\newblock


\bibitem[Touvron et~al\mbox{.}(2021)]%
        {touvron2021training}
\bibfield{author}{\bibinfo{person}{Hugo Touvron}, \bibinfo{person}{Matthieu Cord}, \bibinfo{person}{Matthijs Douze}, \bibinfo{person}{Francisco Massa}, \bibinfo{person}{Alexandre Sablayrolles}, {and} \bibinfo{person}{Herv{\'e} J{\'e}gou}.} \bibinfo{year}{2021}\natexlab{}.
\newblock \showarticletitle{Training data-efficient image transformers \& distillation through attention}. In \bibinfo{booktitle}{\emph{Proc. ICML}}. PMLR, \bibinfo{pages}{10347--10357}.
\newblock


\bibitem[Tran et~al\mbox{.}(2015)]%
        {tran2015learning}
\bibfield{author}{\bibinfo{person}{Du Tran}, \bibinfo{person}{Lubomir Bourdev}, \bibinfo{person}{Rob Fergus}, \bibinfo{person}{Lorenzo Torresani}, {and} \bibinfo{person}{Manohar Paluri}.} \bibinfo{year}{2015}\natexlab{}.
\newblock \showarticletitle{Learning spatiotemporal features with 3d convolutional networks}. In \bibinfo{booktitle}{\emph{Proc. ICCV}}. \bibinfo{pages}{4489--4497}.
\newblock


\bibitem[Tripuraneni et~al\mbox{.}(2020)]%
        {tripuraneni2020theory}
\bibfield{author}{\bibinfo{person}{Nilesh Tripuraneni}, \bibinfo{person}{Michael Jordan}, {and} \bibinfo{person}{Chi Jin}.} \bibinfo{year}{2020}\natexlab{}.
\newblock \showarticletitle{On the theory of transfer learning: The importance of task diversity}. In \bibinfo{booktitle}{\emph{Proc. NeurIPS}}, Vol.~\bibinfo{volume}{33}. \bibinfo{pages}{7852--7862}.
\newblock


\bibitem[Tsubota et~al\mbox{.}(2023)]%
        {tsubota2023universal}
\bibfield{author}{\bibinfo{person}{Koki Tsubota}, \bibinfo{person}{Hiroaki Akutsu}, {and} \bibinfo{person}{Kiyoharu Aizawa}.} \bibinfo{year}{2023}\natexlab{}.
\newblock \showarticletitle{Universal Deep Image Compression via Content-Adaptive Optimization with Adapters}. In \bibinfo{booktitle}{\emph{Proc. WACV}}. \bibinfo{pages}{2529--2538}.
\newblock


\bibitem[Tu et~al\mbox{.}(2022)]%
        {tu2022visual}
\bibfield{author}{\bibinfo{person}{Cheng-Hao Tu}, \bibinfo{person}{Zheda Mai}, {and} \bibinfo{person}{Wei-Lun Chao}.} \bibinfo{year}{2022}\natexlab{}.
\newblock \showarticletitle{Visual Query Tuning: Towards Effective Usage of Intermediate Representations for Parameter and Memory Efficient Transfer Learning}.
\newblock \bibinfo{journal}{\emph{Preprint at https://arxiv.org/abs/2212.03220}} (\bibinfo{year}{2022}).
\newblock


\bibitem[Usynin et~al\mbox{.}(2021)]%
        {usynin2021adversarial}
\bibfield{author}{\bibinfo{person}{Dmitrii Usynin}, \bibinfo{person}{Alexander Ziller}, \bibinfo{person}{Marcus Makowski}, \bibinfo{person}{Rickmer Braren}, \bibinfo{person}{Daniel Rueckert}, \bibinfo{person}{Ben Glocker}, \bibinfo{person}{Georgios Kaissis}, {and} \bibinfo{person}{Jonathan Passerat-Palmbach}.} \bibinfo{year}{2021}\natexlab{}.
\newblock \showarticletitle{Adversarial interference and its mitigations in privacy-preserving collaborative machine learning}.
\newblock \bibinfo{journal}{\emph{Nature Machine Intelligence}} \bibinfo{volume}{3}, \bibinfo{number}{9} (\bibinfo{year}{2021}), \bibinfo{pages}{749--758}.
\newblock


\bibitem[Valipour et~al\mbox{.}(2022)]%
        {valipour2022dylora}
\bibfield{author}{\bibinfo{person}{Mojtaba Valipour}, \bibinfo{person}{Mehdi Rezagholizadeh}, \bibinfo{person}{Ivan Kobyzev}, {and} \bibinfo{person}{Ali Ghodsi}.} \bibinfo{year}{2022}\natexlab{}.
\newblock \showarticletitle{DyLoRA: Parameter Efficient Tuning of Pre-trained Models using Dynamic Search-Free Low-Rank Adaptation}.
\newblock \bibinfo{journal}{\emph{Preprint at https://arxiv.org/abs/2210.07558}} (\bibinfo{year}{2022}).
\newblock


\bibitem[Vemprala et~al\mbox{.}(2023)]%
        {vemprala2023chatgpt}
\bibfield{author}{\bibinfo{person}{Sai Vemprala}, \bibinfo{person}{Rogerio Bonatti}, \bibinfo{person}{Arthur Bucker}, {and} \bibinfo{person}{Ashish Kapoor}.} \bibinfo{year}{2023}\natexlab{}.
\newblock \showarticletitle{Chatgpt for robotics: Design principles and model abilities}.
\newblock  (\bibinfo{year}{2023}).
\newblock


\bibitem[Verkuil et~al\mbox{.}(2022)]%
        {verkuil2022language}
\bibfield{author}{\bibinfo{person}{Robert Verkuil}, \bibinfo{person}{Ori Kabeli}, \bibinfo{person}{Yilun Du}, \bibinfo{person}{Basile~IM Wicky}, \bibinfo{person}{Lukas~F Milles}, \bibinfo{person}{Justas Dauparas}, \bibinfo{person}{David Baker}, \bibinfo{person}{Sergey Ovchinnikov}, \bibinfo{person}{Tom Sercu}, {and} \bibinfo{person}{Alexander Rives}.} \bibinfo{year}{2022}\natexlab{}.
\newblock \showarticletitle{Language models generalize beyond natural proteins}.
\newblock \bibinfo{journal}{\emph{bioRxiv}} (\bibinfo{year}{2022}), \bibinfo{pages}{2022--12}.
\newblock


\bibitem[Wang et~al\mbox{.}(2022d)]%
        {wang2022learning}
\bibfield{author}{\bibinfo{person}{Feng Wang}, \bibinfo{person}{Manling Li}, \bibinfo{person}{Xudong Lin}, \bibinfo{person}{Hairong Lv}, \bibinfo{person}{Alexander~G Schwing}, {and} \bibinfo{person}{Heng Ji}.} \bibinfo{year}{2022}\natexlab{d}.
\newblock \showarticletitle{Learning to Decompose Visual Features with Latent Textual Prompts}.
\newblock \bibinfo{journal}{\emph{Preprint at https://arxiv.org/abs/2210.04287}} (\bibinfo{year}{2022}).
\newblock


\bibitem[Wang et~al\mbox{.}(2023b)]%
        {wang2023lion}
\bibfield{author}{\bibinfo{person}{Haixin Wang}, \bibinfo{person}{Jianlong Chang}, \bibinfo{person}{Xiao Luo}, \bibinfo{person}{Jinan Sun}, \bibinfo{person}{Zhouchen Lin}, {and} \bibinfo{person}{Qi Tian}.} \bibinfo{year}{2023}\natexlab{b}.
\newblock \showarticletitle{LION: Implicit Vision Prompt Tuning}.
\newblock \bibinfo{journal}{\emph{Preprint at https://arxiv.org/abs/2303.09992}} (\bibinfo{year}{2023}).
\newblock


\bibitem[Wang et~al\mbox{.}(2023c)]%
        {wang2023mode}
\bibfield{author}{\bibinfo{person}{Haixin Wang}, \bibinfo{person}{Xinlong Yang}, \bibinfo{person}{Jianlong Chang}, \bibinfo{person}{Dian Jin}, \bibinfo{person}{Jinan Sun}, \bibinfo{person}{Shikun Zhang}, \bibinfo{person}{Xiao Luo}, {and} \bibinfo{person}{Qi Tian}.} \bibinfo{year}{2023}\natexlab{c}.
\newblock \showarticletitle{Mode Approximation Makes Good Vision-Language Prompts}.
\newblock \bibinfo{journal}{\emph{arXiv preprint arXiv:2305.08381}} (\bibinfo{year}{2023}).
\newblock


\bibitem[Wang et~al\mbox{.}(2020a)]%
        {wang2020stacking}
\bibfield{author}{\bibinfo{person}{Haixin Wang}, \bibinfo{person}{Tianhao Zhang}, \bibinfo{person}{Muzhi Yu}, \bibinfo{person}{Jinan Sun}, \bibinfo{person}{Wei Ye}, \bibinfo{person}{Chen Wang}, {and} \bibinfo{person}{Shikun Zhang}.} \bibinfo{year}{2020}\natexlab{a}.
\newblock \showarticletitle{Stacking networks dynamically for image restoration based on the Plug-and-Play framework}. In \bibinfo{booktitle}{\emph{Proc. ECCV}}. Springer, \bibinfo{pages}{446--462}.
\newblock


\bibitem[Wang et~al\mbox{.}(2023a)]%
        {DBLP:conf/cvpr/WangCLWO023}
\bibfield{author}{\bibinfo{person}{Shijie Wang}, \bibinfo{person}{Jianlong Chang}, \bibinfo{person}{Haojie Li}, \bibinfo{person}{Zhihui Wang}, \bibinfo{person}{Wanli Ouyang}, {and} \bibinfo{person}{Qi Tian}.} \bibinfo{year}{2023}\natexlab{a}.
\newblock \showarticletitle{Open-Set Fine-Grained Retrieval via Prompting Vision-Language Evaluator}. In \bibinfo{booktitle}{\emph{{IEEE/CVF} Conference on Computer Vision and Pattern Recognition, {CVPR} 2023, Vancouver, BC, Canada, June 17-24, 2023}}. \bibinfo{publisher}{{IEEE}}, \bibinfo{pages}{19381--19391}.
\newblock


\bibitem[Wang et~al\mbox{.}(2022b)]%
        {wang2022fine}
\bibfield{author}{\bibinfo{person}{Shijie Wang}, \bibinfo{person}{Jianlong Chang}, \bibinfo{person}{Zhihui Wang}, \bibinfo{person}{Haojie Li}, \bibinfo{person}{Wanli Ouyang}, {and} \bibinfo{person}{Qi Tian}.} \bibinfo{year}{2022}\natexlab{b}.
\newblock \showarticletitle{Fine-grained Retrieval Prompt Tuning}.
\newblock \bibinfo{journal}{\emph{Preprint at https://arxiv.org/abs/2207.14465}} (\bibinfo{year}{2022}).
\newblock


\bibitem[Wang et~al\mbox{.}(2022a)]%
        {wang2022image}
\bibfield{author}{\bibinfo{person}{Wenhui Wang}, \bibinfo{person}{Hangbo Bao}, \bibinfo{person}{Li Dong}, \bibinfo{person}{Johan Bjorck}, \bibinfo{person}{Zhiliang Peng}, \bibinfo{person}{Qiang Liu}, \bibinfo{person}{Kriti Aggarwal}, \bibinfo{person}{Owais~Khan Mohammed}, \bibinfo{person}{Saksham Singhal}, \bibinfo{person}{Subhojit Som}, {et~al\mbox{.}}} \bibinfo{year}{2022}\natexlab{a}.
\newblock \showarticletitle{Image as a foreign language: Beit pretraining for all vision and vision-language tasks}.
\newblock \bibinfo{journal}{\emph{Preprint at https://arxiv.org/abs/2208.10442}} (\bibinfo{year}{2022}).
\newblock


\bibitem[Wang et~al\mbox{.}(2021b)]%
        {wang2021pyramid}
\bibfield{author}{\bibinfo{person}{Wenhai Wang}, \bibinfo{person}{Enze Xie}, \bibinfo{person}{Xiang Li}, \bibinfo{person}{Deng-Ping Fan}, \bibinfo{person}{Kaitao Song}, \bibinfo{person}{Ding Liang}, \bibinfo{person}{Tong Lu}, \bibinfo{person}{Ping Luo}, {and} \bibinfo{person}{Ling Shao}.} \bibinfo{year}{2021}\natexlab{b}.
\newblock \showarticletitle{Pyramid vision transformer: A versatile backbone for dense prediction without convolutions}. In \bibinfo{booktitle}{\emph{Proc. CVPR}}. \bibinfo{pages}{568--578}.
\newblock


\bibitem[Wang et~al\mbox{.}(2022e)]%
        {wang2022images}
\bibfield{author}{\bibinfo{person}{Xinlong Wang}, \bibinfo{person}{Wen Wang}, \bibinfo{person}{Yue Cao}, \bibinfo{person}{Chunhua Shen}, {and} \bibinfo{person}{Tiejun Huang}.} \bibinfo{year}{2022}\natexlab{e}.
\newblock \showarticletitle{Images Speak in Images: A Generalist Painter for In-Context Visual Learning}.
\newblock \bibinfo{journal}{\emph{Preprint at https://arxiv.org/abs/2212.02499}} (\bibinfo{year}{2022}).
\newblock


\bibitem[Wang et~al\mbox{.}(2022c)]%
        {wang2022s}
\bibfield{author}{\bibinfo{person}{Yabin Wang}, \bibinfo{person}{Zhiwu Huang}, {and} \bibinfo{person}{Xiaopeng Hong}.} \bibinfo{year}{2022}\natexlab{c}.
\newblock \showarticletitle{S-Prompts Learning with Pre-trained Transformers: An Occam's Razor for Domain Incremental Learning}.
\newblock \bibinfo{journal}{\emph{Preprint at https://arxiv.org/abs/2207.12819}} (\bibinfo{year}{2022}).
\newblock


\bibitem[Wang et~al\mbox{.}(2021a)]%
        {wang2021knowledge}
\bibfield{author}{\bibinfo{person}{Yiran Wang}, \bibinfo{person}{Xingyi Li}, \bibinfo{person}{Min Shi}, \bibinfo{person}{Ke Xian}, {and} \bibinfo{person}{Zhiguo Cao}.} \bibinfo{year}{2021}\natexlab{a}.
\newblock \showarticletitle{Knowledge distillation for fast and accurate monocular depth estimation on mobile devices}. In \bibinfo{booktitle}{\emph{Proc. CVPR}}. \bibinfo{pages}{2457--2465}.
\newblock


\bibitem[Wang et~al\mbox{.}(2020b)]%
        {wang2020intra}
\bibfield{author}{\bibinfo{person}{Yukang Wang}, \bibinfo{person}{Wei Zhou}, \bibinfo{person}{Tao Jiang}, \bibinfo{person}{Xiang Bai}, {and} \bibinfo{person}{Yongchao Xu}.} \bibinfo{year}{2020}\natexlab{b}.
\newblock \showarticletitle{Intra-class feature variation distillation for semantic segmentation}. In \bibinfo{booktitle}{\emph{Proc. ECCV}}. Springer, \bibinfo{pages}{346--362}.
\newblock


\bibitem[Wang et~al\mbox{.}(2022f)]%
        {wang2022p2p}
\bibfield{author}{\bibinfo{person}{Ziyi Wang}, \bibinfo{person}{Xumin Yu}, \bibinfo{person}{Yongming Rao}, \bibinfo{person}{Jie Zhou}, {and} \bibinfo{person}{Jiwen Lu}.} \bibinfo{year}{2022}\natexlab{f}.
\newblock \showarticletitle{P2p: Tuning pre-trained image models for point cloud analysis with point-to-pixel prompting}.
\newblock \bibinfo{journal}{\emph{Preprint at https://arxiv.org/abs/2208.02812}} (\bibinfo{year}{2022}).
\newblock


\bibitem[Wen et~al\mbox{.}(2016)]%
        {wen2016learning}
\bibfield{author}{\bibinfo{person}{Wei Wen}, \bibinfo{person}{Chunpeng Wu}, \bibinfo{person}{Yandan Wang}, \bibinfo{person}{Yiran Chen}, {and} \bibinfo{person}{Hai Li}.} \bibinfo{year}{2016}\natexlab{}.
\newblock \showarticletitle{Learning structured sparsity in deep neural networks}. In \bibinfo{booktitle}{\emph{Proc. NeurIPS}}, Vol.~\bibinfo{volume}{29}.
\newblock


\bibitem[Wu et~al\mbox{.}(2023b)]%
        {wu2023visual}
\bibfield{author}{\bibinfo{person}{Chenfei Wu}, \bibinfo{person}{Shengming Yin}, \bibinfo{person}{Weizhen Qi}, \bibinfo{person}{Xiaodong Wang}, \bibinfo{person}{Zecheng Tang}, {and} \bibinfo{person}{Nan Duan}.} \bibinfo{year}{2023}\natexlab{b}.
\newblock \showarticletitle{Visual ChatGPT: Talking, Drawing and Editing with Visual Foundation Models}.
\newblock \bibinfo{journal}{\emph{Preprint at https://arxiv.org/abs/2303.04671}} (\bibinfo{year}{2023}).
\newblock


\bibitem[Wu et~al\mbox{.}(2022b)]%
        {wu2022generative}
\bibfield{author}{\bibinfo{person}{Chen~Henry Wu}, \bibinfo{person}{Saman Motamed}, \bibinfo{person}{Shaunak Srivastava}, {and} \bibinfo{person}{Fernando De~la Torre}.} \bibinfo{year}{2022}\natexlab{b}.
\newblock \showarticletitle{Generative visual prompt: Unifying distributional control of pre-trained generative models}. In \bibinfo{booktitle}{\emph{Proc. NeurIPS}}.
\newblock


\bibitem[Wu and Chen(2022)]%
        {wu2022pruning}
\bibfield{author}{\bibinfo{person}{Jiarun Wu} {and} \bibinfo{person}{Qingliang Chen}.} \bibinfo{year}{2022}\natexlab{}.
\newblock \showarticletitle{Pruning Adapters with Lottery Ticket}.
\newblock \bibinfo{journal}{\emph{Algorithms}} \bibinfo{volume}{15}, \bibinfo{number}{2} (\bibinfo{year}{2022}), \bibinfo{pages}{63}.
\newblock


\bibitem[Wu et~al\mbox{.}(2022a)]%
        {wu2022unleashing}
\bibfield{author}{\bibinfo{person}{Junyang Wu}, \bibinfo{person}{Xianhang Li}, \bibinfo{person}{Chen Wei}, \bibinfo{person}{Huiyu Wang}, \bibinfo{person}{Alan Yuille}, \bibinfo{person}{Yuyin Zhou}, {and} \bibinfo{person}{Cihang Xie}.} \bibinfo{year}{2022}\natexlab{a}.
\newblock \showarticletitle{Unleashing the Power of Visual Prompting At the Pixel Level}.
\newblock \bibinfo{journal}{\emph{Preprint at https://arxiv.org/abs/2212.10556}} (\bibinfo{year}{2022}).
\newblock


\bibitem[Wu et~al\mbox{.}(2023a)]%
        {wu2023tune}
\bibfield{author}{\bibinfo{person}{Jay~Zhangjie Wu}, \bibinfo{person}{Yixiao Ge}, \bibinfo{person}{Xintao Wang}, \bibinfo{person}{Stan~Weixian Lei}, \bibinfo{person}{Yuchao Gu}, \bibinfo{person}{Yufei Shi}, \bibinfo{person}{Wynne Hsu}, \bibinfo{person}{Ying Shan}, \bibinfo{person}{Xiaohu Qie}, {and} \bibinfo{person}{Mike~Zheng Shou}.} \bibinfo{year}{2023}\natexlab{a}.
\newblock \showarticletitle{Tune-a-video: One-shot tuning of image diffusion models for text-to-video generation}. In \bibinfo{booktitle}{\emph{Proc. CVPR}}. \bibinfo{pages}{7623--7633}.
\newblock


\bibitem[Xie et~al\mbox{.}(2021)]%
        {xie2021weight}
\bibfield{author}{\bibinfo{person}{Lingxi Xie}, \bibinfo{person}{Xin Chen}, \bibinfo{person}{Kaifeng Bi}, \bibinfo{person}{Longhui Wei}, \bibinfo{person}{Yuhui Xu}, \bibinfo{person}{Lanfei Wang}, \bibinfo{person}{Zhengsu Chen}, \bibinfo{person}{An Xiao}, \bibinfo{person}{Jianlong Chang}, \bibinfo{person}{Xiaopeng Zhang}, {et~al\mbox{.}}} \bibinfo{year}{2021}\natexlab{}.
\newblock \showarticletitle{Weight-sharing neural architecture search: A battle to shrink the optimization gap}.
\newblock \bibinfo{journal}{\emph{ACM Computing Surveys (CSUR)}} \bibinfo{volume}{54}, \bibinfo{number}{9} (\bibinfo{year}{2021}), \bibinfo{pages}{1--37}.
\newblock


\bibitem[Xie et~al\mbox{.}(2018)]%
        {xie2018rethinking}
\bibfield{author}{\bibinfo{person}{Saining Xie}, \bibinfo{person}{Chen Sun}, \bibinfo{person}{Jonathan Huang}, \bibinfo{person}{Zhuowen Tu}, {and} \bibinfo{person}{Kevin Murphy}.} \bibinfo{year}{2018}\natexlab{}.
\newblock \showarticletitle{Rethinking spatiotemporal feature learning: Speed-accuracy trade-offs in video classification}. In \bibinfo{booktitle}{\emph{Proc. ECCV}}. \bibinfo{pages}{305--321}.
\newblock


\bibitem[Xie et~al\mbox{.}(2019)]%
        {xie2019snas}
\bibfield{author}{\bibinfo{person}{Sirui Xie}, \bibinfo{person}{Hehui Zheng}, \bibinfo{person}{Chunxiao Liu}, {and} \bibinfo{person}{Liang Lin}.} \bibinfo{year}{2019}\natexlab{}.
\newblock \showarticletitle{SNAS: stochastic neural architecture search}. In \bibinfo{booktitle}{\emph{Proc. ICLR}}.
\newblock


\bibitem[Xing et~al\mbox{.}(2022)]%
        {xing2022class}
\bibfield{author}{\bibinfo{person}{Yinghui Xing}, \bibinfo{person}{Qirui Wu}, \bibinfo{person}{De Cheng}, \bibinfo{person}{Shizhou Zhang}, \bibinfo{person}{Guoqiang Liang}, {and} \bibinfo{person}{Yanning Zhang}.} \bibinfo{year}{2022}\natexlab{}.
\newblock \showarticletitle{Class-aware visual prompt tuning for vision-language pre-trained model}.
\newblock \bibinfo{journal}{\emph{Preprint at https://arxiv.org/abs/2208.08340}} (\bibinfo{year}{2022}).
\newblock


\bibitem[Xu et~al\mbox{.}(2023a)]%
        {xu2023exploring}
\bibfield{author}{\bibinfo{person}{Chengming Xu}, \bibinfo{person}{Siqian Yang}, \bibinfo{person}{Yabiao Wang}, \bibinfo{person}{Zhanxiong Wang}, \bibinfo{person}{Yanwei Fu}, {and} \bibinfo{person}{Xiangyang Xue}.} \bibinfo{year}{2023}\natexlab{a}.
\newblock \showarticletitle{Exploring Efficient Few-shot Adaptation for Vision Transformers}.
\newblock \bibinfo{journal}{\emph{Preprint at https://arxiv.org/abs/2301.02419}} (\bibinfo{year}{2023}).
\newblock


\bibitem[Xu et~al\mbox{.}(2020a)]%
        {xu2020feature}
\bibfield{author}{\bibinfo{person}{Kunran Xu}, \bibinfo{person}{Lai Rui}, \bibinfo{person}{Yishi Li}, {and} \bibinfo{person}{Lin Gu}.} \bibinfo{year}{2020}\natexlab{a}.
\newblock \showarticletitle{Feature normalized knowledge distillation for image classification}. In \bibinfo{booktitle}{\emph{Proc. ECCV}}. Springer, \bibinfo{pages}{664--680}.
\newblock


\bibitem[Xu et~al\mbox{.}(2023b)]%
        {xu2023side}
\bibfield{author}{\bibinfo{person}{Mengde Xu}, \bibinfo{person}{Zheng Zhang}, \bibinfo{person}{Fangyun Wei}, \bibinfo{person}{Han Hu}, {and} \bibinfo{person}{Xiang Bai}.} \bibinfo{year}{2023}\natexlab{b}.
\newblock \showarticletitle{Side Adapter Network for Open-Vocabulary Semantic Segmentation}.
\newblock \bibinfo{journal}{\emph{Preprint at https://arxiv.org/abs/2302.12242}} (\bibinfo{year}{2023}).
\newblock


\bibitem[Xu et~al\mbox{.}(2020b)]%
        {xu2020pc}
\bibfield{author}{\bibinfo{person}{Yuhui Xu}, \bibinfo{person}{Lingxi Xie}, \bibinfo{person}{Xiaopeng Zhang}, \bibinfo{person}{Xin Chen}, \bibinfo{person}{Guo-Jun Qi}, \bibinfo{person}{Qi Tian}, {and} \bibinfo{person}{Hongkai Xiong}.} \bibinfo{year}{2020}\natexlab{b}.
\newblock \showarticletitle{Pc-darts: Partial channel connections for memory-efficient architecture search}. In \bibinfo{booktitle}{\emph{Proc. ICLR}}.
\newblock


\bibitem[Yan et~al\mbox{.}(2018)]%
        {yan2018spatial}
\bibfield{author}{\bibinfo{person}{Sijie Yan}, \bibinfo{person}{Yuanjun Xiong}, {and} \bibinfo{person}{Dahua Lin}.} \bibinfo{year}{2018}\natexlab{}.
\newblock \showarticletitle{Spatial temporal graph convolutional networks for skeleton-based action recognition}. In \bibinfo{booktitle}{\emph{Proc. AAAI}}, Vol.~\bibinfo{volume}{32}.
\newblock


\bibitem[Yang et~al\mbox{.}(2022a)]%
        {yang2022cross}
\bibfield{author}{\bibinfo{person}{Chuanguang Yang}, \bibinfo{person}{Helong Zhou}, \bibinfo{person}{Zhulin An}, \bibinfo{person}{Xue Jiang}, \bibinfo{person}{Yongjun Xu}, {and} \bibinfo{person}{Qian Zhang}.} \bibinfo{year}{2022}\natexlab{a}.
\newblock \showarticletitle{Cross-image relational knowledge distillation for semantic segmentation}. In \bibinfo{booktitle}{\emph{Proc. CVPR}}. \bibinfo{pages}{12319--12328}.
\newblock


\bibitem[Yang et~al\mbox{.}(2020)]%
        {yang2020knowledge}
\bibfield{author}{\bibinfo{person}{Jing Yang}, \bibinfo{person}{Brais Martinez}, \bibinfo{person}{Adrian Bulat}, {and} \bibinfo{person}{Georgios Tzimiropoulos}.} \bibinfo{year}{2020}\natexlab{}.
\newblock \showarticletitle{Knowledge distillation via adaptive instance normalization}.
\newblock \bibinfo{journal}{\emph{Preprint at https://arxiv.org/abs/2003.04289}} (\bibinfo{year}{2020}).
\newblock


\bibitem[Yang et~al\mbox{.}(2023a)]%
        {yang2023diffusion}
\bibfield{author}{\bibinfo{person}{Ling Yang}, \bibinfo{person}{Zhilong Zhang}, \bibinfo{person}{Yang Song}, \bibinfo{person}{Shenda Hong}, \bibinfo{person}{Runsheng Xu}, \bibinfo{person}{Yue Zhao}, \bibinfo{person}{Wentao Zhang}, \bibinfo{person}{Bin Cui}, {and} \bibinfo{person}{Ming-Hsuan Yang}.} \bibinfo{year}{2023}\natexlab{a}.
\newblock \showarticletitle{Diffusion models: A comprehensive survey of methods and applications}.
\newblock \bibinfo{journal}{\emph{Comput. Surveys}} \bibinfo{volume}{56}, \bibinfo{number}{4} (\bibinfo{year}{2023}), \bibinfo{pages}{1--39}.
\newblock


\bibitem[Yang et~al\mbox{.}(2023b)]%
        {yang2023aim}
\bibfield{author}{\bibinfo{person}{Taojiannan Yang}, \bibinfo{person}{Yi Zhu}, \bibinfo{person}{Yusheng Xie}, \bibinfo{person}{Aston Zhang}, \bibinfo{person}{Chen Chen}, {and} \bibinfo{person}{Mu Li}.} \bibinfo{year}{2023}\natexlab{b}.
\newblock \showarticletitle{{AIM}: Adapting Image Models for Efficient Video Action Recognition}. In \bibinfo{booktitle}{\emph{Proc. ICLR}}.
\newblock
\urldef\tempurl%
\url{https://openreview.net/forum?id=CIoSZ_HKHS7}
\showURL{%
\tempurl}


\bibitem[Yang et~al\mbox{.}(2022b)]%
        {yang2022deep}
\bibfield{author}{\bibinfo{person}{Xingyi Yang}, \bibinfo{person}{Daquan Zhou}, \bibinfo{person}{Songhua Liu}, \bibinfo{person}{Jingwen Ye}, {and} \bibinfo{person}{Xinchao Wang}.} \bibinfo{year}{2022}\natexlab{b}.
\newblock \showarticletitle{Deep model reassembly}. In \bibinfo{booktitle}{\emph{Proc. NeurIPS}}, Vol.~\bibinfo{volume}{35}. \bibinfo{pages}{25739--25753}.
\newblock


\bibitem[Ye et~al\mbox{.}(2021)]%
        {ye2021towards}
\bibfield{author}{\bibinfo{person}{Haotian Ye}, \bibinfo{person}{Chuanlong Xie}, \bibinfo{person}{Tianle Cai}, \bibinfo{person}{Ruichen Li}, \bibinfo{person}{Zhenguo Li}, {and} \bibinfo{person}{Liwei Wang}.} \bibinfo{year}{2021}\natexlab{}.
\newblock \showarticletitle{Towards a theoretical framework of out-of-distribution generalization}. In \bibinfo{booktitle}{\emph{Proc. NeurIPS}}, Vol.~\bibinfo{volume}{34}. \bibinfo{pages}{23519--23531}.
\newblock


\bibitem[Yu et~al\mbox{.}(2022a)]%
        {yu2022towards}
\bibfield{author}{\bibinfo{person}{Bruce~XB Yu}, \bibinfo{person}{Jianlong Chang}, \bibinfo{person}{Lingbo Liu}, \bibinfo{person}{Qi Tian}, {and} \bibinfo{person}{Chang~Wen Chen}.} \bibinfo{year}{2022}\natexlab{a}.
\newblock \showarticletitle{Towards a Unified View on Visual Parameter-Efficient Transfer Learning}.
\newblock \bibinfo{journal}{\emph{Preprint at https://arxiv.org/abs/2210.00788}} (\bibinfo{year}{2022}).
\newblock


\bibitem[Yu et~al\mbox{.}(2023c)]%
        {yu2023gla}
\bibfield{author}{\bibinfo{person}{Bruce~XB Yu}, \bibinfo{person}{Zhi Zhang}, \bibinfo{person}{Yongxu Liu}, \bibinfo{person}{Sheng-hua Zhong}, \bibinfo{person}{Yan Liu}, {and} \bibinfo{person}{Chang~Wen Chen}.} \bibinfo{year}{2023}\natexlab{c}.
\newblock \showarticletitle{Gla-gcn: Global-local adaptive graph convolutional network for 3d human pose estimation from monocular video}. In \bibinfo{booktitle}{\emph{Proc. ICCV}}. \bibinfo{pages}{8818--8829}.
\newblock


\bibitem[Yu et~al\mbox{.}(2022b)]%
        {yu2022coca}
\bibfield{author}{\bibinfo{person}{Jiahui Yu}, \bibinfo{person}{Zirui Wang}, \bibinfo{person}{Vijay Vasudevan}, \bibinfo{person}{Legg Yeung}, \bibinfo{person}{Mojtaba Seyedhosseini}, {and} \bibinfo{person}{Yonghui Wu}.} \bibinfo{year}{2022}\natexlab{b}.
\newblock \showarticletitle{Coca: Contrastive captioners are image-text foundation models}.
\newblock \bibinfo{journal}{\emph{Preprint at https://arxiv.org/abs/2205.01917}} (\bibinfo{year}{2022}).
\newblock


\bibitem[Yu et~al\mbox{.}(2023b)]%
        {yu2023prompting}
\bibfield{author}{\bibinfo{person}{Shengming Yu}, \bibinfo{person}{Zhaopeng Dou}, {and} \bibinfo{person}{Shengjin Wang}.} \bibinfo{year}{2023}\natexlab{b}.
\newblock \showarticletitle{Prompting and Tuning: A Two-Stage Unsupervised Domain Adaptive Person Re-identification Method on Vision Transformer Backbone}.
\newblock \bibinfo{journal}{\emph{Tsinghua Science and Technology}} \bibinfo{volume}{28}, \bibinfo{number}{4} (\bibinfo{year}{2023}), \bibinfo{pages}{799--810}.
\newblock


\bibitem[Yu et~al\mbox{.}(2023a)]%
        {yu2023rethinking}
\bibfield{author}{\bibinfo{person}{Zitong Yu}, \bibinfo{person}{Rizhao Cai}, \bibinfo{person}{Yawen Cui}, \bibinfo{person}{Xin Liu}, \bibinfo{person}{Yongjian Hu}, {and} \bibinfo{person}{Alex Kot}.} \bibinfo{year}{2023}\natexlab{a}.
\newblock \showarticletitle{Rethinking Vision Transformer and Masked Autoencoder in Multimodal Face Anti-Spoofing}.
\newblock \bibinfo{journal}{\emph{Preprint at https://arxiv.org/abs/2302.05744}} (\bibinfo{year}{2023}).
\newblock


\bibitem[Yuan et~al\mbox{.}(2021c)]%
        {yuan2021incorporating}
\bibfield{author}{\bibinfo{person}{Kun Yuan}, \bibinfo{person}{Shaopeng Guo}, \bibinfo{person}{Ziwei Liu}, \bibinfo{person}{Aojun Zhou}, \bibinfo{person}{Fengwei Yu}, {and} \bibinfo{person}{Wei Wu}.} \bibinfo{year}{2021}\natexlab{c}.
\newblock \showarticletitle{Incorporating convolution designs into visual transformers}. In \bibinfo{booktitle}{\emph{Proc. CVPR}}. \bibinfo{pages}{579--588}.
\newblock


\bibitem[Yuan et~al\mbox{.}(2021a)]%
        {yuan2021florence}
\bibfield{author}{\bibinfo{person}{Lu Yuan}, \bibinfo{person}{Dongdong Chen}, \bibinfo{person}{Yi-Ling Chen}, \bibinfo{person}{Noel Codella}, \bibinfo{person}{Xiyang Dai}, \bibinfo{person}{Jianfeng Gao}, \bibinfo{person}{Houdong Hu}, \bibinfo{person}{Xuedong Huang}, \bibinfo{person}{Boxin Li}, \bibinfo{person}{Chunyuan Li}, {et~al\mbox{.}}} \bibinfo{year}{2021}\natexlab{a}.
\newblock \showarticletitle{Florence: A new foundation model for computer vision}.
\newblock \bibinfo{journal}{\emph{Preprint at https://arxiv.org/abs/2111.11432}} (\bibinfo{year}{2021}).
\newblock


\bibitem[Yuan et~al\mbox{.}(2021b)]%
        {yuan2021tokens}
\bibfield{author}{\bibinfo{person}{Li Yuan}, \bibinfo{person}{Yunpeng Chen}, \bibinfo{person}{Tao Wang}, \bibinfo{person}{Weihao Yu}, \bibinfo{person}{Yujun Shi}, \bibinfo{person}{Zi-Hang Jiang}, \bibinfo{person}{Francis~EH Tay}, \bibinfo{person}{Jiashi Feng}, {and} \bibinfo{person}{Shuicheng Yan}.} \bibinfo{year}{2021}\natexlab{b}.
\newblock \showarticletitle{Tokens-to-token vit: Training vision transformers from scratch on imagenet}. In \bibinfo{booktitle}{\emph{Proc. CVPR}}. \bibinfo{pages}{558--567}.
\newblock


\bibitem[Yuan et~al\mbox{.}(2022a)]%
        {yuan2022volo}
\bibfield{author}{\bibinfo{person}{Li Yuan}, \bibinfo{person}{Qibin Hou}, \bibinfo{person}{Zihang Jiang}, \bibinfo{person}{Jiashi Feng}, {and} \bibinfo{person}{Shuicheng Yan}.} \bibinfo{year}{2022}\natexlab{a}.
\newblock \showarticletitle{Volo: Vision outlooker for visual recognition}.
\newblock \bibinfo{journal}{\emph{IEEE Trans. Patt. Anal. Mach. Intell.}} (\bibinfo{year}{2022}).
\newblock


\bibitem[Yuan et~al\mbox{.}(2022b)]%
        {yuan2022roadmap}
\bibfield{author}{\bibinfo{person}{Sha Yuan}, \bibinfo{person}{Hanyu Zhao}, \bibinfo{person}{Shuai Zhao}, \bibinfo{person}{Jiahong Leng}, \bibinfo{person}{Yangxiao Liang}, \bibinfo{person}{Xiaozhi Wang}, \bibinfo{person}{Jifan Yu}, \bibinfo{person}{Xin Lv}, \bibinfo{person}{Zhou Shao}, \bibinfo{person}{Jiaao He}, {et~al\mbox{.}}} \bibinfo{year}{2022}\natexlab{b}.
\newblock \showarticletitle{A Roadmap for Big Model}.
\newblock \bibinfo{journal}{\emph{Preprint at https://arxiv.org/abs/2203.14101}} (\bibinfo{year}{2022}).
\newblock


\bibitem[Zaken et~al\mbox{.}(2022)]%
        {zaken2022bitfit}
\bibfield{author}{\bibinfo{person}{Elad~Ben Zaken}, \bibinfo{person}{Yoav Goldberg}, {and} \bibinfo{person}{Shauli Ravfogel}.} \bibinfo{year}{2022}\natexlab{}.
\newblock \showarticletitle{BitFit: Simple Parameter-efficient Fine-tuning for Transformer-based Masked Language-models}. In \bibinfo{booktitle}{\emph{Proc. ACL (Volume 2: Short Papers)}}. \bibinfo{pages}{1--9}.
\newblock


\bibitem[Zang et~al\mbox{.}(2022)]%
        {zang2022unified}
\bibfield{author}{\bibinfo{person}{Yuhang Zang}, \bibinfo{person}{Wei Li}, \bibinfo{person}{Kaiyang Zhou}, \bibinfo{person}{Chen Huang}, {and} \bibinfo{person}{Chen~Change Loy}.} \bibinfo{year}{2022}\natexlab{}.
\newblock \showarticletitle{Unified vision and language prompt learning}.
\newblock \bibinfo{journal}{\emph{Preprint at https://arxiv.org/abs/2210.07225}} (\bibinfo{year}{2022}).
\newblock


\bibitem[Zhang et~al\mbox{.}(2021d)]%
        {zhangbeyond}
\bibfield{author}{\bibinfo{person}{Aston Zhang}, \bibinfo{person}{Yi Tay}, \bibinfo{person}{SHUAI Zhang}, \bibinfo{person}{Alvin Chan}, \bibinfo{person}{Anh~Tuan Luu}, \bibinfo{person}{Siu Hui}, {and} \bibinfo{person}{Jie Fu}.} \bibinfo{year}{2021}\natexlab{d}.
\newblock \showarticletitle{Beyond Fully-Connected Layers with Quaternions: Parameterization of Hypercomplex Multiplications with $1/n $ Parameters}. In \bibinfo{booktitle}{\emph{Proc. ICLR}}.
\newblock


\bibitem[Zhang et~al\mbox{.}(2023a)]%
        {zhang2023multimodal}
\bibfield{author}{\bibinfo{person}{Bowen Zhang}, \bibinfo{person}{Xiaojie Jin}, \bibinfo{person}{Weibo Gong}, \bibinfo{person}{Kai Xu}, \bibinfo{person}{Zhao Zhang}, \bibinfo{person}{Peng Wang}, \bibinfo{person}{Xiaohui Shen}, {and} \bibinfo{person}{Jiashi Feng}.} \bibinfo{year}{2023}\natexlab{a}.
\newblock \showarticletitle{Multimodal Video Adapter for Parameter Efficient Video Text Retrieval}.
\newblock \bibinfo{journal}{\emph{Preprint at https://arxiv.org/abs/2301.07868}} (\bibinfo{year}{2023}).
\newblock


\bibitem[Zhang et~al\mbox{.}(2022e)]%
        {zhang2022feature}
\bibfield{author}{\bibinfo{person}{Jian-Wei Zhang}, \bibinfo{person}{Yifan Sun}, \bibinfo{person}{Yi Yang}, {and} \bibinfo{person}{Wei Chen}.} \bibinfo{year}{2022}\natexlab{e}.
\newblock \showarticletitle{Feature-Proxy Transformer for Few-Shot Segmentation}.
\newblock \bibinfo{journal}{\emph{Preprint at https://arxiv.org/abs/2210.06908}} (\bibinfo{year}{2022}).
\newblock


\bibitem[Zhang and Ma(2021)]%
        {zhang2021improve}
\bibfield{author}{\bibinfo{person}{Linfeng Zhang} {and} \bibinfo{person}{Kaisheng Ma}.} \bibinfo{year}{2021}\natexlab{}.
\newblock \showarticletitle{Improve object detection with feature-based knowledge distillation: Towards accurate and efficient detectors}. In \bibinfo{booktitle}{\emph{Proc. ICLR}}.
\newblock


\bibitem[Zhang et~al\mbox{.}(2022a)]%
        {zhang2022collaboration}
\bibfield{author}{\bibinfo{person}{Renrui Zhang}, \bibinfo{person}{Hanqiu Deng}, \bibinfo{person}{Bohao Li}, \bibinfo{person}{Wei Zhang}, \bibinfo{person}{Hao Dong}, \bibinfo{person}{Hongsheng Li}, \bibinfo{person}{Peng Gao}, {and} \bibinfo{person}{Yu Qiao}.} \bibinfo{year}{2022}\natexlab{a}.
\newblock \showarticletitle{Collaboration of Pre-trained Models Makes Better Few-shot Learner}.
\newblock \bibinfo{journal}{\emph{Preprint at https://arxiv.org/abs/2209.12255}} (\bibinfo{year}{2022}).
\newblock


\bibitem[Zhang et~al\mbox{.}(2021c)]%
        {zhang2021tip}
\bibfield{author}{\bibinfo{person}{Renrui Zhang}, \bibinfo{person}{Rongyao Fang}, \bibinfo{person}{Wei Zhang}, \bibinfo{person}{Peng Gao}, \bibinfo{person}{Kunchang Li}, \bibinfo{person}{Jifeng Dai}, \bibinfo{person}{Yu Qiao}, {and} \bibinfo{person}{Hongsheng Li}.} \bibinfo{year}{2021}\natexlab{c}.
\newblock \showarticletitle{Tip-adapter: Training-free clip-adapter for better vision-language modeling}.
\newblock \bibinfo{journal}{\emph{Preprint at https://arxiv.org/abs/2111.03930}} (\bibinfo{year}{2021}).
\newblock


\bibitem[Zhang et~al\mbox{.}(2022d)]%
        {zhang2022pointclip}
\bibfield{author}{\bibinfo{person}{Renrui Zhang}, \bibinfo{person}{Ziyu Guo}, \bibinfo{person}{Wei Zhang}, \bibinfo{person}{Kunchang Li}, \bibinfo{person}{Xupeng Miao}, \bibinfo{person}{Bin Cui}, \bibinfo{person}{Yu Qiao}, \bibinfo{person}{Peng Gao}, {and} \bibinfo{person}{Hongsheng Li}.} \bibinfo{year}{2022}\natexlab{d}.
\newblock \showarticletitle{Pointclip: Point cloud understanding by clip}. In \bibinfo{booktitle}{\emph{Proc. CVPR}}. \bibinfo{pages}{8552--8562}.
\newblock


\bibitem[Zhang et~al\mbox{.}(2021a)]%
        {DBLP:journals/pami/ZhangCGMXLP21}
\bibfield{author}{\bibinfo{person}{Xinbang Zhang}, \bibinfo{person}{Jianlong Chang}, \bibinfo{person}{Yiwen Guo}, \bibinfo{person}{Gaofeng Meng}, \bibinfo{person}{Shiming Xiang}, \bibinfo{person}{Zhouchen Lin}, {and} \bibinfo{person}{Chunhong Pan}.} \bibinfo{year}{2021}\natexlab{a}.
\newblock \showarticletitle{{DATA:} Differentiable ArchiTecture Approximation With Distribution Guided Sampling}.
\newblock \bibinfo{journal}{\emph{{IEEE} Trans. Pattern Anal. Mach. Intell.}} \bibinfo{volume}{43}, \bibinfo{number}{9} (\bibinfo{year}{2021}), \bibinfo{pages}{2905--2920}.
\newblock


\bibitem[Zhang et~al\mbox{.}(2021b)]%
        {zhang2021data}
\bibfield{author}{\bibinfo{person}{Yiman Zhang}, \bibinfo{person}{Hanting Chen}, \bibinfo{person}{Xinghao Chen}, \bibinfo{person}{Yiping Deng}, \bibinfo{person}{Chunjing Xu}, {and} \bibinfo{person}{Yunhe Wang}.} \bibinfo{year}{2021}\natexlab{b}.
\newblock \showarticletitle{Data-free knowledge distillation for image super-resolution}. In \bibinfo{booktitle}{\emph{Proc. CVPR}}. \bibinfo{pages}{7852--7861}.
\newblock


\bibitem[Zhang et~al\mbox{.}(2022b)]%
        {zhang2022prompting}
\bibfield{author}{\bibinfo{person}{Yue Zhang}, \bibinfo{person}{Hongliang Fei}, \bibinfo{person}{Dingcheng Li}, \bibinfo{person}{Tan Yu}, {and} \bibinfo{person}{Ping Li}.} \bibinfo{year}{2022}\natexlab{b}.
\newblock \showarticletitle{Prompting through Prototype: A Prototype-based Prompt Learning on Pretrained Vision-Language Models}.
\newblock \bibinfo{journal}{\emph{Preprint at https://arxiv.org/abs/2210.10841}} (\bibinfo{year}{2022}).
\newblock


\bibitem[Zhang et~al\mbox{.}(2019)]%
        {zhang2019bridging}
\bibfield{author}{\bibinfo{person}{Yuchen Zhang}, \bibinfo{person}{Tianle Liu}, \bibinfo{person}{Mingsheng Long}, {and} \bibinfo{person}{Michael Jordan}.} \bibinfo{year}{2019}\natexlab{}.
\newblock \showarticletitle{Bridging theory and algorithm for domain adaptation}. In \bibinfo{booktitle}{\emph{Proc. ICML}}. PMLR, \bibinfo{pages}{7404--7413}.
\newblock


\bibitem[Zhang and Yang(2021)]%
        {zhang2021survey}
\bibfield{author}{\bibinfo{person}{Yu Zhang} {and} \bibinfo{person}{Qiang Yang}.} \bibinfo{year}{2021}\natexlab{}.
\newblock \showarticletitle{A survey on multi-task learning}.
\newblock \bibinfo{journal}{\emph{IEEE Trans. Know. and Data Engin.}} (\bibinfo{year}{2021}).
\newblock


\bibitem[Zhang et~al\mbox{.}(2022f)]%
        {zhang2022neural}
\bibfield{author}{\bibinfo{person}{Yuanhan Zhang}, \bibinfo{person}{Kaiyang Zhou}, {and} \bibinfo{person}{Ziwei Liu}.} \bibinfo{year}{2022}\natexlab{f}.
\newblock \showarticletitle{Neural Prompt Search}.
\newblock \bibinfo{journal}{\emph{Preprint at https://arxiv.org/abs/2206.04673}} (\bibinfo{year}{2022}).
\newblock


\bibitem[Zhang et~al\mbox{.}(2023b)]%
        {zhang2023makes}
\bibfield{author}{\bibinfo{person}{Yuanhan Zhang}, \bibinfo{person}{Kaiyang Zhou}, {and} \bibinfo{person}{Ziwei Liu}.} \bibinfo{year}{2023}\natexlab{b}.
\newblock \showarticletitle{What Makes Good Examples for Visual In-Context Learning?}
\newblock \bibinfo{journal}{\emph{Preprint at https://arxiv.org/abs/2301.13670}} (\bibinfo{year}{2023}).
\newblock


\bibitem[Zhang et~al\mbox{.}(2022c)]%
        {zhang2022hyperpelt}
\bibfield{author}{\bibinfo{person}{Zhengkun Zhang}, \bibinfo{person}{Wenya Guo}, \bibinfo{person}{Xiaojun Meng}, \bibinfo{person}{Yasheng Wang}, \bibinfo{person}{Yadao Wang}, \bibinfo{person}{Xin Jiang}, \bibinfo{person}{Qun Liu}, {and} \bibinfo{person}{Zhenglu Yang}.} \bibinfo{year}{2022}\natexlab{c}.
\newblock \showarticletitle{Hyperpelt: Unified parameter-efficient language model tuning for both language and vision-and-language tasks}.
\newblock \bibinfo{journal}{\emph{Preprint at https://arxiv.org/abs/2203.03878}} (\bibinfo{year}{2022}).
\newblock


\bibitem[Zhao et~al\mbox{.}(2022a)]%
        {zhao2022decoupled}
\bibfield{author}{\bibinfo{person}{Borui Zhao}, \bibinfo{person}{Quan Cui}, \bibinfo{person}{Renjie Song}, \bibinfo{person}{Yiyu Qiu}, {and} \bibinfo{person}{Jiajun Liang}.} \bibinfo{year}{2022}\natexlab{a}.
\newblock \showarticletitle{Decoupled knowledge distillation}. In \bibinfo{booktitle}{\emph{Proc. CVPR}}. \bibinfo{pages}{11953--11962}.
\newblock


\bibitem[Zhao et~al\mbox{.}(2022b)]%
        {zhao2022learning}
\bibfield{author}{\bibinfo{person}{Cairong Zhao}, \bibinfo{person}{Yubin Wang}, \bibinfo{person}{Xinyang Jiang}, \bibinfo{person}{Yifei Shen}, \bibinfo{person}{Kaitao Song}, \bibinfo{person}{Dongsheng Li}, {and} \bibinfo{person}{Duoqian Miao}.} \bibinfo{year}{2022}\natexlab{b}.
\newblock \showarticletitle{Learning Domain Invariant Prompt for Vision-Language Models}.
\newblock \bibinfo{journal}{\emph{Preprint at https://arxiv.org/abs/2212.04196}} (\bibinfo{year}{2022}).
\newblock


\bibitem[Zheng et~al\mbox{.}(2022)]%
        {zheng2022localization}
\bibfield{author}{\bibinfo{person}{Zhaohui Zheng}, \bibinfo{person}{Rongguang Ye}, \bibinfo{person}{Ping Wang}, \bibinfo{person}{Dongwei Ren}, \bibinfo{person}{Wangmeng Zuo}, \bibinfo{person}{Qibin Hou}, {and} \bibinfo{person}{Ming-Ming Cheng}.} \bibinfo{year}{2022}\natexlab{}.
\newblock \showarticletitle{Localization distillation for dense object detection}. In \bibinfo{booktitle}{\emph{Proc. CVPR}}. \bibinfo{pages}{9407--9416}.
\newblock


\bibitem[Zhou et~al\mbox{.}(2023)]%
        {zhou2023comprehensive}
\bibfield{author}{\bibinfo{person}{Ce Zhou}, \bibinfo{person}{Qian Li}, \bibinfo{person}{Chen Li}, \bibinfo{person}{Jun Yu}, \bibinfo{person}{Yixin Liu}, \bibinfo{person}{Guangjing Wang}, \bibinfo{person}{Kai Zhang}, \bibinfo{person}{Cheng Ji}, \bibinfo{person}{Qiben Yan}, \bibinfo{person}{Lifang He}, {et~al\mbox{.}}} \bibinfo{year}{2023}\natexlab{}.
\newblock \showarticletitle{A Comprehensive Survey on Pretrained Foundation Models: A History from BERT to ChatGPT}.
\newblock \bibinfo{journal}{\emph{Preprint at https://arxiv.org/abs/2302.09419}} (\bibinfo{year}{2023}).
\newblock


\bibitem[Zhou et~al\mbox{.}(2022a)]%
        {zhou2022conditional}
\bibfield{author}{\bibinfo{person}{Kaiyang Zhou}, \bibinfo{person}{Jingkang Yang}, \bibinfo{person}{Chen~Change Loy}, {and} \bibinfo{person}{Ziwei Liu}.} \bibinfo{year}{2022}\natexlab{a}.
\newblock \showarticletitle{Conditional prompt learning for vision-language models}. In \bibinfo{booktitle}{\emph{Proc. CVPR}}. \bibinfo{pages}{16816--16825}.
\newblock


\bibitem[Zhou et~al\mbox{.}(2022b)]%
        {zhou2022learning}
\bibfield{author}{\bibinfo{person}{Kaiyang Zhou}, \bibinfo{person}{Jingkang Yang}, \bibinfo{person}{Chen~Change Loy}, {and} \bibinfo{person}{Ziwei Liu}.} \bibinfo{year}{2022}\natexlab{b}.
\newblock \showarticletitle{Learning to prompt for vision-language models}.
\newblock \bibinfo{journal}{\emph{IJCV}} \bibinfo{volume}{130}, \bibinfo{number}{9} (\bibinfo{year}{2022}), \bibinfo{pages}{2337--2348}.
\newblock


\bibitem[Zhou et~al\mbox{.}(2021)]%
        {zhou2021distilling}
\bibfield{author}{\bibinfo{person}{Sheng Zhou}, \bibinfo{person}{Yucheng Wang}, \bibinfo{person}{Defang Chen}, \bibinfo{person}{Jiawei Chen}, \bibinfo{person}{Xin Wang}, \bibinfo{person}{Can Wang}, {and} \bibinfo{person}{Jiajun Bu}.} \bibinfo{year}{2021}\natexlab{}.
\newblock \showarticletitle{Distilling holistic knowledge with graph neural networks}. In \bibinfo{booktitle}{\emph{Proc. CVPR}}. \bibinfo{pages}{10387--10396}.
\newblock


\bibitem[Zhou et~al\mbox{.}(2022c)]%
        {zhou2022zegclip}
\bibfield{author}{\bibinfo{person}{Ziqin Zhou}, \bibinfo{person}{Bowen Zhang}, \bibinfo{person}{Yinjie Lei}, \bibinfo{person}{Lingqiao Liu}, {and} \bibinfo{person}{Yifan Liu}.} \bibinfo{year}{2022}\natexlab{c}.
\newblock \showarticletitle{ZegCLIP: Towards Adapting CLIP for Zero-shot Semantic Segmentation}.
\newblock \bibinfo{journal}{\emph{Preprint at https://arxiv.org/abs/2212.03588}} (\bibinfo{year}{2022}).
\newblock


\bibitem[Zhu et~al\mbox{.}(2022a)]%
        {zhu2022prompt}
\bibfield{author}{\bibinfo{person}{Beier Zhu}, \bibinfo{person}{Yulei Niu}, \bibinfo{person}{Yucheng Han}, \bibinfo{person}{Yue Wu}, {and} \bibinfo{person}{Hanwang Zhang}.} \bibinfo{year}{2022}\natexlab{a}.
\newblock \showarticletitle{Prompt-aligned gradient for prompt tuning}.
\newblock \bibinfo{journal}{\emph{Preprint at https://arxiv.org/abs/2205.14865}} (\bibinfo{year}{2022}).
\newblock


\bibitem[Zhu et~al\mbox{.}(2022b)]%
        {zhu2022pointclip}
\bibfield{author}{\bibinfo{person}{Xiangyang Zhu}, \bibinfo{person}{Renrui Zhang}, \bibinfo{person}{Bowei He}, \bibinfo{person}{Ziyao Zeng}, \bibinfo{person}{Shanghang Zhang}, {and} \bibinfo{person}{Peng Gao}.} \bibinfo{year}{2022}\natexlab{b}.
\newblock \showarticletitle{PointCLIP V2: Adapting CLIP for Powerful 3D Open-world Learning}.
\newblock \bibinfo{journal}{\emph{Preprint at https://arxiv.org/abs/2211.11682}} (\bibinfo{year}{2022}).
\newblock


\bibitem[Zhuang et~al\mbox{.}(2020)]%
        {zhuang2020comprehensive}
\bibfield{author}{\bibinfo{person}{Fuzhen Zhuang}, \bibinfo{person}{Zhiyuan Qi}, \bibinfo{person}{Keyu Duan}, \bibinfo{person}{Dongbo Xi}, \bibinfo{person}{Yongchun Zhu}, \bibinfo{person}{Hengshu Zhu}, \bibinfo{person}{Hui Xiong}, {and} \bibinfo{person}{Qing He}.} \bibinfo{year}{2020}\natexlab{}.
\newblock \showarticletitle{A comprehensive survey on transfer learning}.
\newblock \bibinfo{journal}{\emph{Proc. IEEE}} \bibinfo{volume}{109}, \bibinfo{number}{1} (\bibinfo{year}{2020}), \bibinfo{pages}{43--76}.
\newblock


\end{thebibliography}

\end{document}